\documentclass[export, accepted]{uai2022} 

\usepackage[american]{babel}

\usepackage{natbib} 
\usepackage{mathtools} 
\usepackage{booktabs} 
\usepackage{tikz} 

\newcommand{\GC}{ G_{\mb{c}} }
\newcommand{\GR}{ G_{\mb{r}} }

\newcommand{\CSet}{ \mc{S}_{\text{comp}} }

\newcommand{\traintask}{\mc{M}^\text{train}}
\newcommand{\testtask}{\mc{M}^\text{test}}

\newcommand{\subtask}{ \Phi }
\newcommand{\cond}{ \GC }
\newcommand{\subrew}{\GR}

\newcommand{\pieval}{ \pi^\text{eval} }
\newcommand{\piadapt}{ \pi^\text{adapt} }
\newcommand{\taup}{ \tau^\text{p} }

\newcommand{\mb}{\mathbf}
\newcommand{\tb}{\textbf}
\newcommand{\mbb}{\mathbb}
\newcommand{\mc}{\mathcal}
\newcommand{\wt}{\widetilde}

\DeclareRobustCommand\onedot{\futurelet\@let@token\@onedot}
\def\onedot{.}
\def\eg{\emph{e.g}\onedot} 
\def\ie{\emph{i.e}\onedot}

\usepackage{amsmath,amsfonts,bm}

\def\eqref#1{equation~\ref{#1}}

\def\1{\bm{1}}

\DeclareMathAlphabet{\mathsfit}{\encodingdefault}{\sfdefault}{m}{sl}
\SetMathAlphabet{\mathsfit}{bold}{\encodingdefault}{\sfdefault}{bx}{n}

\newcommand{\softmax}{\mathrm{softmax}}

\newcommand{\softplus}{\zeta}

\DeclareMathOperator*{\argmax}{arg\,max}

\newcommand{\amazon}{\textbf{Amazon}\xspace}
\newcommand{\apple}{\textbf{Apple}\xspace}
\newcommand{\bestbuy}{\textbf{BestBuy}\xspace}
\newcommand{\dicks}{\textbf{Dick's}\xspace}
\newcommand{\ebay}{\textbf{eBay}\xspace}
\newcommand{\expedia}{\textbf{Expedia}\xspace}
\newcommand{\ikea}{\textbf{Ikea}\xspace}
\newcommand{\lego}{\textbf{Lego}\xspace}
\newcommand{\lenox}{\textbf{Lenox}\xspace}
\newcommand{\omahasteaks}{\textbf{Omahasteaks}\xspace}
\newcommand{\swarovski}{\textbf{Swarovski}\xspace}
\newcommand{\thriftbooks}{\textbf{Thriftbooks}\xspace}
\newcommand{\todaytix}{\textbf{Todaytix}\xspace}
\newcommand{\walgreens}{\textbf{Walgreens}\xspace}
\newcommand{\walmart}{\textbf{Walmart}\xspace}
\def\wob{\textit{SymWoB}\xspace}
\def\mining{\textit{Mining}\xspace}

\newcommand{\figleft}{{\em (Left)}\xspace}

\newcommand{\figright}{{\em (Right)}\xspace}
\newcommand{\figtop}{{\em (Top)}\xspace}
\newcommand{\figbottom}{{\em (Bottom)}\xspace}

\newcommand{\figA}{\textbf{\color{blue} A}\xspace}
\newcommand{\figB}{\textbf{\color{blue} B}\xspace}
\newcommand{\figC}{\textbf{\color{blue} C}\xspace}
\usepackage{algorithm}
\usepackage{algorithmic}
\usepackage{boldline}
\usepackage{xspace}
\usepackage{comment}
\usepackage{microtype}
\usepackage{cleveref}
\usepackage{adjustbox}

\newcommand{\cutsectionup}{\vspace*{-0.05in}}
\newcommand{\cutsectiondown}{\vspace*{-0.0in}}

\newcommand{\cutparagraphup}{\vspace{-0pt}}

\newcommand{\cutitemizeup}{\vspace{-5pt}}
\newcommand{\cutitemizedown}{\vspace{-5pt}}

\title{Fast Inference and Transfer of Compositional Task Structures\\ for Few-shot Task Generalization}

\author[1, 2]{\href{mailto:<srsohn@umich.edu>}{Sungryull Sohn}{}}
\author[1]{Hyunjae Woo}
\author[1]{Jongwook Choi}
\author[1]{Lyubing Qiang}
\author[3]{Izzeddin Gur}
\author[3]{Aleksandra Faust}
\author[1,2]{Honglak Lee}

\affil[1]{%
    University of Michigan
}
\affil[2]{%
    LG AI Research Center Ann Arbor
}
\affil[3]{%
    Google Research
}
  
\begin{document}
\maketitle

\begin{abstract}
We tackle real-world problems with complex structures beyond the pixel-based game or simulator.
We formulate it as a few-shot reinforcement learning problem where a task is characterized by a subtask graph that defines a set of subtasks and their dependencies that are unknown to the agent.  
Different from the previous meta-RL methods trying to directly infer the unstructured task embedding, our multi-task subtask graph inferencer (MTSGI) first infers the common high-level task structure in terms of the subtask graph from the training tasks, and use it as a prior to improve the task inference in testing.
Our experiment results on 2D grid-world and complex web navigation domains show that the proposed method can learn and leverage the common underlying structure of the tasks for faster adaptation to the unseen tasks than various existing algorithms such as meta reinforcement learning, hierarchical reinforcement learning, and other heuristic agents.
\end{abstract}

\begin{figure*}[!htp]
    \centering
    \includegraphics[draft=false,width=0.98\linewidth, valign=b]{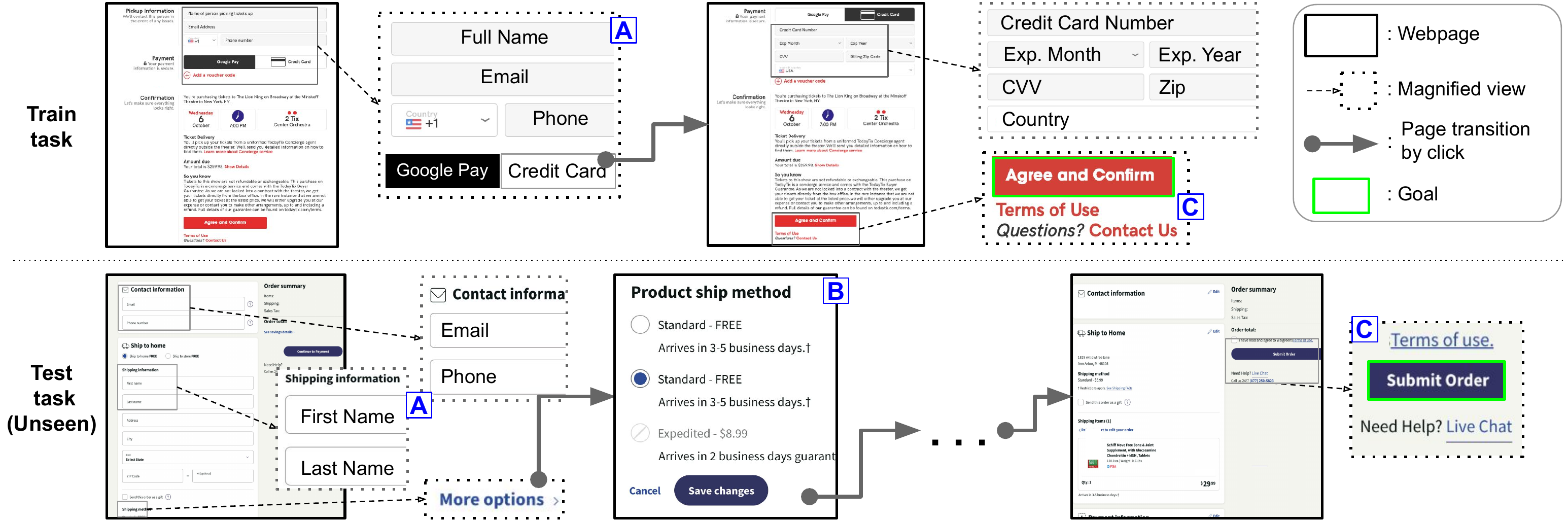}%
    \caption{%
        An illustration of the train \figtop and test task \figbottom in our \wob domain. Some selected actionable web-elements (\eg, text fields and buttons) are magnified (dotted arrow and box) for readability. The agent’s goal ({\color{green} green box}) is to checkout the products in unseen test website by interacting with the web elements in a correct order. 
        For example, in train task, the agent should fill out all the text fields in ({\em Top}, \figA) \textbf{before} clicking the \texttt{credit\_card} button to transition ({\color{gray} gray arrow}) to next page.
        The high-level checkout processes in different websites have many commonalities while certain details may differ. 
        For example, in both train and test tasks, the agent should fill out the user information ({\em Top} and {\em Bottom}, \figA) before proceeding to the next page or there exist similar elements ({\em Top} and {\em Bottom}, \figC). However, the details may differ; \eg, the train task ({\em Top}, \figA) has a single text field for full name, while the test task ({\em Bottom}, \figA) has separate text fields for the first and last name, respectively. Also, only the test website ({\em Bottom}, \figB) requires shipping information since the training website does not ship the product.
    }
    \label{fig:task_overview}
\end{figure*}
\cutsectionup
\section{Introduction}\label{sec:i}
\cutsectiondown
Recently, deep reinforcement learning (RL) has shown an outstanding performance on various domains such as video games~\citep{mnih2015human, vinyals2019grandmaster} and board games~\citep{silver2017mastering}.
However, most of the successes of deep RL were focused on a single-task setting where the agent is allowed to interact with the environment for hundreds of millions of time steps. 
In numerous real-world scenarios, interacting with the environment is expensive or limited, and the agent is often presented with a novel task that is not seen during its training time.
%
To overcome this limitation, many recent works focused on scaling the RL algorithm beyond the single-task setting.
Recent works on multi-task RL aim to build a single, contextual policy that can solve multiple related tasks and generalize to unseen tasks.
However, they require a certain form of task embedding as an extra input that often fully characterizes the given task~\citep{oh2017zero, Andreas2017Modular, yu2017deep, chaplot2018aaai}, or requires a human demonstration~\cite{huang2018neural}, which are not readily available in practice.
%
Meta RL~\citep{finn2017model,duan2016rl} focuses on a more general setting where the agent should learn about the unseen task purely via interacting with the environment without any additional information. However, such meta-RL algorithms either require a large amount of experience on the diverse set of tasks or are limited to a relatively smaller set of simple tasks with a simple task structure.

%
On the contrary, real-world problems require the agent to solve much more complex and compositional tasks without human supervision.
Consider a web-navigating RL agent given the task of checking out the products from an online store as shown in~\Cref{fig:task_overview}.
The agent can complete the task by filling out the required web elements with the correct information such as shipping or payment information, navigating between the web pages, and placing the order.
Note that the task consists of multiple \textit{subtasks} and the subtasks have complex dependencies in the form of \textit{precondition};
for instance, the agent may proceed to the payment web page (see {\em Bottom}, {\color{blue} B}) \textit{after} all the required shipping information has been correctly filled in (see {\em Bottom}, {\color{blue} A}), or the \texttt{credit\_card\_number} field will appear \textit{after} selecting the \texttt{credit\_card} as a payment method (see {\em Top, Middle} in~\Cref{fig:task_overview}).
Learning to perform such a task can be quite challenging if the reward is given only after yielding meaningful outcomes (\ie, sparse reward task).
This is the problem scope we focus on in this work: solving and generalizing to unseen compositional sparse-reward tasks with complex subtask dependencies without human supervision.

%
Recent works~\citep{sohn2019meta, NTP, huang2018neural, liu2016GTSvisual, ghazanfari2017autonomous} tackled the compositional tasks by explicitly inferring the underlying task structure in a graph form.
Specifically, the subtask graph inference (SGI) framework~\citep{sohn2019meta} uses inductive logic programming (ILP) on the agent's own experience to infer the task structure in terms of \textit{subtask graph} and learns a contextual policy to \textit{execute} the inferred task in few-shot RL setting.
However, it only meta-learned the adaptation policy that relates to the efficient exploration, while the task inference and execution policy learning were limited to a single task (\ie, both task inference and policy learning were done from scratch for each task), limiting its capability of handling large variance in the task structure.
We claim that the inefficient task inference may hinder applying the SGI framework to a more complex domain such as web navigation~\citep{shi2017world, liu2018reinforcement} where a task may have a large number of subtasks and complex dependencies between them.
We note that humans can navigate an unseen website by transferring the high-level process learned from previously seen websites.

\begin{figure*}[!htp]
    \centering
    \includegraphics[draft=false,width=0.98\linewidth, valign=b]{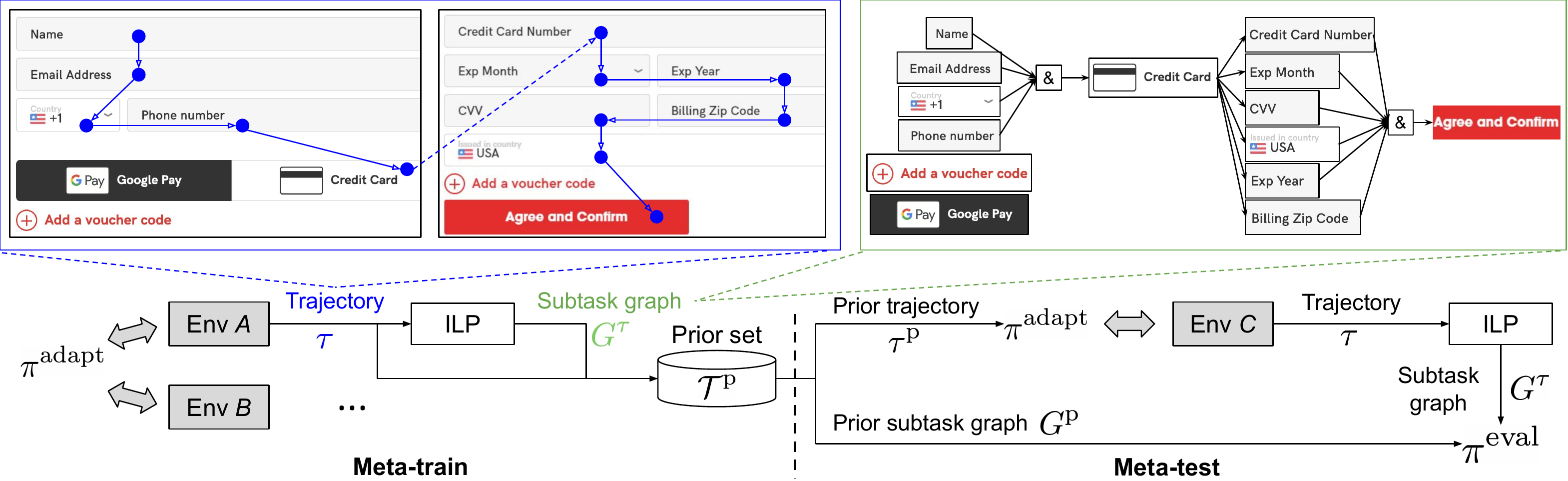}%
    \vspace{-6pt}
    \caption{%
        The overview of our algorithm and the example of agent's {\color{blue} trajectory} and the inferred {\color[rgb]{0.5,0.75,0.4}subtask graph}. In meta-train \figleft, the adaptation policy $\piadapt$ interacts with the environment and collects the trajectory $\tau$. The inductive logic programming (ILP) module takes as input the trajectory, and infers the task structure in terms of the subtask graph $G^\tau$. The trajectory and the subtask graph are stored as a prior. In meta-testing \figright, the adaptation policy incorporates the prior trajectory $\tau^\text{p}$ to efficiently explore the environment, and ILP module infers the subtask graph $G^\tau$ from the adaptation trajectory $\tau$. Finally, the evaluation policy $\pieval$ takes as input the prior and inferred subtask graphs $(G^\text{p}, G^\tau)$ to solve the test task.
    }
    \label{fig:method_overview}
\end{figure*}

Inspired by this, we extend the SGI framework to a \textit{multi-task subtask graph inferencer} (MTSGI) that can generalize the previously learned task structure to the unseen task for faster adaptation and stronger generalization.~\Cref{fig:method_overview} outlines our method.
MTSGI estimates the prior model of the subtask graphs from the training tasks. When an unseen task is presented, MTSGI samples the prior that best matches with the current task, and incorporates the sampled prior model to improve the latent subtask graph inference, which in turn improves the performance of the evaluation policy.
We demonstrate results in the 2D grid-world domain and the web navigation domain that simulates the interaction with 15 actual websites.
We compare our method with MSGI~\citep{sohn2019meta} that learns the task hierarchy from scratch for each task, and two other baselines including hierarchical RL and a heuristic algorithm. 
We find that MTSGI significantly outperforms all other baselines, and the learned prior model enables more efficient task inference compared to MSGI.
\cutsectionup
\section{Preliminaries}\label{sec:pre}
\cutsectiondown

\paragraph{Few-shot Reinforcement Learning}
\label{sec:metarl}
A \emph{task} is defined by an MDP $\mc{M}_{G}=(\mc{S, A}, \mc{P}_{G}, \mc{R}_{G})$ parameterized by a task parameter $G$ 
with a set of states $\mc{S}$, a set of actions $\mc{A}$, transition dynamics $\mc{P}_{G}$, reward function $\mc{R}_{G}$.
The goal of $K$-shot RL~\citep{duan2016rl,finn2017model},
is to efficiently solve a distribution of unseen test tasks $\testtask{}$ by learning and transferring the common knowledge from the training tasks $\traintask{}$.
It is assumed that the training and test tasks do not overlap (\ie, $\traintask\cap \testtask = \emptyset$) but share a certain commonality such that the knowledge learned from the training tasks may be helpful for learning the test tasks.
For each task $\mc{M}_G$, the agent is given $K$ steps budget for interacting with the environment.
During meta-training, the goal of multi-task RL agent is to learn a prior (\ie, slow-learning) over the training tasks $\traintask{}$. 
Then, the learned prior may be exploited during the meta-test to enable faster adaptation on unseen test tasks $\testtask$.
For each task, the agent faces two phases:
an \emph{adaptation phase} where the agent learns a task-specific behavior (\ie, fast-learning) for $K$ environment steps, which often spans over multiple episodes,
and a \emph{evaluation phase} where the adapted behavior is evaluated.
In the evaluation phase, the agent is not allowed to perform any form of learning, and agent's performance on the task $\mc{M}_G$ is measured in terms of the return:
\begin{align}
    \mc{R}_{\mc{M}_{G}}(\pi_{\phi_K}) = \mathbb{E}_{\pi_{\phi_K}, \mc{M}_{G} }\left[\textstyle\sum^{H}_{t=1}{r_t}\right],\label{eq:loss}
\end{align}
where $\pi_{\phi_{K}}$ is the policy after $K$ update steps of adaptation,
$H$ is the horizon of evaluation phase,
and $r_t$ is the reward at time $t$ in the evaluation phase.

\cutsectionup
\section{Subtask Graph Inference Problem}\label{sec:problem}
\cutsectiondown
The \textit{subtask graph inference} problem~\citep{sohn2019meta} is a few-shot RL problem where a task is parameterized by a set of subtasks and their dependencies.
Formally, a task consists of $N$ subtasks $\mathbf{\Phi}=\{\subtask^1, \ldots, \subtask^N\}$, and each subtask $\subtask^i$ is parameterized by a tuple ($\CSet{}^i, \cond^i, \subrew^i$).
The \textit{goal state} $\CSet{}^i\subset \mc{S}$ and \textit{precondition} $\cond^i: \mc{S} \rightarrow \{0, 1\}$ defines the condition that a subtask is \textit{completed}: the current state should be contained in its goal states  (\ie, $\mb{s}_t\in\smash{\CSet^i}$) and the precondition should be satisfied (\ie, $\cond^i(\mb{s}_t)=1$).
If the precondition is not satisfied (\ie, $\cond^i(\mb{s}_t)=0$), the subtask cannot be completed and the agent receives no reward even if the goal state is achieved.
The \textit{subtask reward function} $\subrew^i$ defines the amount of reward given to the agent when it \textit{completes} the subtask $i$: $r_t\sim\subrew^i$.
We note that the subtasks $\{\subtask^1, \ldots, \subtask^N\}$ are unknown to the agent. Thus, the agent should learn to infer the underlying task structure and complete the subtasks in an optimal order while satisfying the required preconditions.

\paragraph{State}
In the subtask graph inference problem, it is assumed that the state input provides the high-level status of the subtasks. 
Specifically, the state consists of the followings: $\mb{s}_t=(\text{obs}_t, \mb{x}_t, \mb{e}_t, \text{step}_{\text{epi}, t}, \text{step}_{\text{phase}, t})$.
The $\text{obs}_t\in\{0, 1\}^{W\times H\times C}$ is a visual observation of the environment.
The completion vector $\smash{\mb{x}_t\in\{0, 1\}^N }$ indicates whether each subtask is complete. The eligibility vector $\smash{\mb{e}_t\in\{0, 1\}^N }$ indicates whether each subtask is eligible (\ie, precondition is satisfied). Following the few-shot RL setting, the agent observes two scalar-valued time features: the remaining time steps until the episode termination $\text{step}_{\text{epi}, t}\in\mathbb{R}$ and the remaining time steps until the phase termination $\text{step}_{\text{phase}, t}\in\mathbb{R}$.

\paragraph{Options} 
For each subtask $\Phi^i$, the agent can learn an option $\mc{O}^i$~\citep{sutton1999between} that reaches the goal state of the subtask.
Following~\cite{sohn2019meta}, such options are pre-learned individually by maximizing the goal-reaching reward: $r_t=\mbb{I}(\mb{s}_t\in\smash{\CSet^i})$. At time step $t$, we denote the option taken by the agent as $\mb{o}_t$ and the binary variable that indicates whether episode is terminated as $d_t$.
\cutsectionup
\section{Method}\label{sec:m}
\cutsectiondown

We propose a novel Multi-Task Subtask Graph Inference (MTSGI) framework that can perform an efficient inference of latent task embedding (\ie, subtask graph). The overall method is outlined in~\Cref{fig:method_overview}.
Specifically, in meta-training, MTSGI models the prior in terms of (1) adaptation trajectory $\tau$ and (2) subtask graph $G$ from the agent's experience. In meta-testing, MTSGI samples (1) the prior trajectory $\tau^\text{p}$ for more efficient exploration in adaptation and (2) the prior subtask graph $G^\text{p}$ for more accurate task inference.

\subsection{Multi-task Adaptation Policy}\label{sec:adapt-policy}
The goal of \textit{adaptation policy} is to efficiently explore and gather the information about the task. 
Intuitively, if the adaptation policy completes more diverse subtasks, then it can provide more data to the task inference module (ILP), which in turn can more accurately infer the task structure.
To this end, we extend the upper confidence bound (UCB)-based adaptation policy proposed in~\citet{sohn2019meta} as follows:
\begin{align}
\piadapt(o=\mc{O}^i \mid s) \propto \exp \left(r^i+\sqrt{2}\frac{\log \left(\sum_{j} n^j\right)}{n^i}\right),\label{eq:UCB-policy}
\end{align}
where $r^i$ is the empirical mean of the reward received after executing subtask $i$ and $n^i$ is the number of times subtask $i$ has been executed within the current task. Note that the exploration parameters $\{r^i, n^i\}_{i=1}^N$ can be computed from the agent's trajectory.
In meta-train, the exploration parameters are initialized to zero when a new task is sampled. In meta-test, the exploration parameters are initialized with those of the sampled prior. Intuitively, this helps the agent visit novel states that were unseen during meta-training.

\begin{algorithm}[t]
\caption{Meta-training: learning the prior}\label{alg:meta-train}
\begin{algorithmic}[1]
\REQUIRE{ Adaptation policy $\piadapt$}
\ENSURE{ Prior set $\mc{T}^{\text{p}}$}
\STATE $\mc{T}^{\text{p}} \gets \emptyset$
\FOR{each task $\mc{M}\in\traintask$}
    \STATE Rollout adaptation policy: \\
    $\tau=\{\mb{s}_t, \mb{o}_t,r_t, d_t\}_{t=1}^{K} \sim \piadapt$ in task $\mc{M}$
    \STATE Infer subtask graph $G^\tau = \argmax_{G} p(\tau|G)$ 
    \STATE $\pieval= \text{GRProp}(G^\tau)$
    \STATE Evaluate the agent: $\tau^\text{eval}\sim \pieval$ in task $\mc{M}$
    \STATE Update prior $\mc{T}^{\text{p}}\gets \mc{T}^{\text{p}} \cup \left( G^\tau,  \tau\right)$
\ENDFOR
\end{algorithmic}
\end{algorithm}

\subsection{Meta-train: Learning the Prior Subtask Graph}
Let $\tau$ be an adaptation trajectory of the agent for $K$ steps.
The goal is to infer the latent subtask graph $G$ for the given training task $\mc{M}_{G}\in\traintask$,
specified by preconditions $\GC{}$ and subtask rewards $\GR{}$.
We find the maximum-likelihood estimate (MLE) of $G = (\GC{}, \GR{})$
that maximizes the likelihood of the adaptation trajectory $\tau$: 
\begin{align}
    \widehat{G}^\text{MLE} = \argmax_{ \GC{}, \GR{} } p(\tau| \GC{}, \GR{} ).  \label{eq:objective}
\end{align}
Following~\citet{sohn2019meta}, we infer the precondition $\GC{}$ and the subtask reward $\GR{}$ as follows (See Appendix for the detailed derivation):
\begin{align}
\widehat{G}_\mb{c}^\text{MLE}&=\argmax_{\GC{}} \prod_{t=1}^{H}{ p(\mb{e}_{t}|\mb{x}_{t}, \GC{}) },\label{eq:infer-precond}\\
\widehat{G}_\mb{r}^\text{MLE}&=\argmax_{\GR{}} \prod_{t=1}^{H}{ p(r_t|\mb{e}_t,\mb{o}_t, \GR{})}.\label{eq:infer-reward}
\end{align}
where $\mb{e}_t$ is the eligibility vector, $\mb{x}_t$ is the completion vector, $\mb{o}_t$ is the option taken, $r_t$ is the reward at time step $t$.

\paragraph{Precondition inference} 
The problem in~\Cref{eq:infer-precond} is known as the inductive logic programming (ILP) problem that finds a boolean function that satisfies all the indicator functions.
Specifically, $\{\mb{x}_{t}\}^{H}_{t=1}$ forms binary vector inputs to programs,
and $\{e^{i}_{t}\}^{H}_{t=1}$ forms Boolean-valued outputs of the $i$-th program that predicts the eligibility of the $i$-th subtask.
We use the \textit{classification and regression tree} (CART) to infer the precondition function $f_{\GC{}}: \mb{x}\rightarrow \mb{e}$ for each subtask based on Gini impurity~\citep{breiman1984classification}. 
Intuitively, the constructed decision tree is the simplest boolean function approximation for the given input-output pairs $\{\mb{x}_t, \mb{e}_t\}$.
The decision tree is converted to a logic expression (\ie, precondition) in sum-of-product (SOP) form to build the subtask graph.

\paragraph{Subtask reward inference} 
To infer the subtask reward $\widehat{G}_\mb{r}^{\text{MLE}}$ in~\Cref{eq:infer-reward},
we model the reward for $i$-th subtask as a Gaussian distribution: $G_\mb{r}^{i} \sim \mathcal{N}(\widehat{\mu}^{i}, \widehat{\sigma}^{i} )$.
Then, the MLE of subtask reward is given as the empirical mean and variance of the rewards received after taking the eligible option $\mc{O}^i$ in adaptation phase:
\begin{align}
\widehat{\mu}^{i}_{\text{MLE}}
 &   = \mathbb E \left[ r_t|\mb{o}_t=\mc{O}^i, \mb{e}_t^i=1 \right],\label{eq:infer-reward-mu}\\
\widehat{\sigma^2}^{i}_{\text{MLE}}
 &   = \mathbb E \left[ (r_t - \widehat{\mu}^{i}_{\text{MLE}})^2 |\mb{o}_t=\mc{O}^i, \mb{e}_t^i=1 \right],\label{eq:infer-reward-sigma}
\end{align}
where  $\mc{O}^i$ is the option corresponding to the $i$-th subtask.
\Cref{alg:meta-train} outlines the meta-training process.

\begin{algorithm}[t]
\caption{meta-testing: multi-task SGI}\label{alg:meta-eval}
\begin{algorithmic}[1]
\REQUIRE{  Adaptation policy $\piadapt$, prior set $\mc{T}^{\text{p}}$}
\FOR{each task $\mc{M}\in\testtask$}
    \STATE Sample prior: $(G^{\text{p}}, \taup) \sim p(\mc{T}^{\text{p}})$
    %
    \STATE Rollout adaptation policy:\\
    $\tau=\{\mb{s}_t, \mb{o}_t,r_t, d_t\}_{t=1}^{K} \sim \piadapt$ in task $\mc{M}$
    \STATE Infer subtask graph $G^{\tau} = \argmax_{G} p(\tau|G)$ 
    \STATE $\pieval(\cdot| \tau, \taup)
\propto \text{GRProp}(\cdot| G^{\tau})^\alpha\text{GRProp}(\cdot|G^\text{p})^{(1-\alpha)}$
    \STATE Evaluate the agent: $\tau^\text{eval}\sim \pieval$ in task $\mc{M}$
\ENDFOR
\end{algorithmic}
\end{algorithm}

\subsection{Evaluation: Graph-reward Propagation Policy}\label{sec:test-policy}
In both meta-training and meta-testing, the agent's adapted behavior is evaluated during the test phase. Following~\citet{sohn2019meta}, we adopted the graph reward propagation (GRProp) policy as an evaluation policy $\pieval$ that takes as input the inferred subtask graph $\widehat{G}$ and outputs the subtasks to execute to maximize the return.
Intuitively, the GRProp policy approximates a subtask graph to a differentiable form
such that we can compute the gradient of return with respect to the completion vector to measure how much each subtask is likely to increase the return.
Due to space limitations, we give a detail of the GRProp policy in Appendix.
The overall meta-training process is summarized in Appendix.

\subsection{Meta-testing: Multi-task Task Inference}

\paragraph{Prior sampling} 
In meta-testing, MTSGI first chooses the prior task that most resembles the given evaluation task. Specifically, we define the pair-wise similarity between a prior task $\mathcal{M}_G^\text{prior }$ and the evaluation task $\mathcal{M}_G$ as follows:
\begin{align}
\operatorname{sim}\left(\mathcal{M}_G, \mathcal{M}_G^\text{prior }\right)=F_{\beta}\left(\mathbf{\Phi}, \mathbf{\Phi}^{\text {prior }}\right)+\kappa R\left(\tau^{\text {prior }}\right) ,
\end{align}
where $F_{\beta}$ is the F-score with weight parameter $\beta$, $\mathbf{\Phi}$ is the subtask set of $\mc{M}_{G}$, $\mathbf{\Phi}^{\text {prior }}$ is the subtask set of $\mathcal{M}_G^\text{prior }$, $R\left(\tau^{\text {prior }}\right)$ is the agent's empirical performance on the prior task $\mathcal{M}_G^\text{prior }$, and $\kappa$ is a scalar-valued weight which we used $\kappa=1.0$ in experiment.
$F_{\beta}$ measures how many subtasks overlap between current and prior tasks in terms of precision and recall as follows:
\begin{align}
F_{\beta}&=\left(1+\beta^{2}\right) \cdot \frac{\text { precision } \cdot \text { recall }}{\left(\beta^{2} \cdot \text { precision }\right)+\text { recall }},\\
\text{Precision} &= \lvert \mathbf{\Phi} \cap \mathbf{\Phi}^{\text {prior }}\rvert / \lvert  \mathbf{\Phi}^{\text {prior }} \rvert,\\
\text{Recall} &=  \lvert \mathbf{\Phi} \cap \mathbf{\Phi}^{\text {prior }}\rvert / \lvert \mathbf{\Phi}\rvert.
\end{align}
We used $\beta=10$ to assign a higher weight to the current task (\ie, recall) than the prior task (\ie, precision). 

\paragraph{Multi-task subtask graph inference} 
Let $\tau$ be the adaptation trajectory, and $\taup$ be the sampled prior adaptation trajectory. Then, we model our evaluation policy as follows:
\begin{align}
\pi(o|s,\tau, \taup)
\simeq \pi(o|s, G^{\tau})^\alpha\pi(o|s, G^\text{p})^{(1-\alpha)}.\label{eq:policy-mix}
\end{align}
Due to the limited space, we include the detailed derivation of~\Cref{eq:policy-mix} in Appendix. Finally, we deploy the GRProp policy as a contextual policy:
\begin{align}
\pieval(\cdot| \tau, \taup)
= \text{GRProp}(\cdot| G^{\tau})^\alpha\text{GRProp}(\cdot|G^\text{p})^{(1-\alpha)}.\label{eq:MTSGI-policy-mix}
\end{align}
Note that~\Cref{eq:MTSGI-policy-mix} is the weighted sum of the logits of two GRProp policies induced by prior $\taup$ and current experience $\tau$.
We claim that such form of ensemble induces the positive transfer in compositional tasks.
Intuitively, ensembling GRProp is taking a union of preconditions since GRProp assigns a positive logit to task-relevant subtask and non-positive logit to other subtasks.
As motivated in the Introduction, related tasks often share the task-relevant preconditions; thus, taking the union of task-relevant preconditions is likely to be a positive transfer and improve the generalization.
The pseudo-code of the multi-task subtask graph inference process is summarized in~\Cref{alg:meta-eval}.

\cutsectionup
\section{Related Work}\label{sec:r}
\cutsectiondown
\paragraph{Web navigating RL agent}
Previous work introduced MiniWoB \citep{shi2017world} and MiniWoB++ \citep{liu2018reinforcement} benchmarks that are manually curated sets of simulated toy environments for the web navigation problem.
They formulated the problem as acting on a page represented as a Document Object Model (DOM), a hierarchy of objects in the page.
The agent is trained with human demonstrations and online episodes in an RL loop.
\citet{jia2018domqnet} proposed a graph neural network based DOM encoder and a multi-task formulation of the problem similar to this work.
\citet{gur2018learning} introduced a manually-designed curriculum learning method and an LSTM based DOM encoder.
DOM level representations of web pages pose a significant sim-to-real gap as simulated websites are considerably smaller (100s of nodes) compared to noisy real websites (1000s of nodes).
As a result, these models are trained and evaluated on the same simulated environments which are difficult to deploy on real websites.
Our work formulates the problem as abstract web navigation on real websites where the objective is to learn a latent subtask dependency graph similar to a sitemap of websites.
We propose a multi-task training objective that generalizes from a fixed set of real websites to unseen websites without any demonstration, illustrating an agent capable of navigating real websites for the first time.
\paragraph{Meta-reinforcement learning}
To tackle the few-shot RL problem,
researchers have proposed two broad categories of meta-RL approaches: RNN- and gradient-based methods.
The RNN-based meta-RL methods~\citep{duan2016rl, wang2016learning, hochreiter2001learning} encode the common knowledge of the task into the hidden states and the parameters of the RNN. 
The gradient-based meta-RL methods~\citep{finn2017model,nichol2018first,Gupta:1802.07245,Finn2018PMAML,Kim:2018:BMAML}
encode the task embedding in terms of the initial policy parameter for fast adaptation through meta gradient.
Existing meta-RL approaches, however, often require a large amount of environment interaction due to the long-horizon nature of the few-shot RL tasks.
Our work instead explicitly infers the underlying task parameter in terms of subtask graph, which can be efficiently inferred using the inductive logic programming (ILP) method and be transferred across different, unseen tasks.
%
%
\paragraph{More Related Works}
Please refer to the Appendix for further discussions about other related works.
\begin{figure*}[!htp]
    \centering
    \includegraphics[draft=false,width=0.19\linewidth, valign=b]{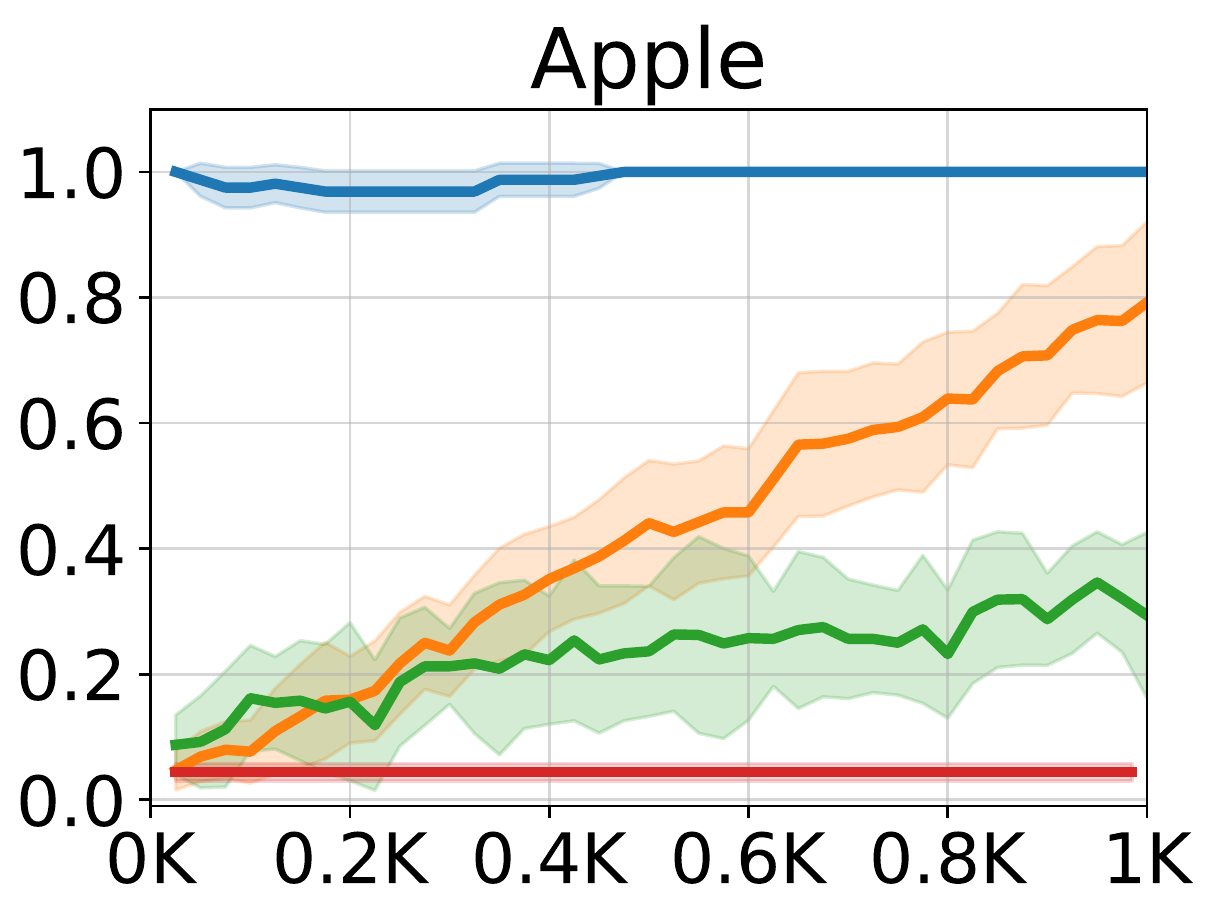}
    \includegraphics[draft=false,width=0.19\linewidth, valign=b]{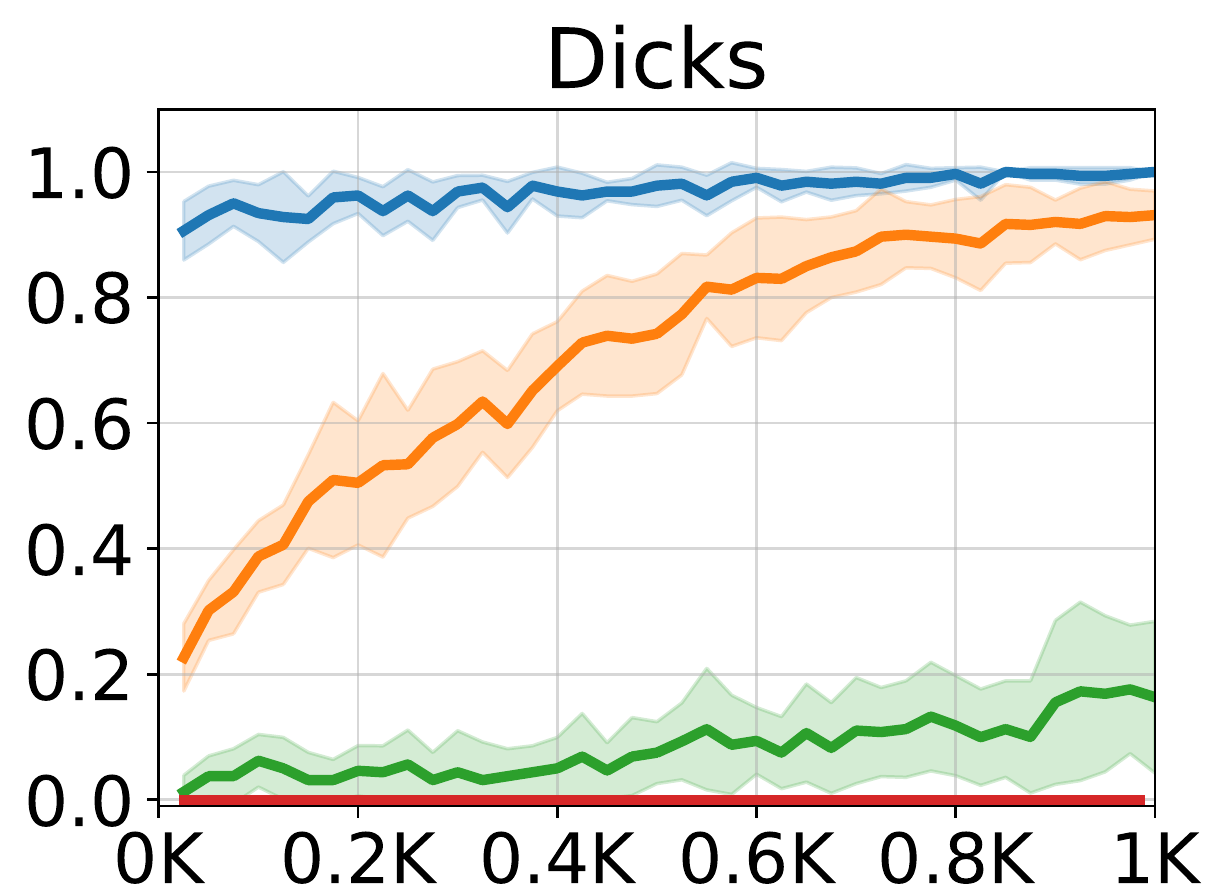}
    \includegraphics[draft=false,width=0.185\linewidth, valign=b]{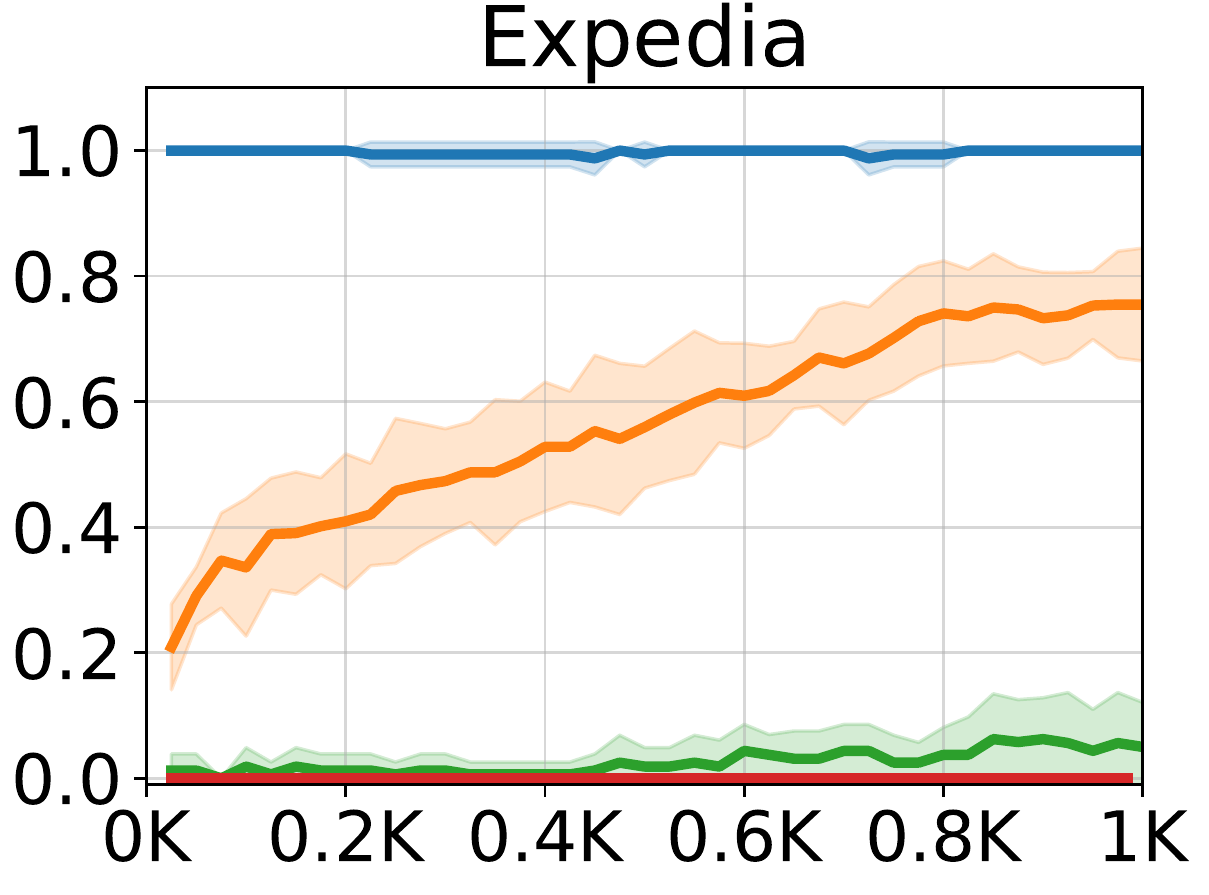}
    \includegraphics[draft=false,width=0.19\linewidth, valign=b]{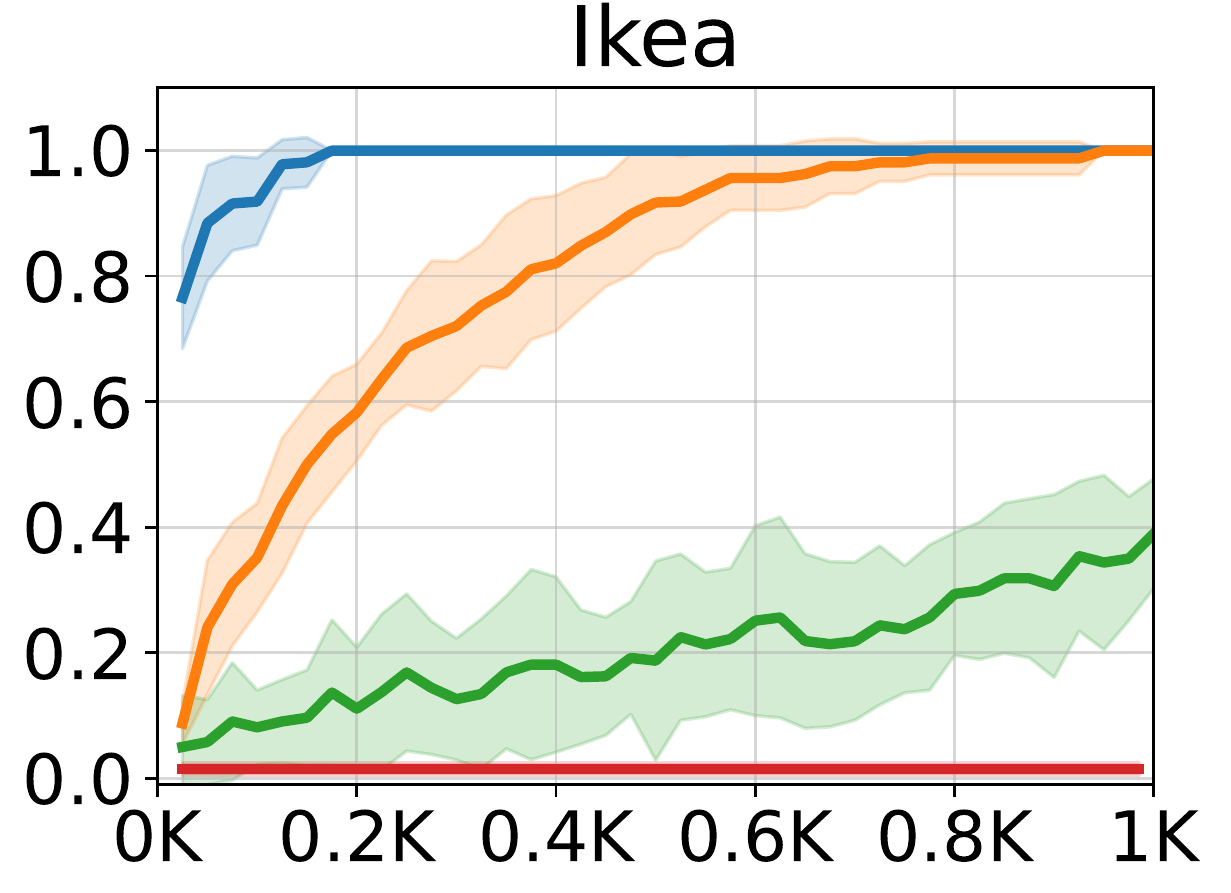}
    \includegraphics[draft=false,width=0.19\linewidth, valign=b]{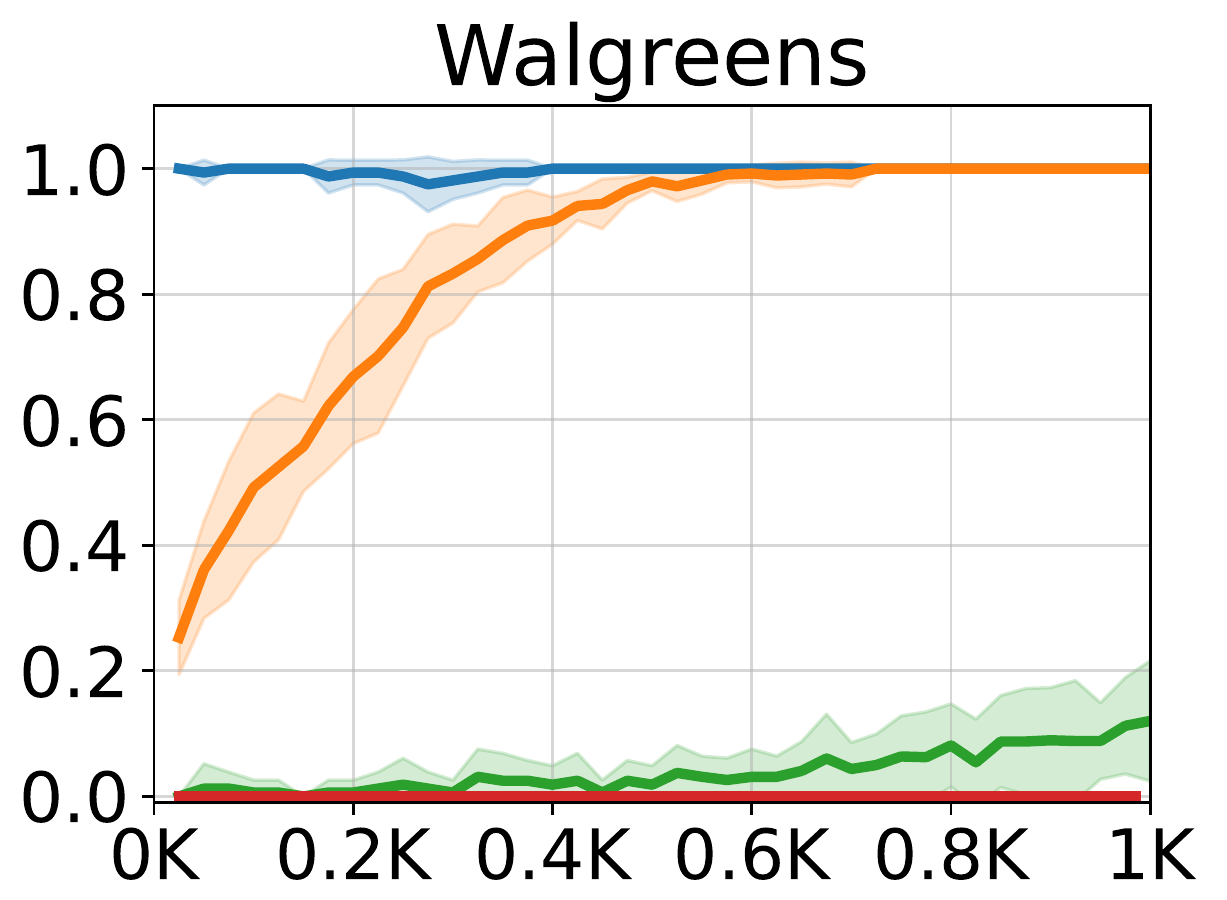}\\
    \includegraphics[draft=false,width=0.98\linewidth, valign=b]{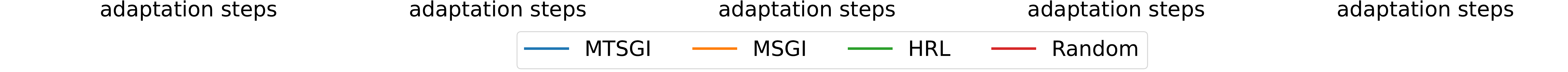}
    \vspace{-8pt}
    \caption{%
        The success rate (y-axis) of the compared methods in the test phase in terms of the environment step during the adaptation phase (x-axis) on \wob{} domain. See Appendix for the results on other tasks.
    }
    \label{fig:fewshot_wob}
    \vspace{-7pt}
\end{figure*}

\section{Experiment}
\subsection{Domains}
%
\paragraph{Mining}
\mining~\citep{sohn2018hierarchical} is a 2D grid-world domain inspired by Minecraft game
where the agent receives a reward by picking up raw materials in the world or crafting items with raw materials.
The subtask dependency in \mining{} domain comes from the crafting recipe implemented in the game. 
Following~\citet{sohn2018hierarchical}, we used the pre-generated training/testing task splits generated with four different random seeds. Each split set consists of 3200 training tasks and 440 testing tasks for meta-training and meta-testing, respectively. We report the performance averaged over the four task split sets.

\paragraph{SymWoB}
We implement a symbolic version of the checkout process on the 15 real-world websites such as \amazon{}, \bestbuy{}, and \walmart{}, etc. 

\textbf{Subtask and option policy.}
Each actionable web element (\eg, text field, button drop-down list, and hyperlink) is considered as a subtask. We assume the agent has pre-learned the option policies that correctly interact with each element (\eg, click the button or fill out the text field). Thus, the agent should learn a policy over the option.

\textbf{Completion and eligibility.}
For each subtask, the completion and eligibility are determined based on the status of the corresponding web element. 
For example, the subtask of a text field is \textit{completed} if the text field is filled with the correct information, and the subtask of a \texttt{confirm\_credit\_info} button is \textit{eligible} if all the required subtasks (\ie, filling out credit card information) on the webpage are completed. Executing an option will complete the corresponding subtask \textit{only if} the subtask is eligible.

\textbf{Reward function and episode termination.}
The agent may receive a non-zero reward only at the end of the episode (\ie, sparse-reward task). When the episode terminates due to the time budget, the agent may not receive any reward. Otherwise, the following  two types of subtasks terminate the episode and give a non-zero reward upon completion:
\cutitemizeup
\begin{itemize}
\setlength{\itemsep}{1pt}\setlength{\parskip}{1pt}
\item{\tb{Goal subtask}} refers to the button that completes the order (See the green boxes in~\Cref{fig:task_overview}). Completing this subtask gives the agent a +5 reward, and the episode is terminated. 
\item{\tb{Distractor subtask}} does not contribute to solving the given task but terminates the episode with a -1 reward. It models the web elements that lead to external web pages such as \texttt{Contact\_Us} button in~\Cref{fig:task_overview}. 
\end{itemize}
\cutitemizedown

\textbf{Transition dynamics.}
The transition dynamics follow the dynamics of the actual website. 
Each website consists of multiple web pages.
The agent may only execute the subtasks that are currently visible (\ie, on the current web page) and can navigate to the next web page only after filling out all the required fields and clicking the continue button.
The goal subtask is present in the last web page; thus, the agent must learn to navigate through the web pages to solve the task.

For more details about each task, please refer to Appendix.

\subsection{Agents}
We compared the following algorithms in the experiment.
\cutitemizeup
\begin{itemize}
\setlength{\itemsep}{1pt}\setlength{\parskip}{1pt}
\item MTSGI (Ours): our multi-task SGI agent
\item MSGI~\citep{sohn2019meta}: SGI agent without multi-task learning
\item HRL: an Option~\citep{sutton1999between}-based proximal policy optimization (PPO)~\citep{schulman2017proximal} agent with the gated rectifier unit (GRU)
\item Random: a heuristic policy that uniform randomly executes an eligible subtask
\end{itemize}
\cutitemizedown
More details on the architectures and the hyperparameters can be found in Appendix.

\paragraph{Meta-training}
In \wob{}, for each task chosen for a meta-testing, we randomly sampled $N_\text{train}$ tasks among the remaining 14 tasks and used it for meta-training.
We used $N_\text{train}=1$ in the experiment (See~\Cref{fig:ablation_nprior} for the impact of the choice of $N_\text{train}$).
For example, we meta-trained our MTSGI on \amazon{} and meta-tested on \expedia{}. 
For \mining{}, we used the train/test task split provided in the environment.
The RL agents (\eg, HRL) were individually trained on each test task; the policy was initialized when a new task is sampled and trained during the adaptation phase.
All the experiments were repeated with four random seeds, where different training tasks were sampled for different seeds.

\subsection{Result: Few-shot Generalization Performance}

\begin{figure}[!t]
    \centering
    \includegraphics[draft=false,width=0.49\linewidth, valign=b]{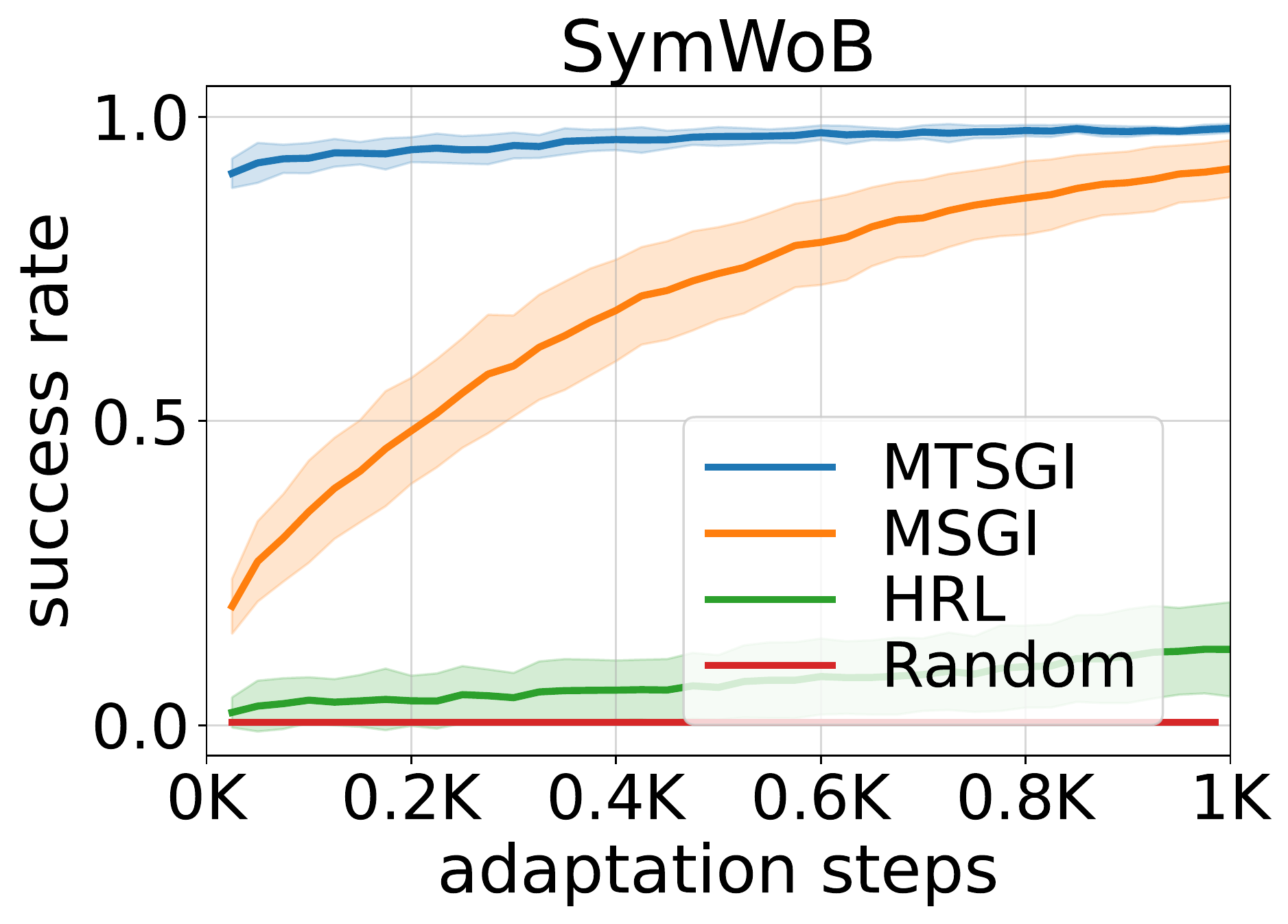}%
    \includegraphics[draft=false,width=0.49\linewidth, valign=b]{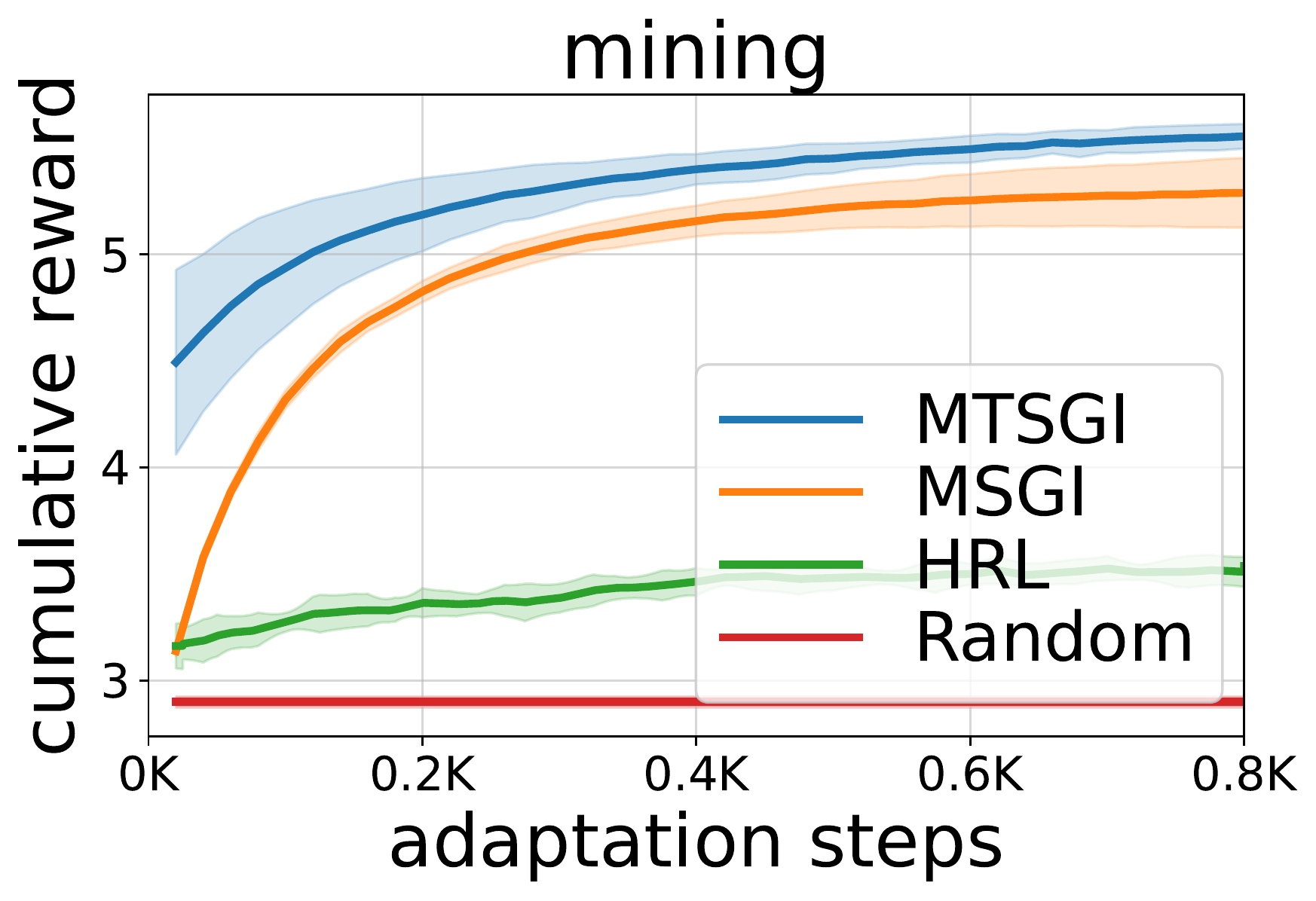}%
    \vspace{-8pt}
    \caption{%
        The performance of the compared methods in terms of the adaptation steps averaged over all the tasks in \wob{} \figleft and \mining{} \figright domains. 
    }
    \label{fig:fewshot_mining}
    \vspace{-7pt}
\end{figure}
\begin{figure}[!t]
    \centering
    \includegraphics[draft=false,width=0.49\linewidth, valign=b]{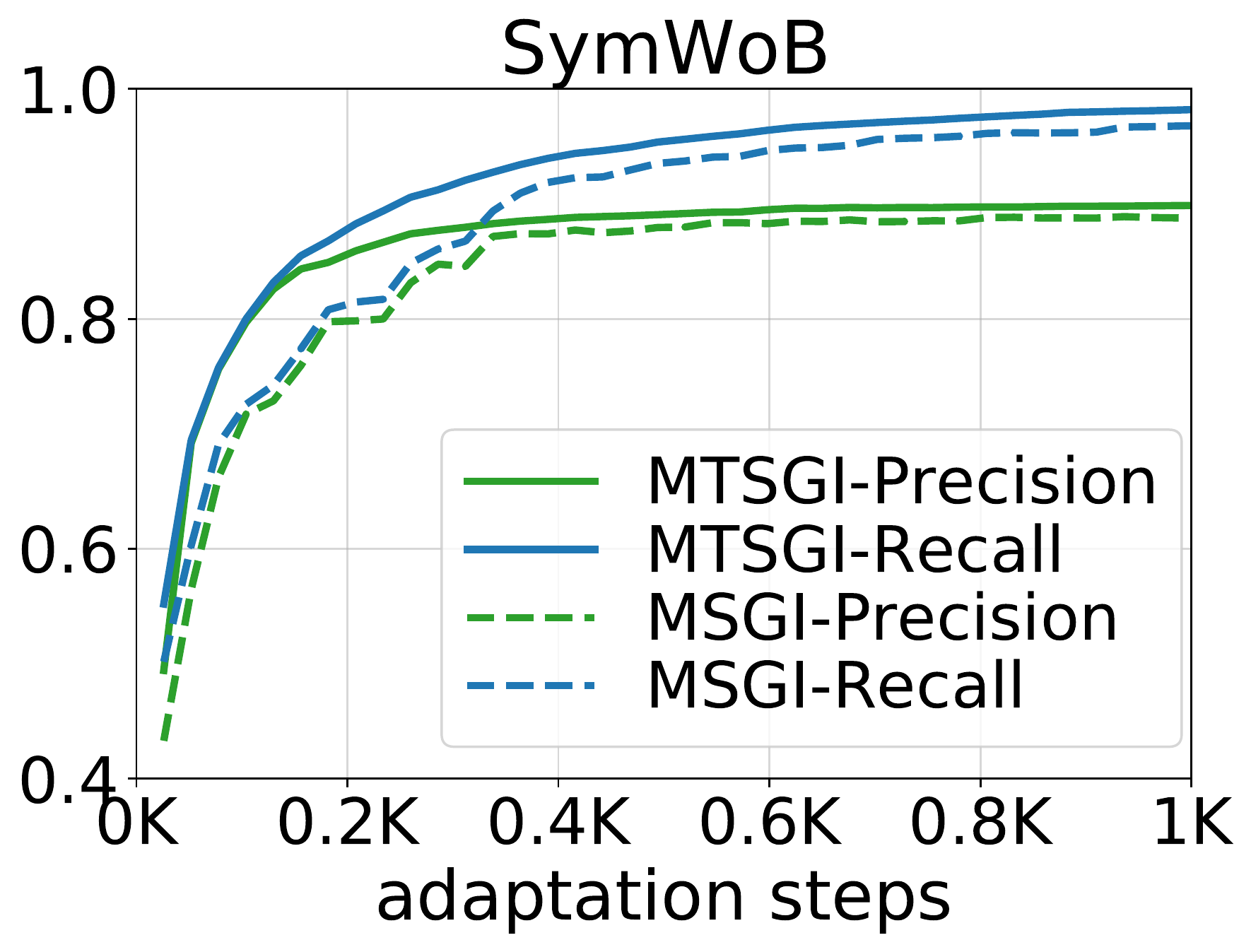}
    \includegraphics[draft=false,width=0.49\linewidth, valign=b]{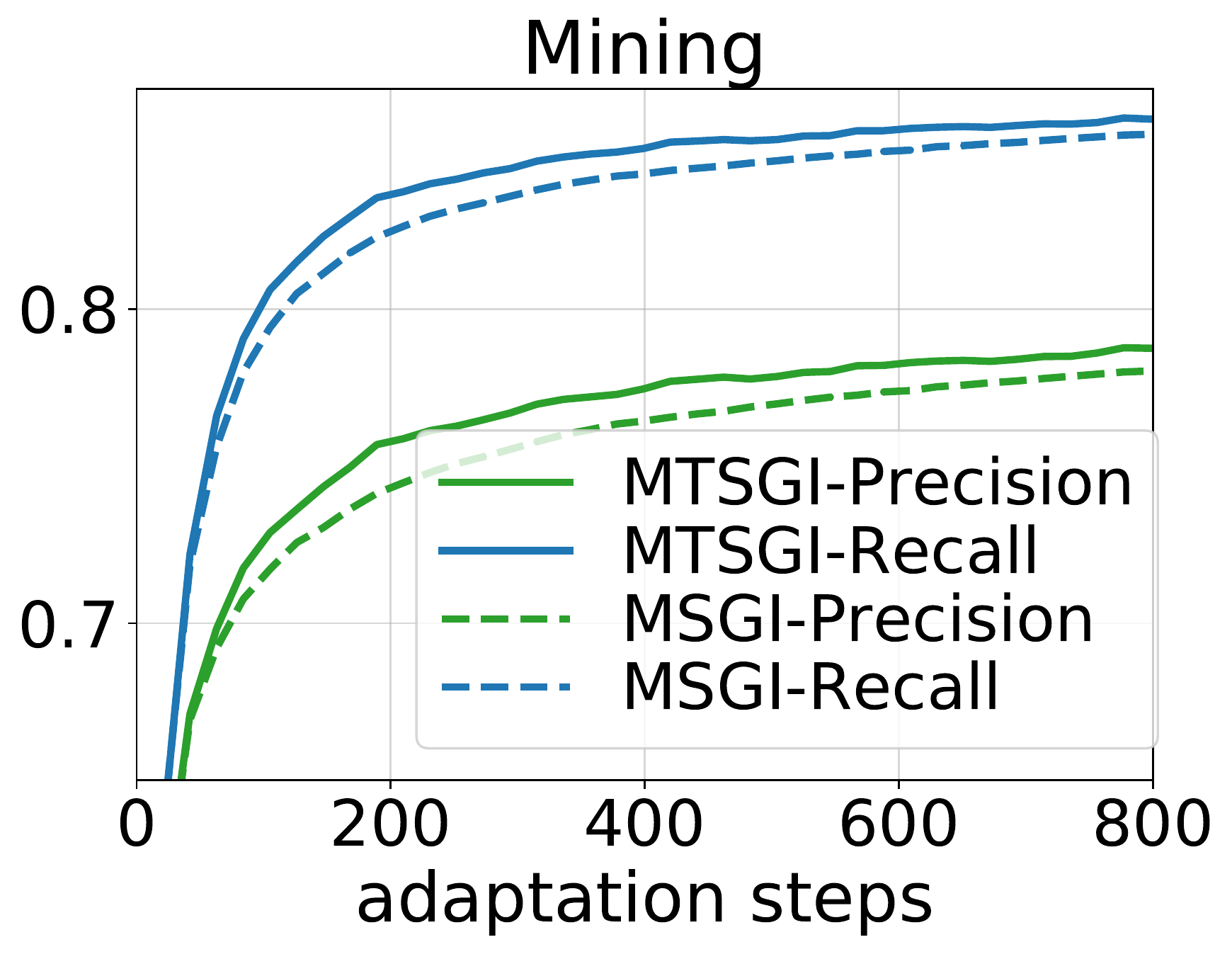}
    \vspace{-8pt}
    \caption{%
        The precision and recall of the subtask graphs inferred by MTSGI and MSGI on \wob and \mining.
    }
    \label{fig:precision_recall}
    \vspace{-7pt}
\end{figure}
\begin{figure*}[!t]
    \centering
    \includegraphics[draft=false,width=0.9\linewidth, valign=b]{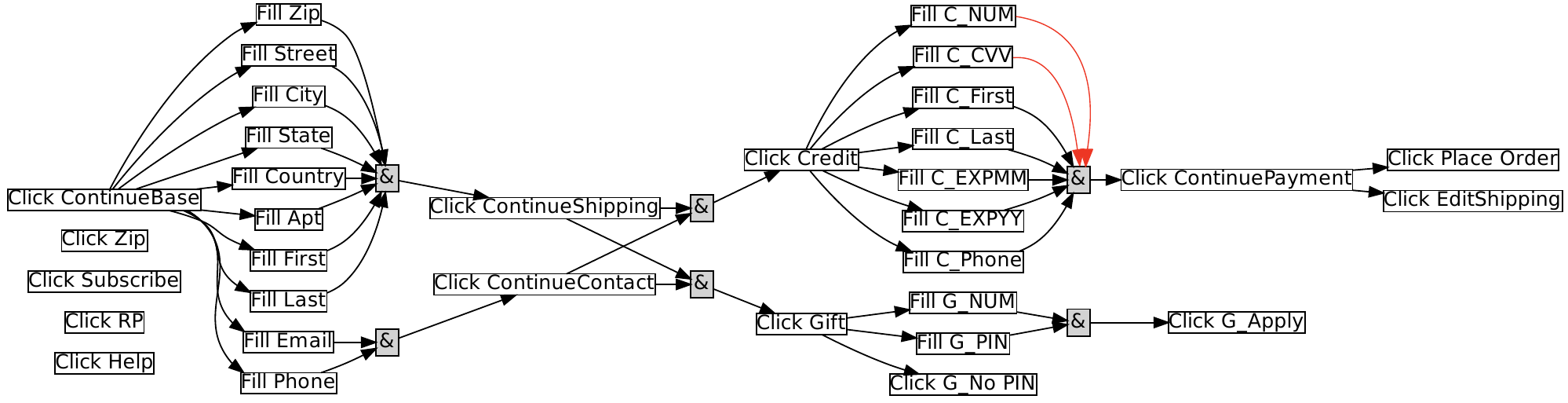}%
    \vspace{-8pt}
    \caption{%
        The visualization of the subtask graph inferred by our MTSGI after 1,000 steps of environment interaction on \walmart{} domain. Compared to the ground-truth subtask graph (not available to the agent), there was no error in the nodes and only two missing edges ({\color{red} in red}). See Appendix for the progression of the inferred subtask graph with varying adaptation steps.
    }
    \label{fig:inferred_graph_wob}
    \vspace{-7pt}
\end{figure*}
\begin{figure}[!t]
    \centering
    \includegraphics[draft=false,width=0.49\linewidth, valign=b]{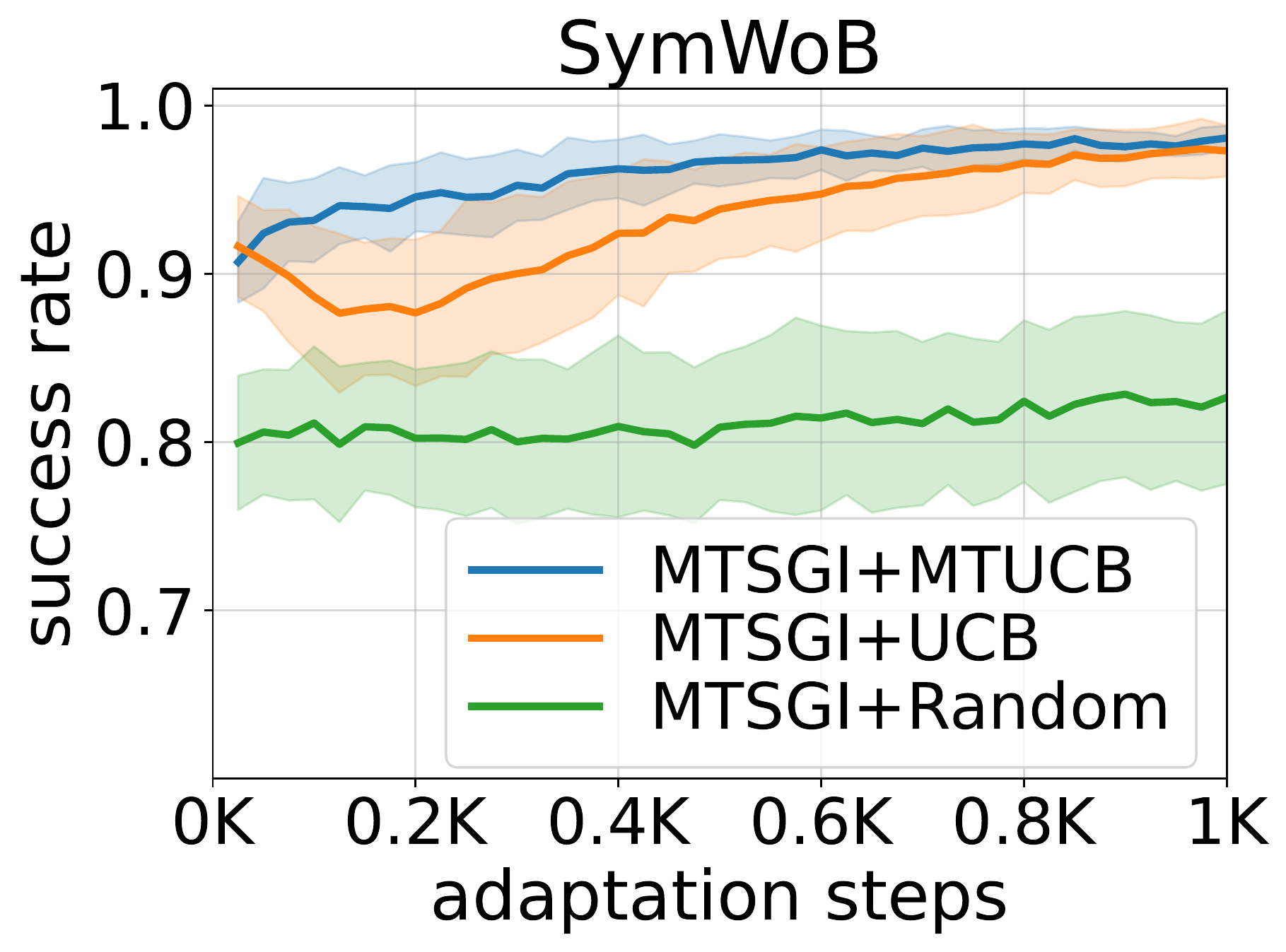}%
    \includegraphics[draft=false,width=0.49\linewidth, valign=b]{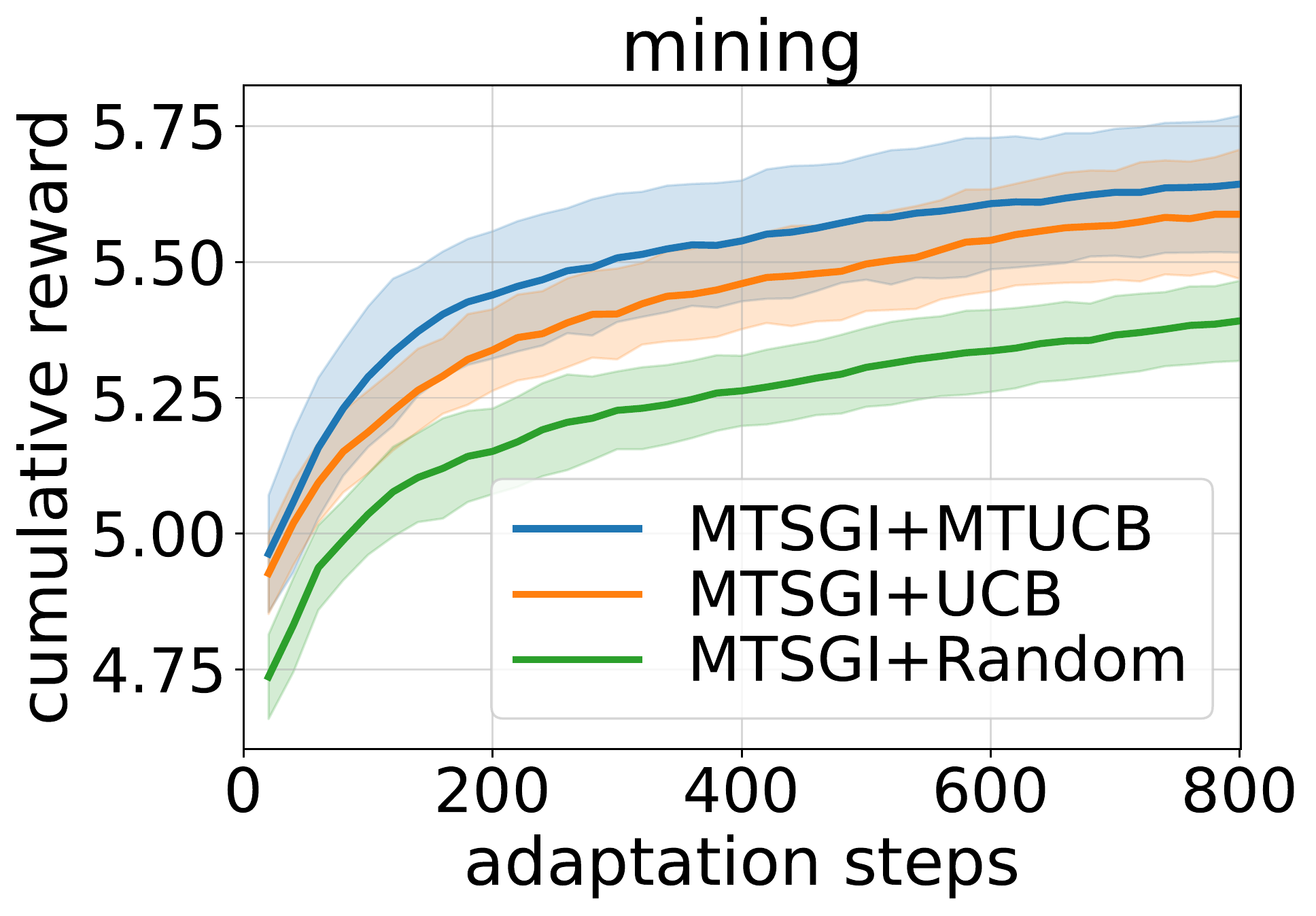}%
    \vspace{-8pt}
    \caption{%
        Comparison of different exploration strategies for MTSGI used in adaptation phase for \wob and \mining. 
    }
    \label{fig:ablation_exploration}
    \vspace{-7pt}
\end{figure}

\Cref{fig:fewshot_wob} and~\Cref{fig:fewshot_mining} show the few-shot generalization performance of the compared methods on \wob{} and \mining{}. 
In~\Cref{fig:fewshot_wob}, MTSGI achieves more than 75\% zero-shot success rate (\ie, success rate at x-axis=0) on all five tasks, which is significantly higher than the zero-shot performance of MSGI. 
This indicates that the prior learned from the training task significantly improves the subtask graph inference and in turn improves the multi-task evaluation policy.
Moreover, our MTSGI can learn a near-optimal policy on all the tasks after only 1,000 steps of environment interactions, demonstrating that the proposed multi-task learning scheme enables fast adaptation. 
Even though the MSGI agent is learning each task from scratch, it still outperforms the HRL and Random agents. This shows that explicitly inferring the underlying task structure and executing the predicted subtask graph
is significantly more effective than learning the policy from the reward signal (\ie, HRL) on complex compositional tasks.
Given the pre-learned options, HRL agent can slightly improve the success rate during the adaptation via PPO update. However, training the policy only from the sparse reward requires a large number of interactions especially for the tasks with many distractors (\eg, \expedia{} and \walgreens{}).
\subsection{Analysis on the Inferred Subtask Graph}
We compare the inferred subtask graph with the ground-truth subtask graph.
\Cref{fig:inferred_graph_wob} shows the subtask graph inferred by MTSGI in \walmart{}. We can see that MTSGI can accurately infer the subtask graph; the inferred subtask graph is missing only two preconditions (shown in red color) of \texttt{Click\_Continue\_Payment} subtask.
We note that such a small error in the subtask graph has a negligible effect as shown in~\Cref{fig:fewshot_wob}: \ie, MTSGI achieves near-optimal performance on \walmart{} after 1,000 steps of adaptation.~\Cref{fig:precision_recall} measures the precision and recall of the inferred precondition (\ie, the edge of the graph). First, both MTSGI and MSGI achieve high precision and recall after only a few hundred of adaptation. Also, MTSGI outperforms MSGI in the early stage of adaptation. This clearly demonstrates that the MTSGI can perform more accurate task inference due to the prior learned from the training tasks.

\begin{figure}[!t]
    \centering
    \includegraphics[draft=false,width=0.49\linewidth, valign=b]{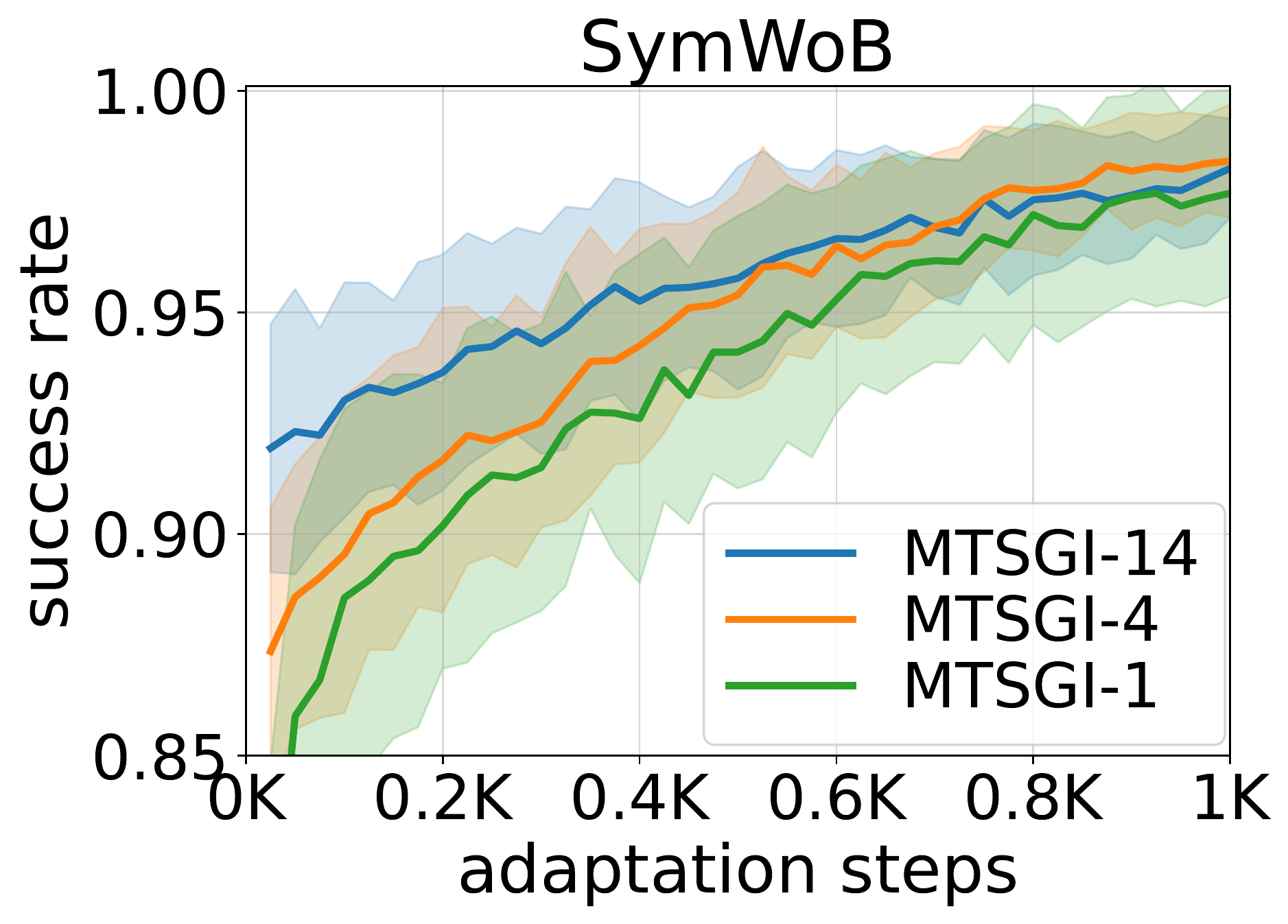}%
    \includegraphics[draft=false,width=0.49\linewidth, valign=b]{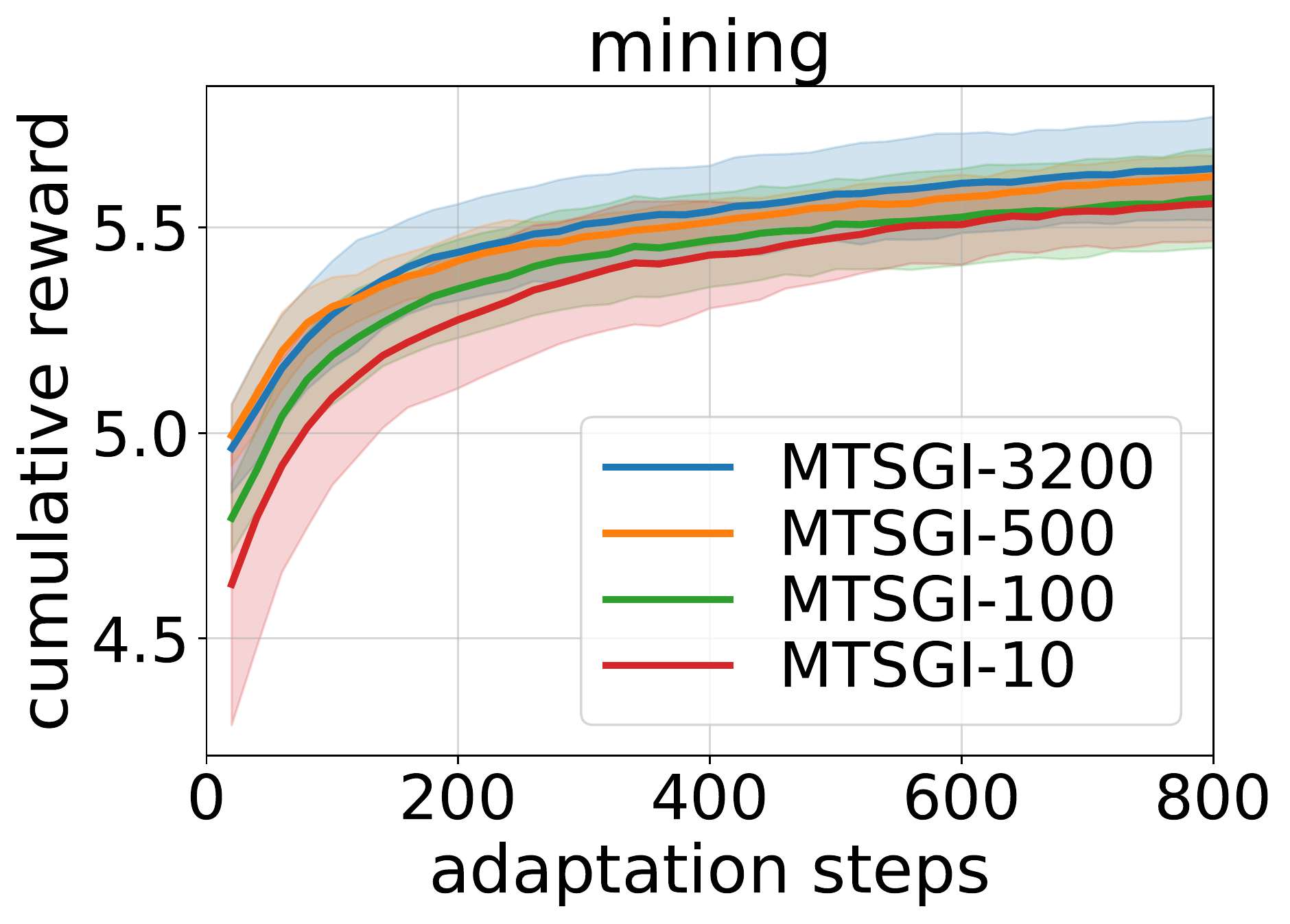}%
    \vspace{-8pt}
    \caption{%
       Comparison of different number of priors for MTSGI on \wob{} and \mining{}.
    }
    \label{fig:ablation_nprior}
    \vspace{-7pt}
\end{figure}

\newcommand{\rand}{MTSGI+Random\xspace}
\newcommand{\ucb}{MTSGI+UCB\xspace}
\newcommand{\mtucb}{MTSGI+MTUCB\xspace}
\subsection{Ablation Study: Effect of Exploration Strategy}
In this section, we investigate the effect of various exploration strategies on the performance of MTSGI. We compared the following three adaptation policies:
\begin{itemize}
\setlength{\itemsep}{1pt}\setlength{\parskip}{1pt}
    \item{Random}: A policy that uniformly randomly executes any eligible subtask. 
    \item{UCB}: The UCB policy defined in~\Cref{eq:UCB-policy} that aims to execute the novel subtask. The exploration parameters are initialized to zero when a new task is sampled.
    \item{MTUCB (Ours)}: Our multi-task extension of UCB policy. When a new task is sampled, the exploration parameter is initialized with those of the sampled prior.
\end{itemize}
\Cref{fig:ablation_exploration} summarizes the result on \wob{} and \mining{} domain, respectively. Using the more sophisticated exploration policy such as \ucb or \mtucb improved the performance of MTSGI compared to \rand, which was also observed in~\citet{sohn2019meta}. This is because better exploration helps the adaptation policy collect more data for logic induction by executing more diverse subtasks. In turn, this results in more accurate subtask graph inference and better performance.
Also, \mtucb outperforms \ucb on both domains. This indicates that transferring the exploration parameters makes the agent's exploration more efficient in meta-testing. Intuitively, the transferred exploration counts inform the agent which subtasks were \textit{under-explored} during meta-training, such that the agent can focus more on exploring those in meta-testing.

\subsection{Ablation Study: Effect of the prior set size}
MTSGI learns the prior from the training tasks. 
We investigated how many training tasks are required for MTSGI to learn a good prior for transfer learning.~\Cref{fig:ablation_nprior} compares the performance of MTSGI with the varying number of training tasks: 1, 4, 14 tasks for \wob and 10, 100, 500, 3200 tasks for \mining.
The training tasks are randomly subsampled from the entire training set.
The result shows that training on a larger number of tasks generally improves the performance.
\mining generally requires more number of training tasks than \wob because the agent is required to solve 440 different tasks in \mining while \wob was evaluated on 15 tasks; the agent is required to capture a wider range of task distribution in \mining than \wob.
Also, we note that MTSGI can still adapt much more efficiently than all other baseline methods even when only a small number of training tasks are available (\eg, one task for \wob and ten tasks for \mining).

\cutsectionup
\section{Conclusion}\label{sec:conclusion}
\cutsectiondown

We introduce a multi-task RL extension of the subtask graph inference framework that can quickly adapt to the unseen tasks by modeling the prior of subtask graph from the training tasks and transferring it to the test tasks.
The empirical results demonstrate that our MTSGI achieves strong zero-shot and few-shot generalization performance on 2D grid-world and complex web navigation domains by transferring the common knowledge learned in the training tasks to the unseen ones in terms of subtask graph.

In this work, we have assumed that the subtasks and the corresponding options are pre-learned and that the environment provides a high-level status of each subtask (\eg, whether the web element is filled in with the correct information).
In future work, our approach may be extended to a more general setting
where the relevant subtask structure is fully learned from (visual) observations,
and the corresponding options are autonomously discovered.

\bibliography{uai2022_mtsgi}

\begin{thebibliography}{43}
\providecommand{\natexlab}[1]{#1}
\providecommand{\url}[1]{\texttt{#1}}
\expandafter\ifx\csname urlstyle\endcsname\relax
  \providecommand{\doi}[1]{doi: #1}\else
  \providecommand{\doi}{doi: \begingroup \urlstyle{rm}\Url}\fi

\bibitem[Andreas et~al.(2017)Andreas, Klein, and Levine]{Andreas2017Modular}
Jacob Andreas, Dan Klein, and Sergey Levine.
\newblock Modular multitask reinforcement learning with policy sketches.
\newblock In \emph{ICML}, 2017.

\bibitem[Breiman(1984)]{breiman1984classification}
Leo Breiman.
\newblock \emph{Classification and regression trees}.
\newblock Routledge, 1984.

\bibitem[Chaplot et~al.(2018)Chaplot, Sathyendra, Pasumarthi, Rajagopal, and
  Salakhutdinov]{chaplot2018aaai}
Devendra~Singh Chaplot, Kanthashree~Mysore Sathyendra, Rama~Kumar Pasumarthi,
  Dheeraj Rajagopal, and Ruslan Salakhutdinov.
\newblock Gated-attention architectures for task-oriented language grounding.
\newblock In \emph{AAAI}, 2018.

\bibitem[Chen et~al.(2018)Chen, Badrinarayanan, Lee, and
  Rabinovich]{chen2018gradnorm}
Zhao Chen, Vijay Badrinarayanan, Chen-Yu Lee, and Andrew Rabinovich.
\newblock Gradnorm: Gradient normalization for adaptive loss balancing in deep
  multitask networks.
\newblock In \emph{International Conference on Machine Learning}, pages
  794--803. PMLR, 2018.

\bibitem[Duan et~al.(2016)Duan, Schulman, Chen, Bartlett, Sutskever, and
  Abbeel]{duan2016rl}
Yan Duan, John Schulman, Xi~Chen, Peter~L Bartlett, Ilya Sutskever, and Pieter
  Abbeel.
\newblock Rl $^2$: Fast reinforcement learning via slow reinforcement learning.
\newblock \emph{arXiv preprint arXiv:1611.02779}, 2016.

\bibitem[Finn et~al.(2017)Finn, Abbeel, and Levine]{finn2017model}
Chelsea Finn, Pieter Abbeel, and Sergey Levine.
\newblock Model-agnostic meta-learning for fast adaptation of deep networks.
\newblock In \emph{Proceedings of the 34th International Conference on Machine
  Learning-Volume 70}, pages 1126--1135. JMLR. org, 2017.

\bibitem[Finn et~al.(2018)Finn, Xu, and Levine]{Finn2018PMAML}
Chelsea Finn, Kelvin Xu, and Sergey Levine.
\newblock Probabilistic model-agnostic meta-learning.
\newblock In \emph{NeurIPS}, pages 9516--9527, 2018.

\bibitem[Ghazanfari and Taylor(2017)]{ghazanfari2017autonomous}
Behzad Ghazanfari and Matthew~E Taylor.
\newblock Autonomous extracting a hierarchical structure of tasks in
  reinforcement learning and multi-task reinforcement learning.
\newblock \emph{arXiv preprint arXiv:1709.04579}, 2017.

\bibitem[Gupta et~al.(2018)Gupta, Mendonca, Liu, Abbeel, and
  Levine]{Gupta:1802.07245}
Abhishek Gupta, Russell Mendonca, YuXuan Liu, Pieter Abbeel, and Sergey Levine.
\newblock Meta-reinforcement learning of structured exploration strategies.
\newblock \emph{arXiv preprint arXiv:1802.07245}, 2018.

\bibitem[Gur et~al.(2018)Gur, Rueckert, Faust, and
  Hakkani-Tur]{gur2018learning}
Izzeddin Gur, Ulrich Rueckert, Aleksandra Faust, and Dilek Hakkani-Tur.
\newblock Learning to navigate the web.
\newblock \emph{arXiv preprint arXiv:1812.09195}, 2018.

\bibitem[Hausman et~al.(2018)Hausman, Springenberg, Wang, Heess, and
  Riedmiller]{hausman2018learning}
Karol Hausman, Jost~Tobias Springenberg, Ziyu Wang, Nicolas Heess, and Martin
  Riedmiller.
\newblock Learning an embedding space for transferable robot skills.
\newblock In \emph{International Conference on Learning Representations}, 2018.

\bibitem[Hochreiter et~al.(2001)Hochreiter, Younger, and
  Conwell]{hochreiter2001learning}
Sepp Hochreiter, A~Steven Younger, and Peter~R Conwell.
\newblock Learning to learn using gradient descent.
\newblock In \emph{International Conference on Artificial Neural Networks},
  pages 87--94. Springer, 2001.

\bibitem[Hoffman et~al.(2020)Hoffman, Shahriari, Aslanides, Barth-Maron,
  Behbahani, Norman, Abdolmaleki, Cassirer, Yang, Baumli, Henderson, Novikov,
  Colmenarejo, Cabi, Gulcehre, Paine, Cowie, Wang, Piot, and
  de~Freitas]{hoffman2020acme}
Matt Hoffman, Bobak Shahriari, John Aslanides, Gabriel Barth-Maron, Feryal
  Behbahani, Tamara Norman, Abbas Abdolmaleki, Albin Cassirer, Fan Yang, Kate
  Baumli, Sarah Henderson, Alex Novikov, Sergio~Gómez Colmenarejo, Serkan
  Cabi, Caglar Gulcehre, Tom~Le Paine, Andrew Cowie, Ziyu Wang, Bilal Piot, and
  Nando de~Freitas.
\newblock Acme: A research framework for distributed reinforcement learning.
\newblock \emph{arXiv preprint arXiv:2006.00979}, 2020.
\newblock URL \url{https://arxiv.org/abs/2006.00979}.

\bibitem[Huang et~al.(2018)Huang, Nair, Xu, Zhu, Garg, Fei-Fei, Savarese, and
  Niebles]{huang2018neural}
De-An Huang, Suraj Nair, Danfei Xu, Yuke Zhu, Animesh Garg, Li~Fei-Fei, Silvio
  Savarese, and Juan~Carlos Niebles.
\newblock Neural task graphs: Generalizing to unseen tasks from a single video
  demonstration.
\newblock \emph{arXiv preprint arXiv:1807.03480}, 2018.

\bibitem[Jia et~al.(2019)Jia, Kiros, and Ba]{jia2018domqnet}
Sheng Jia, Jamie~Ryan Kiros, and Jimmy Ba.
\newblock {DOM}-q-{NET}: Grounded {RL} on structured language.
\newblock In \emph{International Conference on Learning Representations}, 2019.
\newblock URL \url{https://openreview.net/forum?id=HJgd1nAqFX}.

\bibitem[Kim et~al.(2018)Kim, Yoon, Dia, Kim, Bengio, and Ahn]{Kim:2018:BMAML}
Taesup Kim, Jaesik Yoon, Ousmane Dia, Sungwoong Kim, Yoshua Bengio, and Sungjin
  Ahn.
\newblock Bayesian model-agnostic meta-learning.
\newblock \emph{arXiv preprint arXiv:1806.03836}, 2018.

\bibitem[Konidaris and Barto(2006)]{konidaris2006autonomous}
George Konidaris and Andrew Barto.
\newblock Autonomous shaping: Knowledge transfer in reinforcement learning.
\newblock In \emph{Proceedings of the 23rd international conference on Machine
  learning}, pages 489--496, 2006.

\bibitem[Lazaric(2012)]{lazaric2012transfer}
Alessandro Lazaric.
\newblock Transfer in reinforcement learning: a framework and a survey.
\newblock In \emph{Reinforcement Learning}, pages 143--173. Springer, 2012.

\bibitem[Lazaric et~al.(2008)Lazaric, Restelli, and
  Bonarini]{lazaric2008transfer}
Alessandro Lazaric, Marcello Restelli, and Andrea Bonarini.
\newblock Transfer of samples in batch reinforcement learning.
\newblock In \emph{Proceedings of the 25th international conference on Machine
  learning}, pages 544--551, 2008.

\bibitem[Lin et~al.(2019)Lin, Baweja, Kantor, and Held]{lin2019adaptive}
Xingyu Lin, Harjatin~Singh Baweja, George Kantor, and David Held.
\newblock Adaptive auxiliary task weighting for reinforcement learning.
\newblock \emph{Advances in neural information processing systems}, 32, 2019.

\bibitem[Liu et~al.(2016)Liu, Yang, Iaba-Sadiya, Shukla, He, Zhu, and
  Chai]{liu2016GTSvisual}
Changsong Liu, Shaohua Yang, Sari Iaba-Sadiya, Nishant Shukla, Yunzhong He,
  Song-chun Zhu, and Joyce Chai.
\newblock Jointly learning grounded task structures from language instruction
  and visual demonstration.
\newblock In \emph{EMNLP}, 2016.

\bibitem[Liu et~al.(2018)Liu, Guu, Pasupat, Shi, and
  Liang]{liu2018reinforcement}
Evan~Zheran Liu, Kelvin Guu, Panupong Pasupat, Tianlin Shi, and Percy Liang.
\newblock Reinforcement learning on web interfaces using workflow-guided
  exploration.
\newblock \emph{arXiv preprint arXiv:1802.08802}, 2018.

\bibitem[Mnih et~al.(2015)Mnih, Kavukcuoglu, Silver, Rusu, Veness, Bellemare,
  Graves, Riedmiller, Fidjeland, Ostrovski, et~al.]{mnih2015human}
Volodymyr Mnih, Koray Kavukcuoglu, David Silver, Andrei~A Rusu, Joel Veness,
  Marc~G Bellemare, Alex Graves, Martin Riedmiller, Andreas~K Fidjeland, Georg
  Ostrovski, et~al.
\newblock Human-level control through deep reinforcement learning.
\newblock \emph{Nature}, 518\penalty0 (7540):\penalty0 529--533, 2015.

\bibitem[Muggleton(1991)]{Muggleton:1991:ILP}
Stephen Muggleton.
\newblock Inductive logic programming.
\newblock \emph{New Gen. Comput.}, 8\penalty0 (4):\penalty0 295--318, February
  1991.
\newblock ISSN 0288-3635.
\newblock \doi{10.1007/BF03037089}.
\newblock URL \url{http://dx.doi.org/10.1007/BF03037089}.

\bibitem[Nichol et~al.(2018)Nichol, Achiam, and Schulman]{nichol2018first}
Alex Nichol, Joshua Achiam, and John Schulman.
\newblock On first-order meta-learning algorithms.
\newblock \emph{arXiv preprint arXiv:1803.02999}, 2018.

\bibitem[Oh et~al.(2017)Oh, Singh, Lee, and Kohli]{oh2017zero}
Junhyuk Oh, Satinder Singh, Honglak Lee, and Pushmeet Kohli.
\newblock Zero-shot task generalization with multi-task deep reinforcement
  learning.
\newblock In \emph{ICML}, 2017.

\bibitem[Pinto and Gupta(2017)]{pinto2017learning}
Lerrel Pinto and Abhinav Gupta.
\newblock Learning to push by grasping: Using multiple tasks for effective
  learning.
\newblock In \emph{2017 IEEE international conference on robotics and
  automation (ICRA)}, pages 2161--2168. IEEE, 2017.

\bibitem[Sacerdoti(1975{\natexlab{a}})]{sacerdoti1975nonlinear}
Earl~D Sacerdoti.
\newblock The nonlinear nature of plans.
\newblock Technical report, STANFORD RESEARCH INST MENLO PARK CA,
  1975{\natexlab{a}}.

\bibitem[Sacerdoti(1975{\natexlab{b}})]{sacerdoti1975structure}
Earl~D Sacerdoti.
\newblock A structure for plans and behavior.
\newblock Technical report, SRI INTERNATIONAL MENLO PARK CA ARTIFICIAL
  INTELLIGENCE CENTER, 1975{\natexlab{b}}.

\bibitem[Schulman et~al.(2017)Schulman, Wolski, Dhariwal, Radford, and
  Klimov]{schulman2017proximal}
John Schulman, Filip Wolski, Prafulla Dhariwal, Alec Radford, and Oleg Klimov.
\newblock Proximal policy optimization algorithms.
\newblock \emph{arXiv preprint arXiv:1707.06347}, 2017.

\bibitem[Shi et~al.(2017)Shi, Karpathy, Fan, Hernandez, and
  Liang]{shi2017world}
Tianlin Shi, Andrej Karpathy, Linxi Fan, Jonathan Hernandez, and Percy Liang.
\newblock World of bits: An open-domain platform for web-based agents.
\newblock In \emph{International Conference on Machine Learning}, pages
  3135--3144, 2017.

\bibitem[Silver et~al.(2017)Silver, Schrittwieser, Simonyan, Antonoglou, Huang,
  Guez, Hubert, Baker, Lai, Bolton, et~al.]{silver2017mastering}
David Silver, Julian Schrittwieser, Karen Simonyan, Ioannis Antonoglou, Aja
  Huang, Arthur Guez, Thomas Hubert, Lucas Baker, Matthew Lai, Adrian Bolton,
  et~al.
\newblock Mastering the game of go without human knowledge.
\newblock \emph{nature}, 550\penalty0 (7676):\penalty0 354--359, 2017.

\bibitem[Sohn et~al.(2018)Sohn, Oh, and Lee]{sohn2018hierarchical}
Sungryull Sohn, Junhyuk Oh, and Honglak Lee.
\newblock Hierarchical reinforcement learning for zero-shot generalization with
  subtask dependencies.
\newblock In \emph{NeurIPS}, pages 7156--7166, 2018.

\bibitem[Sohn et~al.(2019)Sohn, Woo, Choi, and Lee]{sohn2019meta}
Sungryull Sohn, Hyunjae Woo, Jongwook Choi, and Honglak Lee.
\newblock Meta reinforcement learning with autonomous inference of subtask
  dependencies.
\newblock In \emph{International Conference on Learning Representations}, 2019.

\bibitem[Sutton et~al.(1999)Sutton, Precup, and Singh]{sutton1999between}
Richard~S Sutton, Doina Precup, and Satinder Singh.
\newblock Between mdps and semi-mdps: A framework for temporal abstraction in
  reinforcement learning.
\newblock \emph{Artificial intelligence}, 112\penalty0 (1-2):\penalty0
  181--211, 1999.

\bibitem[Tate(1977)]{tate1977generating}
Austin Tate.
\newblock Generating project networks.
\newblock In \emph{Proceedings of the 5th international joint conference on
  Artificial intelligence-Volume 2}, pages 888--893. Morgan Kaufmann Publishers
  Inc., 1977.

\bibitem[Taylor and Stone(2009)]{taylor2009transfer}
Matthew~E Taylor and Peter Stone.
\newblock Transfer learning for reinforcement learning domains: A survey.
\newblock \emph{Journal of Machine Learning Research}, 10\penalty0 (7), 2009.

\bibitem[Vinyals et~al.(2019)Vinyals, Babuschkin, Czarnecki, Mathieu, Dudzik,
  Chung, Choi, Powell, Ewalds, Georgiev, et~al.]{vinyals2019grandmaster}
Oriol Vinyals, Igor Babuschkin, Wojciech~M Czarnecki, Micha{\"e}l Mathieu,
  Andrew Dudzik, Junyoung Chung, David~H Choi, Richard Powell, Timo Ewalds,
  Petko Georgiev, et~al.
\newblock Grandmaster level in starcraft ii using multi-agent reinforcement
  learning.
\newblock \emph{Nature}, 575\penalty0 (7782):\penalty0 350--354, 2019.

\bibitem[Wang et~al.(2016)Wang, Kurth-Nelson, Tirumala, Soyer, Leibo, Munos,
  Blundell, Kumaran, and Botvinick]{wang2016learning}
Jane~X Wang, Zeb Kurth-Nelson, Dhruva Tirumala, Hubert Soyer, Joel~Z Leibo,
  Remi Munos, Charles Blundell, Dharshan Kumaran, and Matt Botvinick.
\newblock Learning to reinforcement learn.
\newblock \emph{arXiv preprint arXiv:1611.05763}, 2016.

\bibitem[Wilson et~al.(2007)Wilson, Fern, Ray, and Tadepalli]{wilson2007multi}
Aaron Wilson, Alan Fern, Soumya Ray, and Prasad Tadepalli.
\newblock Multi-task reinforcement learning: a hierarchical bayesian approach.
\newblock In \emph{Proceedings of the 24th international conference on Machine
  learning}, pages 1015--1022, 2007.

\bibitem[Xu et~al.(2017)Xu, Nair, Zhu, Gao, Garg, Fei-Fei, and Savarese]{NTP}
Danfei Xu, Suraj Nair, Yuke Zhu, Julian Gao, Animesh Garg, Li~Fei-Fei, and
  Silvio Savarese.
\newblock Neural task programming: Learning to generalize across hierarchical
  tasks.
\newblock \emph{arXiv preprint arXiv:1710.01813}, 2017.

\bibitem[Yu et~al.(2017)Yu, Zhang, and Xu]{yu2017deep}
Haonan Yu, Haichao Zhang, and Wei Xu.
\newblock A deep compositional framework for human-like language acquisition in
  virtual environment.
\newblock \emph{arXiv preprint arXiv:1703.09831}, 2017.

\bibitem[Zhang and Yeung(2014)]{zhang2014regularization}
Yu~Zhang and Dit-Yan Yeung.
\newblock A regularization approach to learning task relationships in multitask
  learning.
\newblock \emph{ACM Transactions on Knowledge Discovery from Data (TKDD)},
  8\penalty0 (3):\penalty0 1--31, 2014.

\end{thebibliography}
\clearpage
\onecolumn
\appendix

    \begin{center}
    {\Large \textbf{ Appendix: 
    Fast Inference and Transfer of Compositional Task Structures \\for Few-shot Task Generalization
    \\[10pt]
    }}
    \end{center}
    \section{Derivation of subtask graph inference}\label{app:sgi}
Following~\citet{sohn2019meta}, we formulate the problem of inferring the precondition $\GC{}$ and the subtask reward $\GR{}$ as a maximum likelihood estimation (MLE) problem.
Let $\tau_{K}=\{ \mb{s}_1, \mb{o}_1, r_1, d_1, \ldots, \mb{s}_K\}$
be an adaptation trajectory of the adaptation policy $\pi^\text{adapt}_{\theta}$
for $K$ steps in adaptation phase.
The goal is to infer the subtask graph $G$ for this task,
specified by preconditions $\GC{}$ and subtask rewards $\GR{}$.
Consider a subtask graph $G$ with subtask reward $\GR{}$ and precondition $\GC{}$.
The maximum-likelihood estimate (MLE) of latent variables $G$, conditioned on the trajectory $\tau_{H}$ can be written as
\begin{align}
    \smash{\widehat{G}}^{\text{MLE}} = \argmax_{ \GC{}, \GR{} } p(\tau_{K}| \GC{}, \GR{} ).  \label{eq:ILP_objective}
\end{align}
The likelihood term can be expanded as
\begin{align}
    p(\tau_{K}|\GC{}, \GR{})  
    &=p(\mb{s}_1|\GC{}) \prod_{t=1}^{K} \pi_{\theta}\left( \mb{o}_t|\tau_{t} \right) p(\mb{s}_{t+1}|\mb{s}_t,\mb{o}_t,\GC{}) p(r_t | \mb{s}_t,\mb{o}_t,\GR{}) p(d_t|\mb{s}_t,\mb{o}_t) \label{eq:ll1}
    \\[-2pt]
    &\propto p(\mb{s}_1|\GC{}) \prod_{t=1}^{K}{ p(\mb{s}_{t+1}|\mb{s}_t,\mb{o}_t,\GC{}) p(r_t | \mb{s}_t,\mb{o}_t,\GR{}) },\label{eq:ll2}
\end{align}
where we dropped the terms that are independent of $G$. From the definitions in \tb{Subtask Graph Inference Problem} section, precondition $\GC{}$ defines the mapping $\mb{x}\mapsto \mb{e}$, and the subtask reward $\GR{}$ determines the reward as $r_t\sim \GR{}^i$ if subtask $i$ is eligible (\ie, $\mb{e}_t^i=1$) and option $\mc{O}^i$ is executed at time $t$.
Therefore, we have
\begin{align}
\widehat{G}^{\text{MLE}} = (\widehat{G}_\mb{c}^\text{MLE}, \widehat{G}_\mb{r}^\text{MLE}) = \left(\argmax_{\GC{}} \prod_{t=1}^{K}{ p(\mb{e}_{t}|\mb{x}_{t}, \GC{}) },\ \argmax_{\GR{}} \prod_{t=1}^{K}{ p(r_t|\mb{e}_t,\mb{o}_t, \GR{})}\right).\label{eq:graph-infer}
\end{align}
Below we explain how to compute the estimate of preconditions $\smash{\widehat{G}_\mb{c}^\text{MLE}}$ and subtask rewards $\smash{\widehat{G}_\mb{r}^\text{MLE}}$.

\cutparagraphup
\paragraph{Precondition inference via logic induction}
From the definition in \tb{Subtask Graph Inference Problem} section, the mapping from a precondition $\GC{}$ and a completion vector $\mb{x}$ to an eligibility vector $\mb{e} = f_{\GC{}}(\mb{x})$ is a deterministic function (\ie, precondition function).
%
Therefore, we can rewrite $\widehat{G}_\mb{c}^\text{MLE}$ in Eq.(\ref{eq:graph-infer}) as:
\begin{align}
\widehat{G}_\mb{c}^\text{MLE}
&= \argmax_{\GC{}} \prod_{t=1}^{K}{ \mbb{I}( \mb{e}_t = f_{\GC{}}(\mb{x}_{t}) ) },\label{eq:cond-infer2}
\end{align}
where $\mbb{I}(\cdot)$ is the indicator function. 
Since the eligibility $\mb{e}$ is factored, the precondition function $f_{\GC{}^i}$ for each subtask is inferred independently.
This problem of finding a boolean function that satisfies all the indicator functions in Eq.(\ref{eq:cond-infer2})
(\ie, $\smash{\prod_{t=1}^{K}{ \mbb{I}( \mb{e}_t = f_{\GC{}}(\mb{x}_{t}) ) }=1}$) is formulated as an \textit{inductive logic programming} (ILP) problem~\citep{Muggleton:1991:ILP}.
Specifically, $\{\mb{x}_{t}\}^{K}_{t=1}$ forms binary vector inputs to programs,
and $\{e^{i}_{t}\}^{K}_{t=1}$ forms Boolean-valued outputs of the $i$-th program that denotes the eligibility of the $i$-th subtask.
We use the \textit{classification and regression tree} (CART) to infer the precondition function $f_{\GC{}}$ for each subtask based on Gini impurity~\citep{breiman1984classification}. Intuitively, the constructed decision tree is the simplest boolean function approximation for the given input-output pairs $\{\mb{x}_t, \mb{e}_t\}$.
Then, we convert it to a logic expression (\ie, precondition) in sum-of-product (SOP) form to build the subtask graph.

\cutparagraphup
\cutparagraphup
\paragraph{Subtask reward inference}

To infer the subtask reward function $\widehat{G}_\mb{r}^{\text{MLE}}$ in Eq.(\ref{eq:graph-infer}),
we model each component of subtask reward as a Gaussian distribution $G_\mb{r}^{i} \sim \mathcal{N}(\widehat{\mu}^{i}, \widehat{\sigma}^{i} )$.
Then, $\widehat{\mu}^{i}_{\text{MLE}}$ becomes the empirical mean of the rewards received after taking the eligible option $\mc{O}^i$ in the trajectory $\tau_{K}$:
\begin{align}
    \widehat{G}_\mb{r}^{\text{MLE},i}
    = \widehat{\mu}^{i}_{\text{MLE}}
    = \mathbb E \left[ r_t|\mb{o}_t=\mc{O}^i, \mb{e}_t^i=1 \right]=\frac{ \sum^{K}_{t=1}{ r_t \mbb{I}(\mb{o}_t=\mc{O}^i, \mb{e}_t^i=1) } }{ \sum^{K}_{t=1}{ \mbb{I}(\mb{o}_t=\mc{O}^i, \mb{e}_t^i=1) } }.
\end{align}
    \newpage
    \section{Details of GRProp policy}
\label{sec:appendix_grprop}

For self-containedness, we repeat the description of GRProp policy in \citet{sohn2019meta}.

Intuitively, GRProp policy modifies the subtask graph to a differentiable form such that we can compute the gradient of modified return with respect to the subtask completion vector in order to measure how much each subtask is likely to increase the modified return.
 Let $\mb{x}_t$ be a completion vector and $\GR{}$ be a subtask reward vector (see \tb{Subtask Graph Inference Problem} section for definitions). Then, the sum of reward until time-step $t$ is given as:
 \begin{align}
    U_t &= \GR{}^{\top} \mb{x}_{t}. \label{eq:cr}
 \end{align}
We first modify the reward formulation such that it gives a half of subtask reward for satisfying the preconditions and the rest for executing the subtask to encourage the agent to satisfy the precondition of a subtask with a large reward:
 \begin{align}
    \widehat{U}_t &= \GR{}^{\top} (\mb{x}_{t}+\mb{e}_t)/2. \label{eq:cr2}
 \end{align}
Let $y_{AND}^j$ be the output of $j$-th \texttt{AND} node. The eligibility vector $\mathbf{e}_t$ can be computed from the subtask graph $G$ and $\mb{x}_t$ as follows:
\begin{align}
  e_{t}^{i} = \underset{j\in Child_i}{\text{OR}} \left( y^{j}_{AND}\right),\quad
  y^{j}_{AND} = \underset{k\in Child_j}{\text{AND}} \left( \widehat{x}_{t}^{j,k}\right),\quad
  \widehat{x}_{t}^{j,k}= x_t^k w^{j,k} + \text{NOT}(x_t^k)(1-w^{j,k}),
  \label{eq:xhat}
\end{align}
where $w^{j, k}=0$ if there is a \texttt{NOT} connection between $j$-th node and $k$-th node, otherwise $w^{j,k}=1$. Intuitively, $\widehat{x}_{t}^{j,k}=1$ when $k$-th node does not violate the precondition of $j$-th node.
The logical AND, OR, and NOT operations in ~\Cref{eq:xhat} 
are substituted by the smoothed counterparts as follows:
  \begin{align}
 p^{i}&= \lambda_{\text{or}} \wt{e}^{i} + \left( 1 - \lambda_{\text{or}}\right) x^i,\label{eq:soft-progress}\\
 \wt{e}^{i} &= \underset{j\in Child_i}{\wt{\text{OR}}} \left( \wt{y}^{j}_{AND}\right),\\
 \wt{y}^{j}_{AND} &= \underset{k\in Child_j}{\wt{\text{AND}}} \left( \hat{x}^{j,k}\right),\\
 \hat{x}^{j,k}&= w^{j,k} p^k + (1-w^{j,k}) \wt{\text{NOT}} \left(p^k \right),
\end{align}
where $\mb{x}\in\mbb{R}^{d}$ is the input completion vector,
\begin{align}
\wt{\text{OR}} \left( \mb{x} \right) &= \softmax(w_{\text{or}}\mb{x})\cdot \mb{x}, \label{eq:soft-or}\\
\wt{\text{AND}} \left( \mb{x} \right) &= \frac{\softplus(\mb{x}, w_{\text{and}})}{\softplus(||\mb{x}||, w_{\text{and}})} ,\label{eq:soft-and}\\
\wt{\text{NOT}} \left( \mb{x} \right) &= -w_{\text{not}} \mb{x},
\end{align}
$||\mb{x}||=d$, $\softplus(\mb{x}, \beta)=\frac{1}{\beta} \log(1+\exp(\beta \mb{x}))$ is a soft-plus function, and $\lambda_{\text{or}}=0.6, w_{\text{or}}=2, w_{\text{and}}=3, w_{\text{not}}=2$ are the hyper-parameters of GRProp. 
Note that we slightly modified the implementation of $\wt{\text{OR}}$ and $\wt{\text{AND}}$ from sigmoid and hyper-tangent functions in~\citep{sohn2018hierarchical} to softmax and softplus functions for better performance. With the smoothed operations, the sum of smoothed and modified reward is given as:
\begin{align}
    \widetilde{U}_t &= \GR{}^\top \mb{p}, \label{eq:cr4}
\end{align}
where $\mb{p}=[p^1,\ldots,p^d]$ and $p^i$ is computed from ~\Cref{eq:soft-progress}.
Finally, the graph reward propagation policy is a softmax policy,  
\begin{align}
 \pi(\mb{o}_{t}|G,\mb{x}_t) =  \text{Softmax}\left(  T\nabla_{\mb{x}_{t}} \widetilde{U}_{t} \right)
 =\text{Softmax}\left( T\GR{}^\top (\lambda_{\text{or}}\nabla_{\mb{x}_{t}} \wt{\mb{e}}_{t}+ (1-\lambda_{\text{or}}) ) \right),
\end{align}
where we used the softmax temperature $T=40$ for \tb{Playground} and \tb{Mining} domain,
and linearly annealed the temperature from $T=1$ to $T=40$ during adaptation phase for \tb{SC2LE} domain.
Intuitively speaking, we act more confidently (\ie, higher temperature $T$) as we collect more data since the inferred subtask graph will become more accurate.
    \newpage
    \section{Derivation of Multi-task subtask graph inference}\label{app:mtsgi}
Let $\tau$ be the adaptation trajectory of the current task $\mc{M}_G$, and $\mc{T}^{\text{p}}=\{\tau^{\text{p}}_1, \ldots, \tau^{\text{p}}_{|\mc{T}^{\text{p}}|}\}$ be the adaptation trajectories of the seen training tasks. 

 Then, from Bayesian rule, we have
\begin{align}
\pi(o|s,\tau, \mc{T}^{\text{p}})
&=\textstyle \sum_G \pi(o|s,G)p(G|\tau, \mc{T}^{\text{p}})\\
&\propto \textstyle \sum_G \pi(o|s,G)p(\tau|G, \mc{T}^{\text{p}})p(G|\mc{T}^{\text{p}})\\
&= \textstyle \sum_G \pi(o|s,G)p(\tau|G)p(G|\mc{T}^{\text{p}})\label{eq:omit_tau}\\
&\propto \textstyle \sum_G \pi(o|s,G)p(\tau|G)p(\mc{T}^{\text{p}}|G)p(G)\label{eq:ell},
\end{align}
where~\Cref{eq:omit_tau} holds because $\tau$ and $\mc{T}^{\text{p}}$ are independently observed variables. 
Since summing over all $G$ is computationally intractable, we instead approximate it by computing the sample estimates of $\pi$. Specifically, we compute the policy $ \pi(o|s,G)$ at the maximum likelihood estimates (MLE) of $G$ for prior and current tasks, that is $G^\tau=\argmax_{G}p(\tau|G)$ and $G^\text{p}=  \argmax_{G}p(\mc{T}^{\text{p}}|G)$ respectively,  and combine them with the weight $\alpha$:
\begin{align}
\pi(o|s,\tau, \mc{T}^{\text{p}})
\simeq \pi(o|s, G^{\tau})^\alpha\pi(o|s, G^\text{p})^{(1-\alpha)}.
\end{align}
Finally, we deploy the GRProp policy as a contextual policy:
\begin{align}
\pieval(\cdot| \tau, \mc{T}^{\text{p}})
\simeq \text{GRProp}(\cdot| G^{\tau})^\alpha\text{GRProp}(\cdot|G^\text{p})^{(1-\alpha)}.
\end{align}

    \section{Pseudo-code of our algorithm}\label{app:algo}
The~\Cref{alg:meta-train-app} below describes the pseudo-code of the meta-training process of our algorithm.
\begin{algorithm}[H]
\caption{Meta-training: learning the prior}\label{alg:meta-train-app}
\begin{algorithmic}[1]
\REQUIRE{ Adaptation policy $\piadapt$}
\ENSURE{ Prior set $\mc{T}^{\text{p}}$}
\STATE $\mc{T}^{\text{p}} \gets \emptyset$
\FOR{each task $\mc{M}\in\traintask$}
    \STATE Rollout adaptation policy: \\
    $\tau=\{\mb{s}_t, \mb{o}_t,r_t, d_t\}_{t=1}^{K} \sim \piadapt$ in task $\mc{M}$
    \STATE Infer subtask graph $G^\tau = \argmax_{G} p(\tau|G)$ 
    \STATE $\pieval= \text{GRProp}(G^\tau)$
    \STATE Evaluate the agent: $\tau^\text{eval}\sim \pieval$ in task $\mc{M}$
    \STATE Update prior $\mc{T}^{\text{p}}\gets \mc{T}^{\text{p}} \cup \left( G^\tau,  \tau\right)$
\ENDFOR
\end{algorithmic}
\end{algorithm}
    \newpage
    \cutsectionup
\section{Extended Related Work}\label{app:r}
\cutsectiondown

\textbf{Multi-task reinforcement learning.}
Multi-task reinforcement learning aims to learn an inductive bias that can be shared and used across a variety of related RL tasks to improve the task generalization.
Early works mostly focused on the transfer learning oriented approaches~\citep{lazaric2012transfer, taylor2009transfer} such as instance transfer~\citep{lazaric2008transfer} or representation transfer~\citep{konidaris2006autonomous}. 
However, these algorithms rely heavily on the prior knowledge about the allowed task differences. 
\citet{hausman2018learning, pinto2017learning, wilson2007multi} proposed to train a multi-task policy with multiple objectives from different tasks.
However, the gradients from different tasks may conflict and hurt the training of other tasks.
To avoid gradient conflict,~\citet{zhang2014regularization, chen2018gradnorm, lin2019adaptive} proposed to explicitly model the task similarity. However, dynamically modulating the loss or the gradient of RL update often results in the instability in optimization. 
Our multi-task learning algorithm also takes the transfer learning oriented viewpoint; MTSGI captures and transfers the task knowledge in terms of the subtask graph. However, our work does not make a strong assumption on the task distribution. We only assume that the task is parameterized by unknown subtask graph, which subsumes many existing compositional tasks (\eg,~\citet{oh2017zero, Andreas2017Modular, huang2018neural}, etc).

\textbf{Extended - web navigating RL agent.}
Previous work introduced MiniWoB \citep{shi2017world} and MiniWoB++ \citep{liu2018reinforcement} benchmarks that are manually curated sets of simulated toy environments for the web navigation problem.
They formulated the problem as acting on a page represented as a Document Object Model (DOM), a hierarchy of objects in the page.
The agent is trained with human demonstrations and online episodes in an RL loop.
\citet{jia2018domqnet} proposed a graph neural network based DOM encoder and a multi-task formulation of the problem similar to this work.
\citet{gur2018learning} introduced a manually-designed curriculum learning method and an LSTM based DOM encoder.
DOM level representations of web pages pose a significant sim-to-real gap as simulated websites are considerably smaller (100s of nodes) compared to noisy real websites (1000s of nodes).
As a result, these models are trained and evaluated on the same simulated environments which is difficult to deploy on real websites.
Our work formulates the problem as abstract web navigation on real websites where the objective is to learn a latent subtask dependency graph similar to sitemap of websites.
We propose a multi-task training objective that generalizes from a fixed set of real websites to unseen websites without any demonstration, illustrating an agent capable of navigating real websites for the first time.

\textbf{Planning Approaches for Compositional Task}
Previous work has tackled the compositional tasks using the Hierarchical Task Network (HTN) planning~\citep{sacerdoti1975structure, sacerdoti1975nonlinear, tate1977generating} in a (single) goal-conditioned RL setting.
The HTN allows the agent to reason the tasks at multiple levels of abstraction, when rich knowledge
at those abstraction levels are available.
Specifically, HTN models the primitive tasks (or the \textit{subtasks} in our terminology) by the precondition and the effects, and aim to find the sequence of actions (or the \textit{options} in our terminology) that execute each subtasks via planning on the HTN.
They aim to execute a single goal task, often with assumptions of simpler subtask dependency structures~\citep{ghazanfari2017autonomous, liu2016GTSvisual}
such that the task structure can be constructed from the successful trajectories.
Also, they often require expensive searching to find the solution.
In contrast, we tackle a more general and challenging setting, where each subtask gives a reward (\ie, multi-goal setting)
and the goal is to maximize the cumulative sum of reward within an episode. Moreover, we avoid any expensive searching and propose to use neural network to directly map the task structure into policy. Lastly, we aim to achieve zero-/few-shot task generalization, which is not achievable with the HTN methods since they require the full specification of the action models in the testing.
%
    \newpage
    \cutsectionup
\section{Additional experiment results}\label{app:result}
\cutsectiondown
\subsection{Full Experiment Results on the performance of the agents on 15 Websites in \wob}
\begin{figure}[H]
    \centering
    \includegraphics[draft=false,width=0.98\linewidth, valign=b]{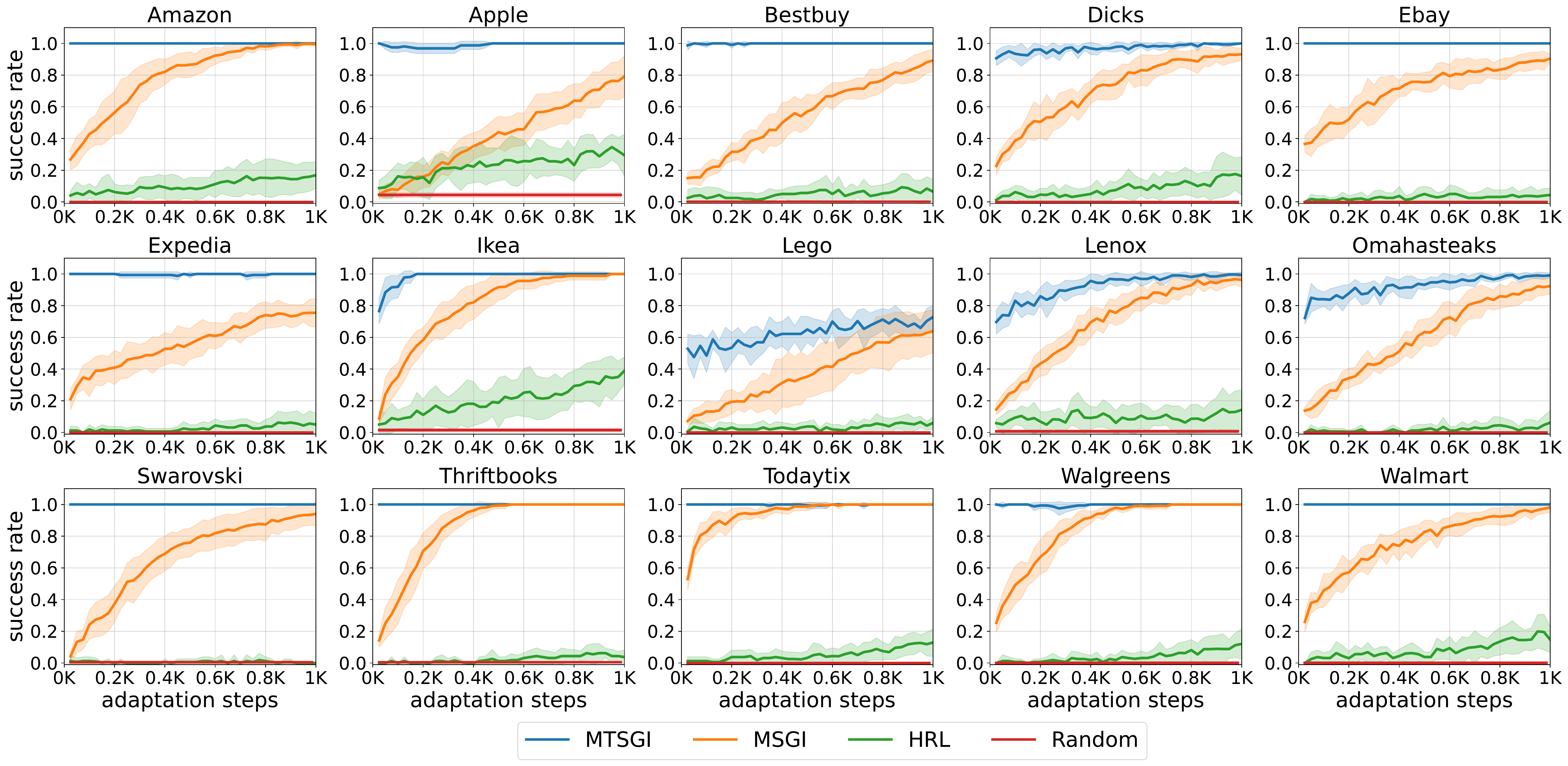}%
    \caption{%
        Results of the success rate for compared methods in the test phase with respect to the adaptation steps on 15 environments in \wob{} domain.
    }
    \label{fig:performance_symwob_all}
\end{figure}

\subsection{Visualization of the multi-task subtask graph inference process}

\begin{figure}[H]
    \centering
    \includegraphics[draft=false,width=0.38\linewidth, valign=b]{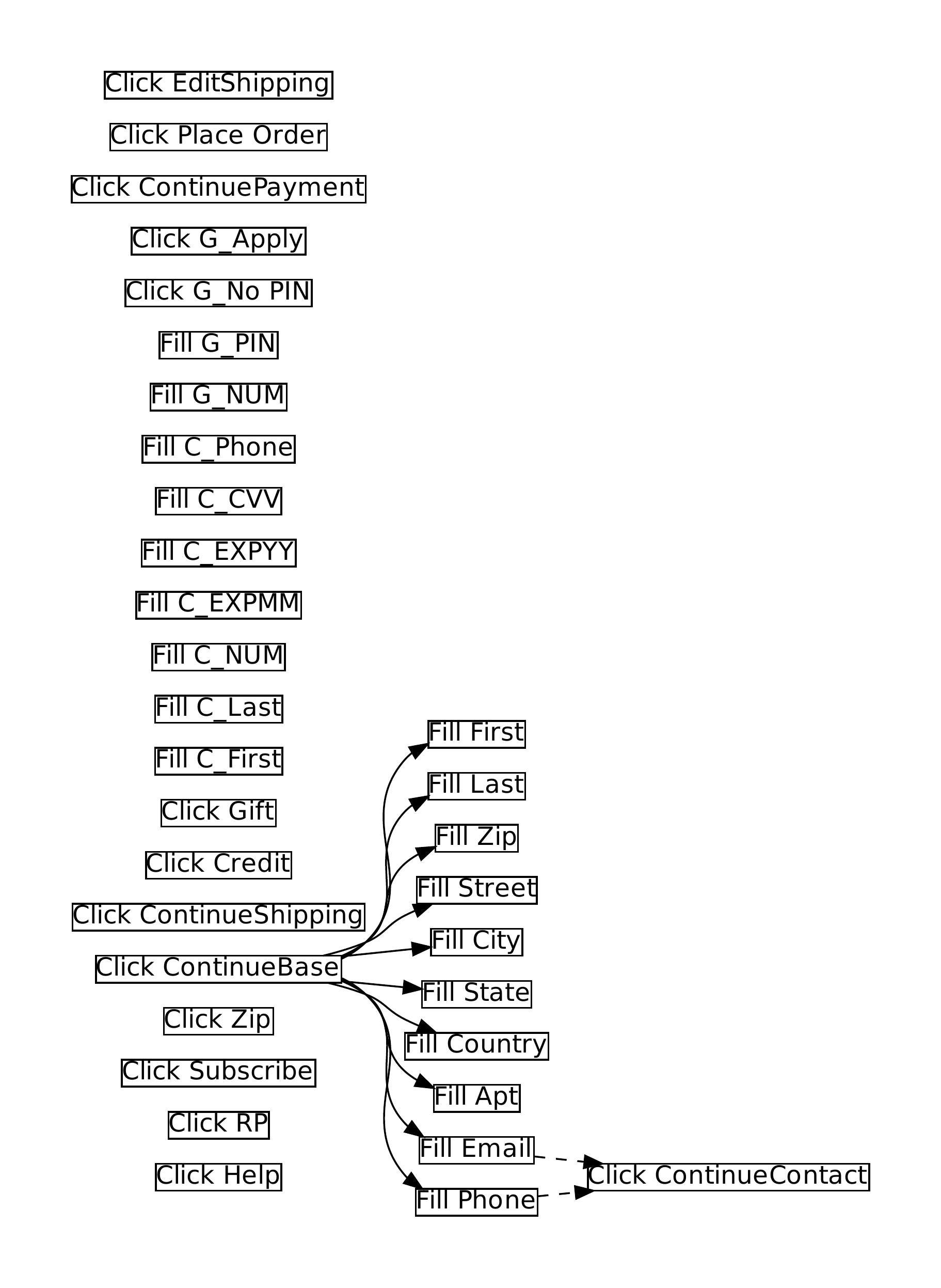}
    \includegraphics[draft=false,width=0.6\linewidth, valign=b]{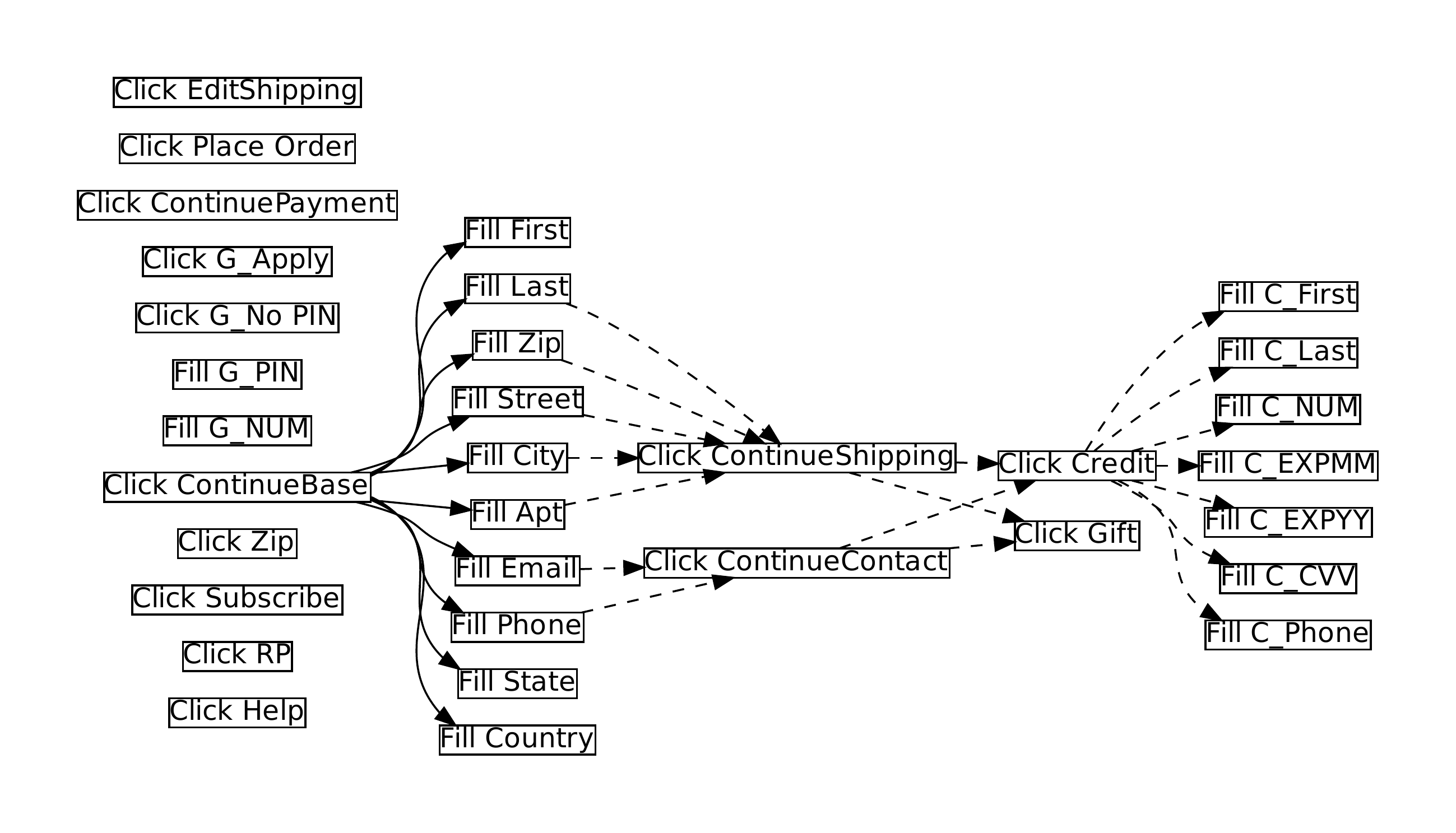}\\
    \includegraphics[draft=false,width=0.9\linewidth, valign=b]{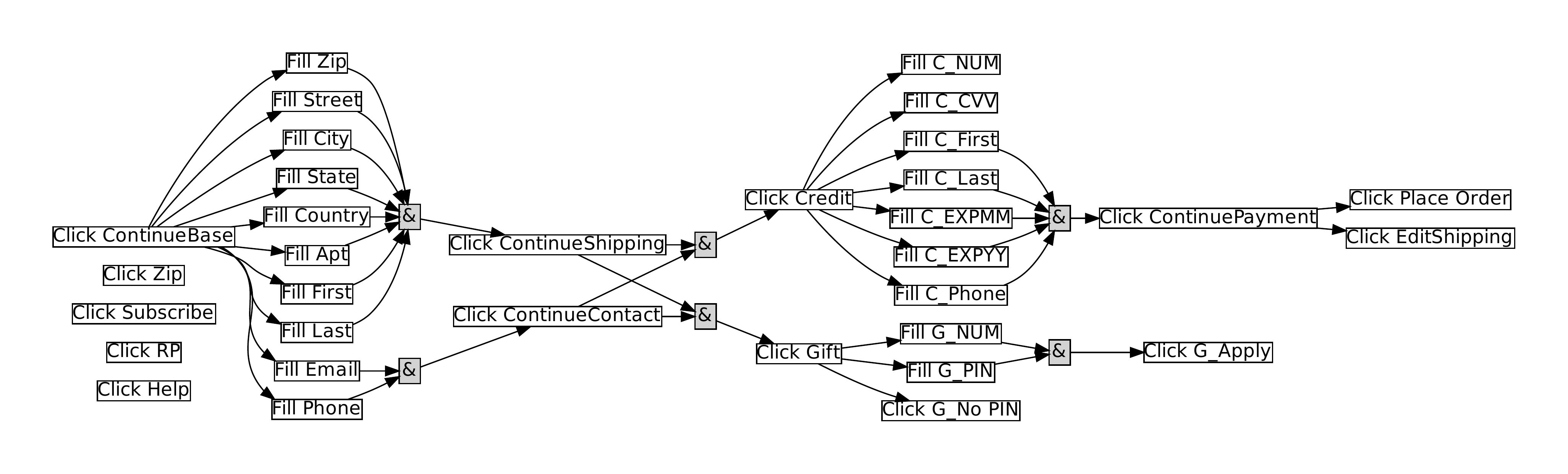}%
    \caption{%
        The subtask graphs inferred by our MTSGI with varying adaptation steps on \walmart domain: 0 steps ({\em Top, Left}), 400 steps ({\em Top, Right}), and 1000 steps ({\em Bottom}).
    }
    \label{fig:graph_inference_progression}
\end{figure}
\Cref{fig:graph_inference_progression} qualitatively shows how the multi-task subtask graph inference proceeds over the adaptation. In the beginning ({\em Top, Left}), the agent has only prior information and the prior provides a partial information about the subtask graph, which is the preconditions in the first webpage. As the agent further explores the webpage ({\em Top, Right}), the subtask graph gets more accurate, but due to insufficient exploration, there are many missing preconditions, especially for the subtasks in the last webpage.
After sufficient adaptation ({\em Bottom}), our MTSGI can infer quite accurate subtask graph compared to the ground-truth subtask graph in~\Cref{fig:wob_walmart}; \ie, only missing two preconditions.

\subsection{Full Experiment Results on the qualitative evaluation of the inferred subtask graph}
Figures~\ref{fig:wob_walmart}-\ref{fig:wob_walgreens} qualitatively evaluate the task inference of our MTSGI by comparing the inferred subtask graph and the ground-truth. It is clear from the figure that the inferred subtask graphs have only a small portion of missing or redundant edges, while most of the nodes and edges are the same as ground-truth graph.
    \newpage
    \section{Details on \wob{} domain}\label{app:wob}
{
\setlength{\tabcolsep}{3pt}
\begin{table*}[htp]
  \centering
  \small
      \begin{tabular}{r|c|c|c|c|c|c|c|c|}
      \hlineB{2}
        \multicolumn{9}{c}{Subtask Graph Setting}\\ \hlineB{2}
         Task   & \amazon    & \apple & \bestbuy  & \dicks & \ebay & \expedia & \ikea & \lego  \\ \hline
  \#Subtasks 	& 31 & 43 & 37 & 39 & 39 & 36 & 39 & 45 \\ \hline
  \#Distractors	& 4  & 5  & 6  & 6  & 5  & 5  & 5  & 6 \\ \hline
Episode length	& 27 & 40 & 37 & 37 & 37 & 40 & 37 & 37 \\ \hline
        \hline
        \hlineB{2}
         Task     & \lenox    & \omahasteaks & \swarovski & \thriftbooks & \todaytix & \walgreens & \walmart&  \\ \hline
  \#Subtasks 	& 45 & 44 & 45 & 43 & 23 & 38 & 46& \\ \hline
  \#Distractors	& 4  & 6  & 7  & 8  & 3  & 7  & 5 & \\ \hline
Episode length	& 41 & 38 & 38 & 33 & 20 & 50 & 43& \\ \hline
      \end{tabular}
      \vspace{-6pt}
\caption{The task configuration of the tasks in \wob{} domain. Each task is parameterized by different subtask graphs, and the episode length is manually set according to the challengeness of the tasks.}
  \label{tab:symwob}
\end{table*}
}
In this paper, we introduce the \wob{} domain, which is a challenging symbolic environment that aims to reflect the hierarchical and compositional aspects of the checkout processes in the real-world websites. There are total 15 different \wob{} websites that are symbolic implementations of the actual websites:  \amazon, \apple, \bestbuy, \dicks, \ebay, \expedia, \ikea, \lego, \lenox, \omahasteaks, \swarovski, \thriftbooks, \todaytix, \walgreens, and \walmart.
All of these websites are generated by analyzing the corresponding real websites and reflecting their key aspects of checkout process.
The main goal of each website is to navigate through the web pages within the website by clicking and filling in the web elements with proper information which leads to the final web page that allows the agent to click on the \texttt{Place\_Order} button, which indicates that the agent has successfully finished the task of checking out.

\subsection{Implementation detail}
In this section, we describe the detailed process of implementing an existing website into a symbolic version. We first fill out the shopping cart with random products, and we proceed until placing the order on the actual website. During the process, we extract all the interactable web elements on the webpage. We repeat this for all the websites and form a shared subtask pool where similar web elements in different websites that have same functionality are mapped to the same subtask in the shared subtask pool. Then, we extract the precondition relationship between subtasks from the website and form the edges in the subtask graph accordingly. Finally, we implement the termination condition and the subtask reward to the failure distractor (See~\Cref{app:distractor}) and the goal subtasks.

\subsection{Comparison of the websites}
\begin{figure}[H]
    \centering
    \includegraphics[draft=false,width=1.0\linewidth, valign=b]{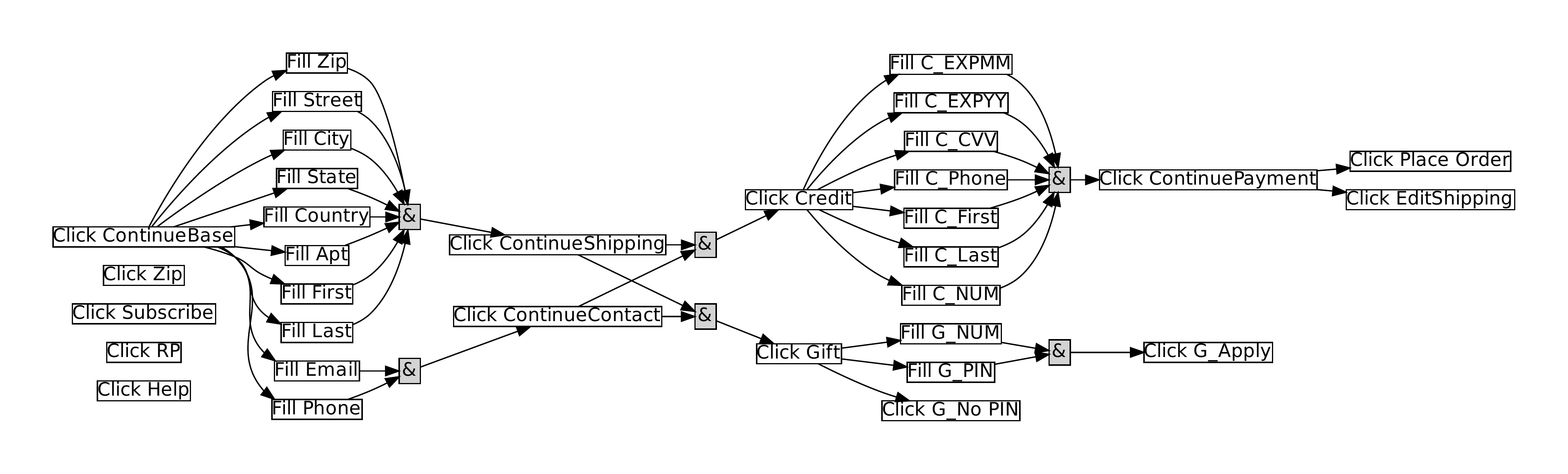}\\
    \includegraphics[draft=false,width=1.0\linewidth, valign=b]{./figure/walmart_inferred_step1040.pdf}%
    \vspace{-8pt}
    \caption{%
        \figtop The ground-truth and \figbottom the inferred subtask graphs of \walmart{} website.
    }
    \label{fig:wob_walmart}
    \vspace{-7pt}
\end{figure}

\begin{figure}[H]
    \centering
    \includegraphics[draft=false,width=1.0\linewidth, valign=b]{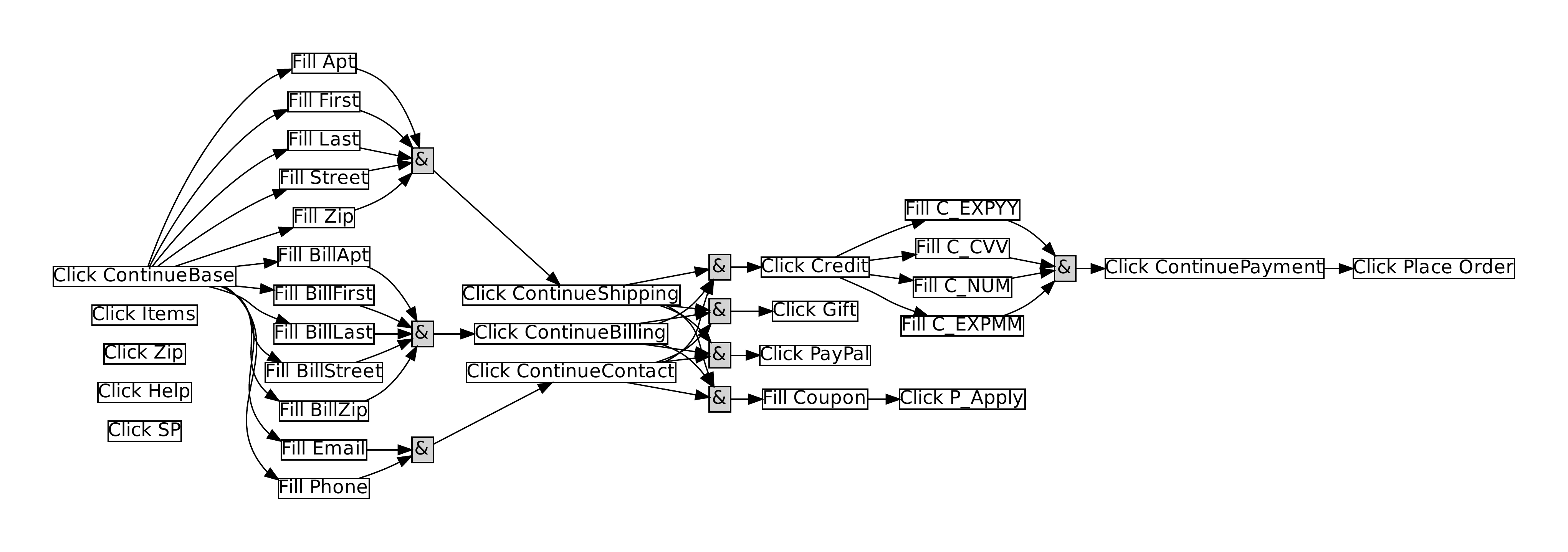}\\
    \includegraphics[draft=false,width=1.0\linewidth, valign=b]{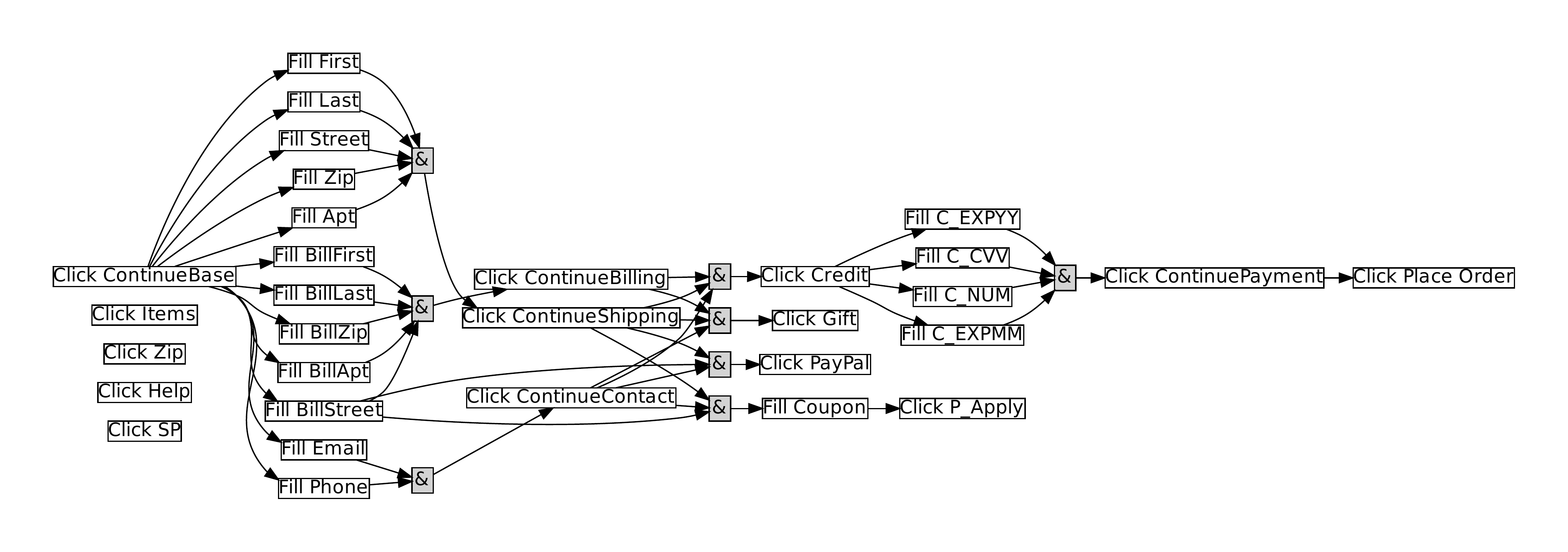}%
    \vspace{-8pt}
    \caption{%
        \figtop The ground-truth and \figbottom the inferred subtask graphs of \dicks{} website.
    }
    \label{fig:wob_dicks}
    \vspace{-7pt}
\end{figure}

\begin{figure}[H]
    \centering
    \includegraphics[draft=false,width=1.0\linewidth, valign=b]{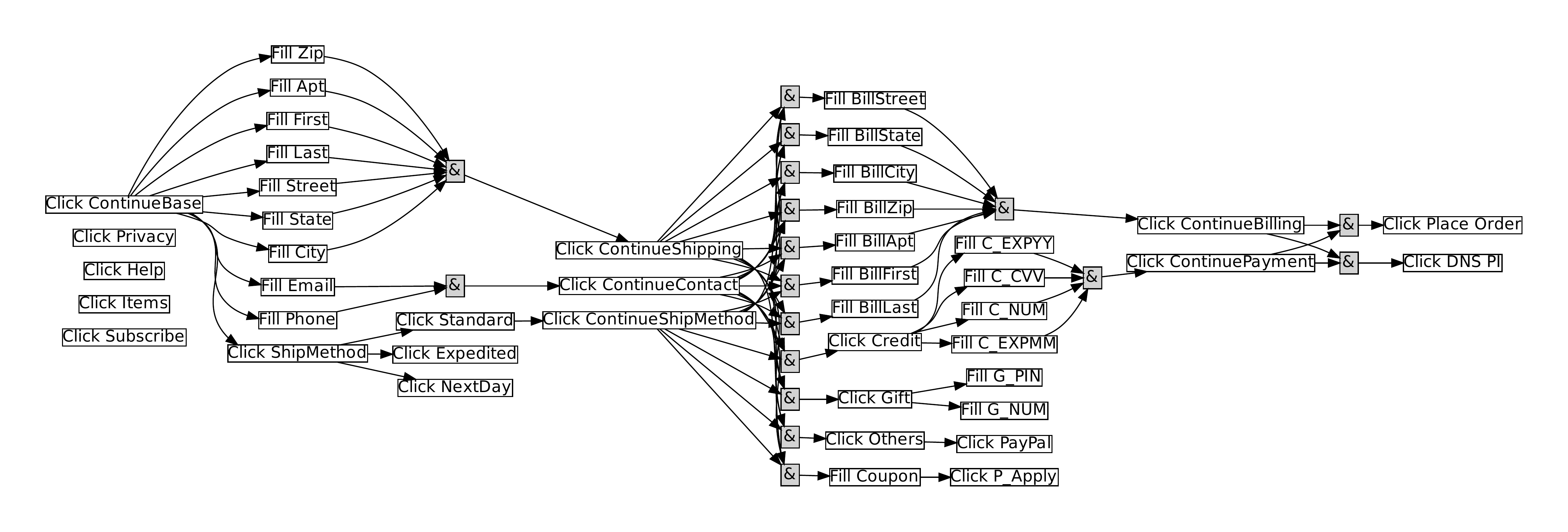}\\
    \includegraphics[draft=false,width=1.0\linewidth, valign=b]{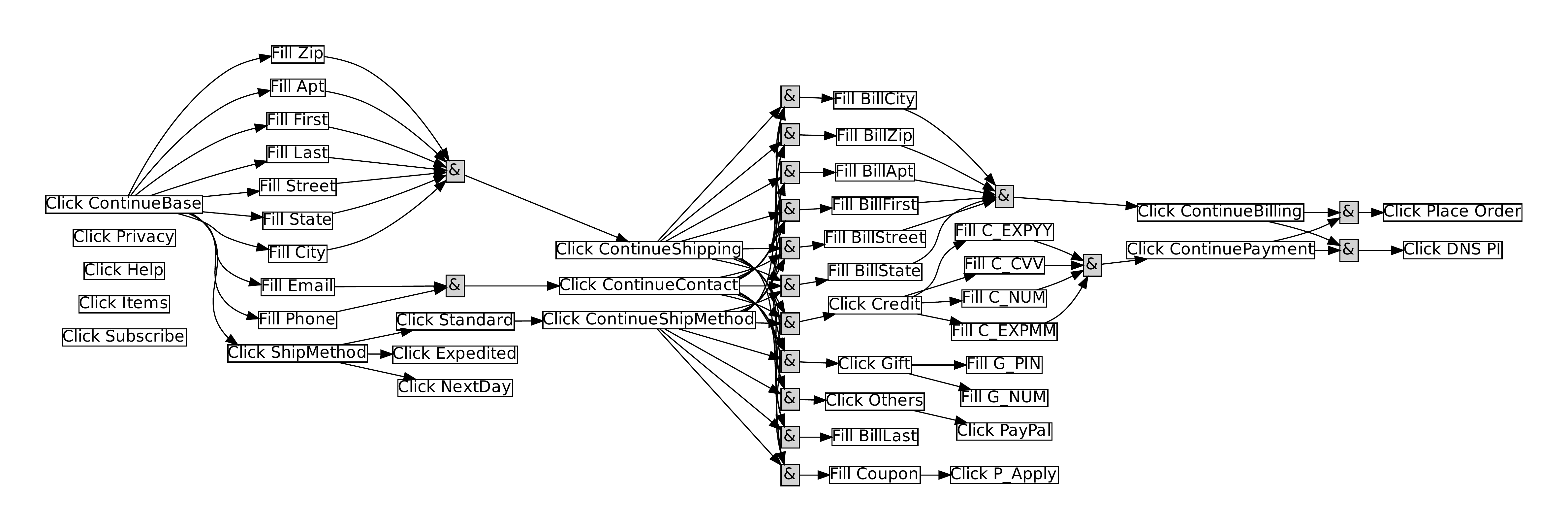}%
    \vspace{-8pt}
    \caption{%
        \figtop The ground-truth and \figbottom the inferred subtask graphs of \bestbuy{} website.
    }
    \label{fig:wob_bestbuy}
    \vspace{-7pt}
\end{figure}

\begin{figure}[H]
    \centering
    \includegraphics[draft=false,width=1.0\linewidth, valign=b]{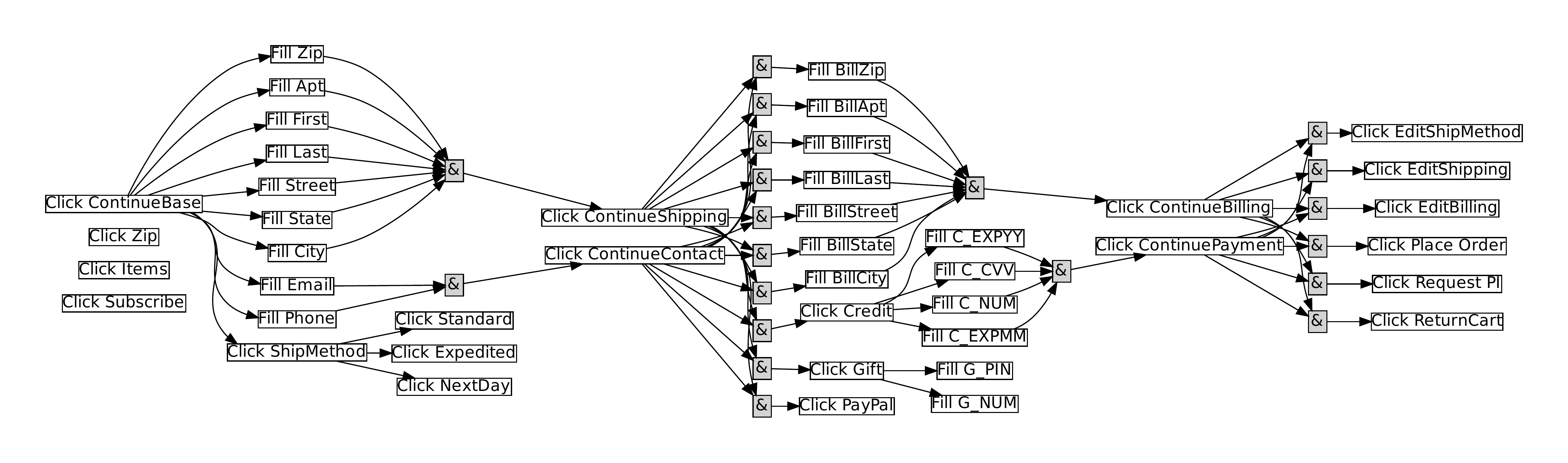}\\
    \includegraphics[draft=false,width=1.0\linewidth, valign=b]{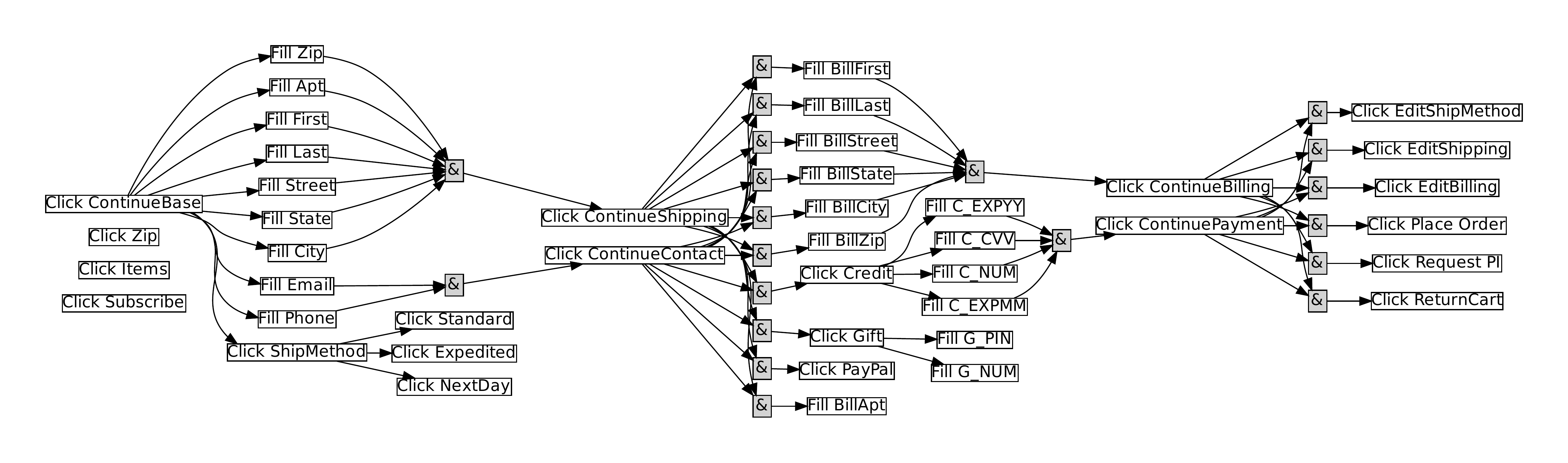}%
    \vspace{-8pt}
    \caption{%
        \figtop The ground-truth and \figbottom the inferred subtask graphs of \apple{} website.
    }
    \label{fig:wob_apple}
    \vspace{-7pt}
\end{figure}

\begin{figure}[H]
    \centering
    \includegraphics[draft=false,width=0.9\linewidth, valign=b]{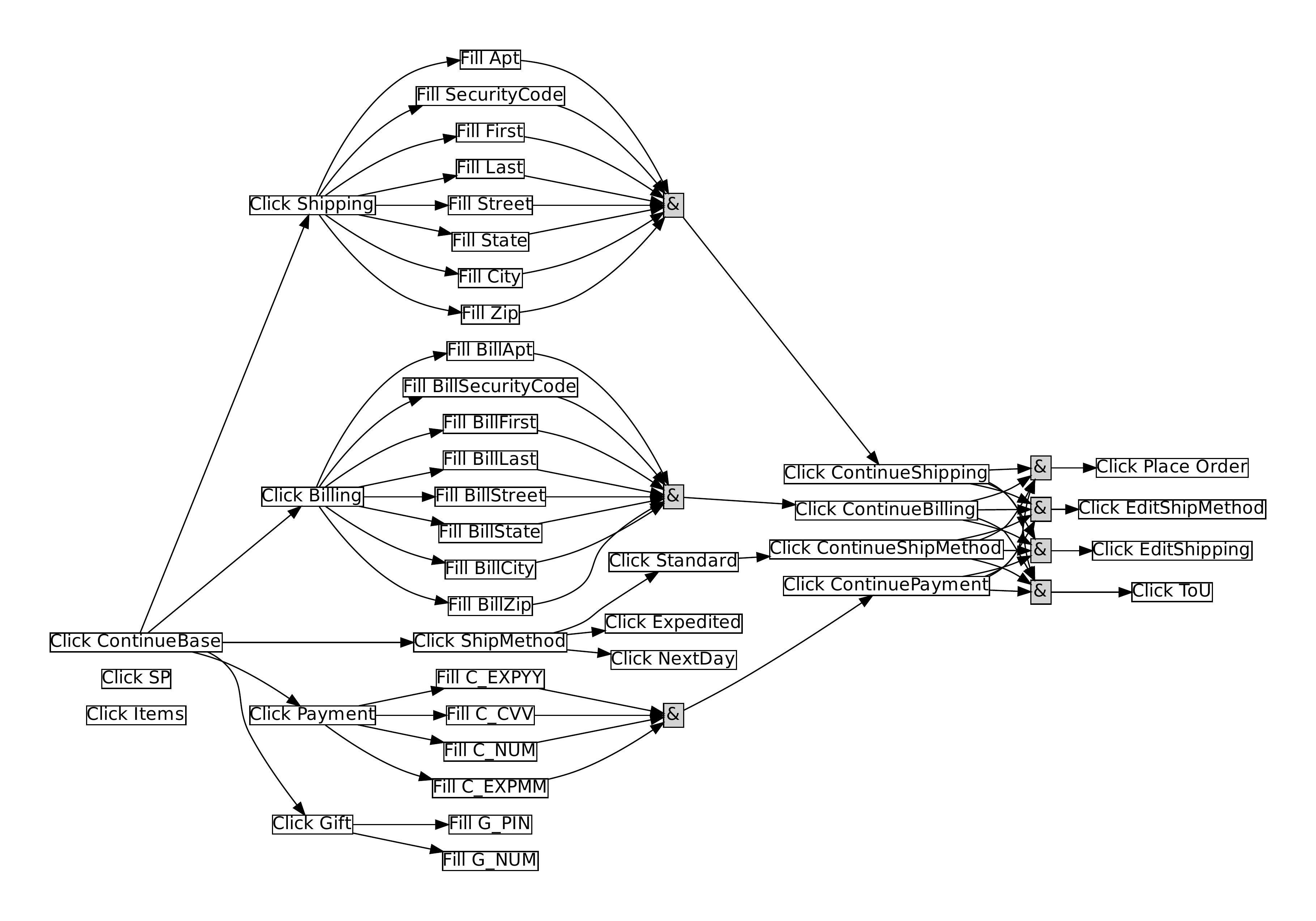}\\
    \includegraphics[draft=false,width=0.9\linewidth, valign=b]{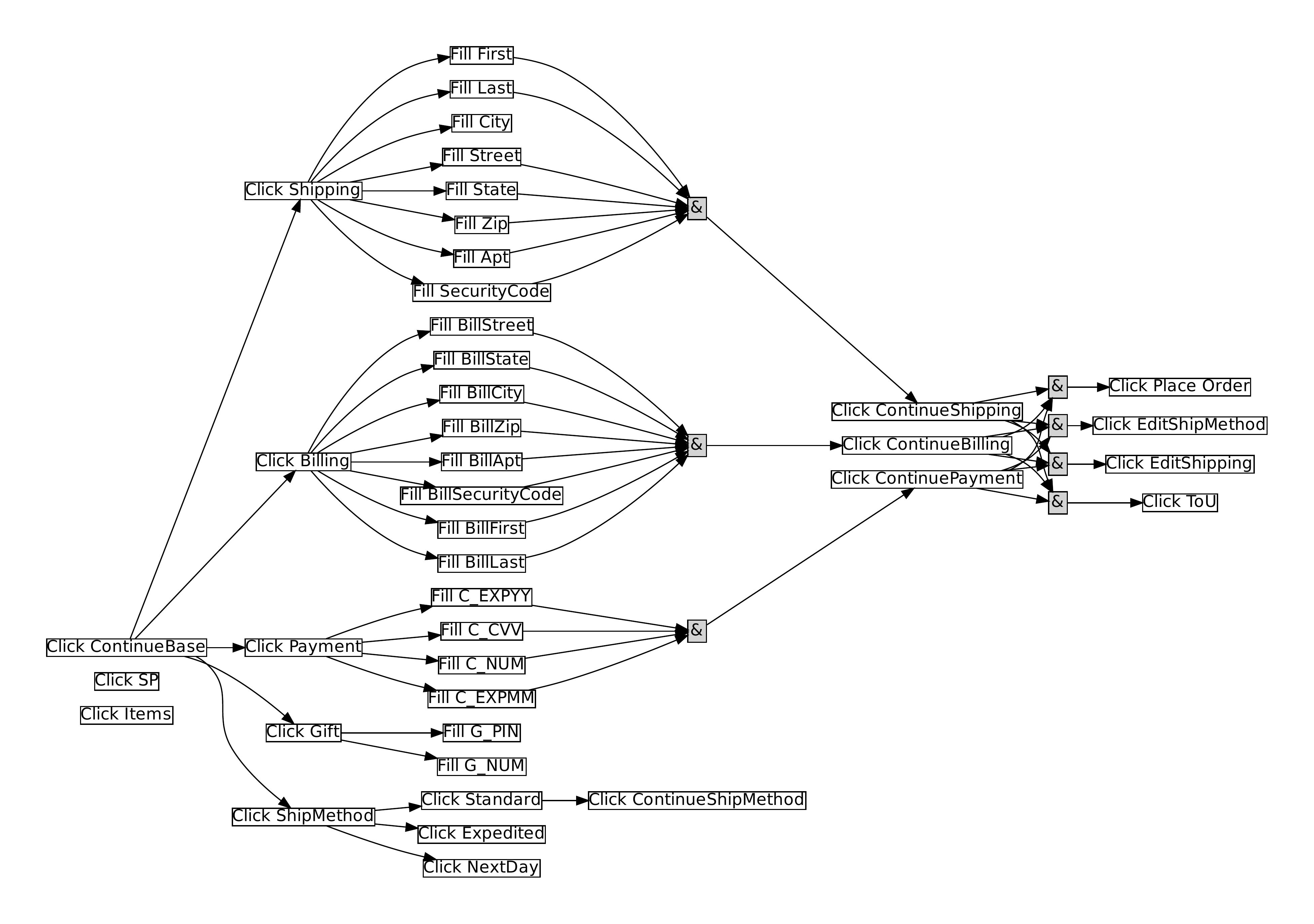}%
    \vspace{-8pt}
    \caption{%
        \figtop The ground-truth and \figbottom the inferred subtask graphs of \amazon{} website.
    }
    \label{fig:wob_amazon}
    \vspace{-7pt}
\end{figure}

\begin{figure}[H]
    \centering
    \includegraphics[draft=false,width=0.8\linewidth, valign=b]{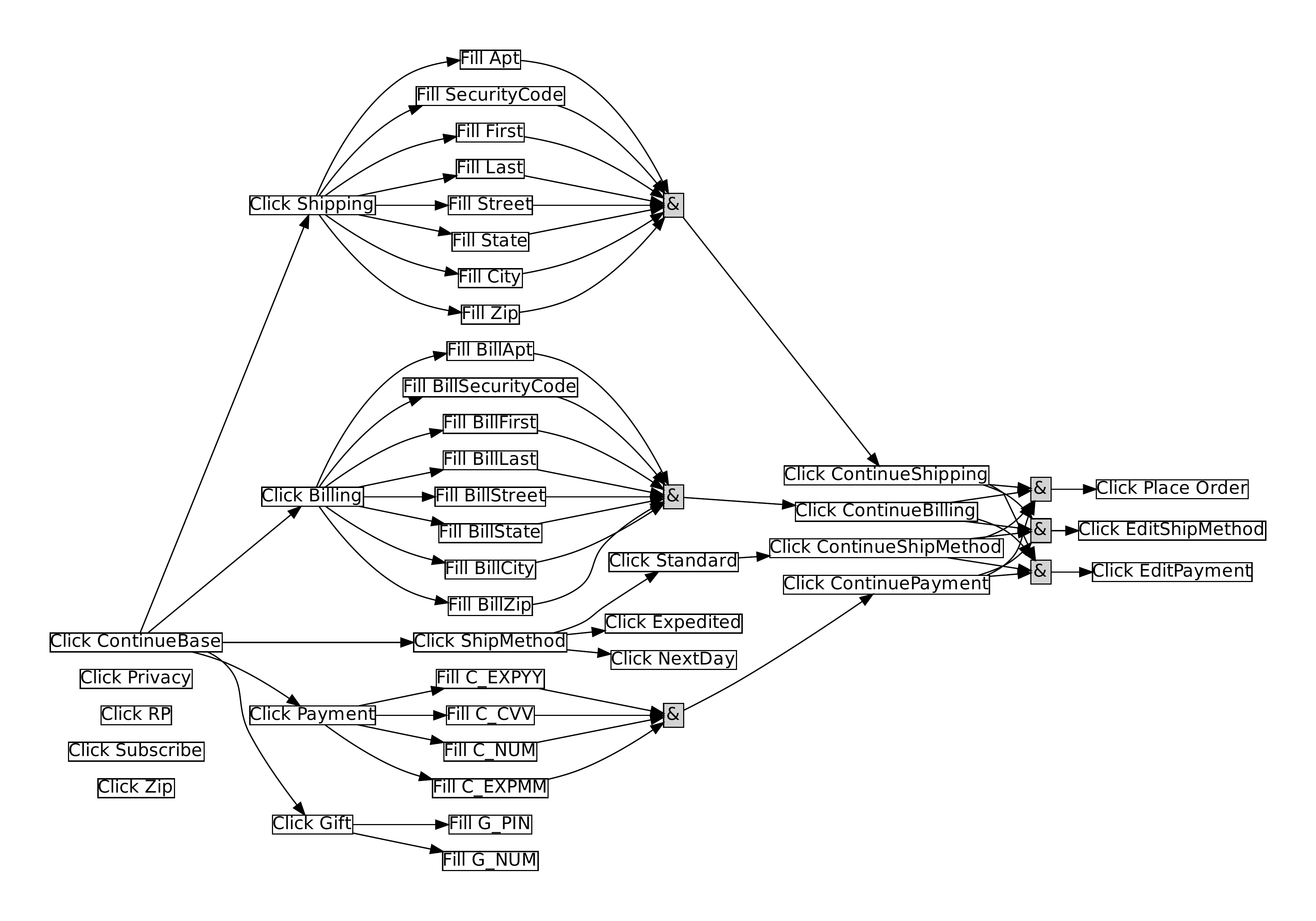}\\
    \includegraphics[draft=false,width=0.73\linewidth, valign=b]{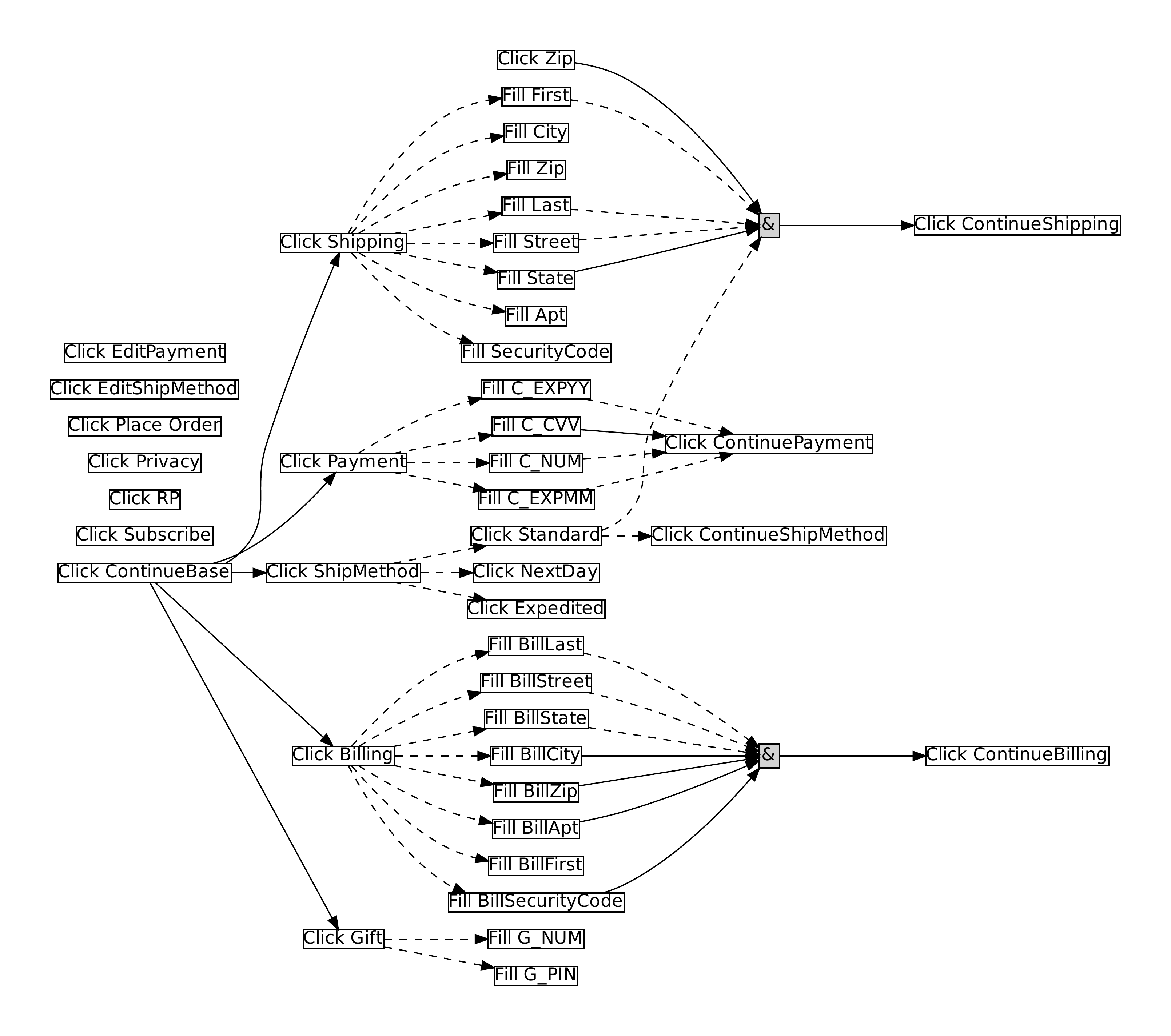}%
    \vspace{-8pt}
    \caption{%
        \figtop The ground-truth and \figbottom the inferred subtask graphs of \ebay{} website.
    }
    \label{fig:wob_ebay}
    \vspace{-7pt}
\end{figure}

\begin{figure}[H]
    \centering
    \includegraphics[draft=false,width=1.0\linewidth, valign=b]{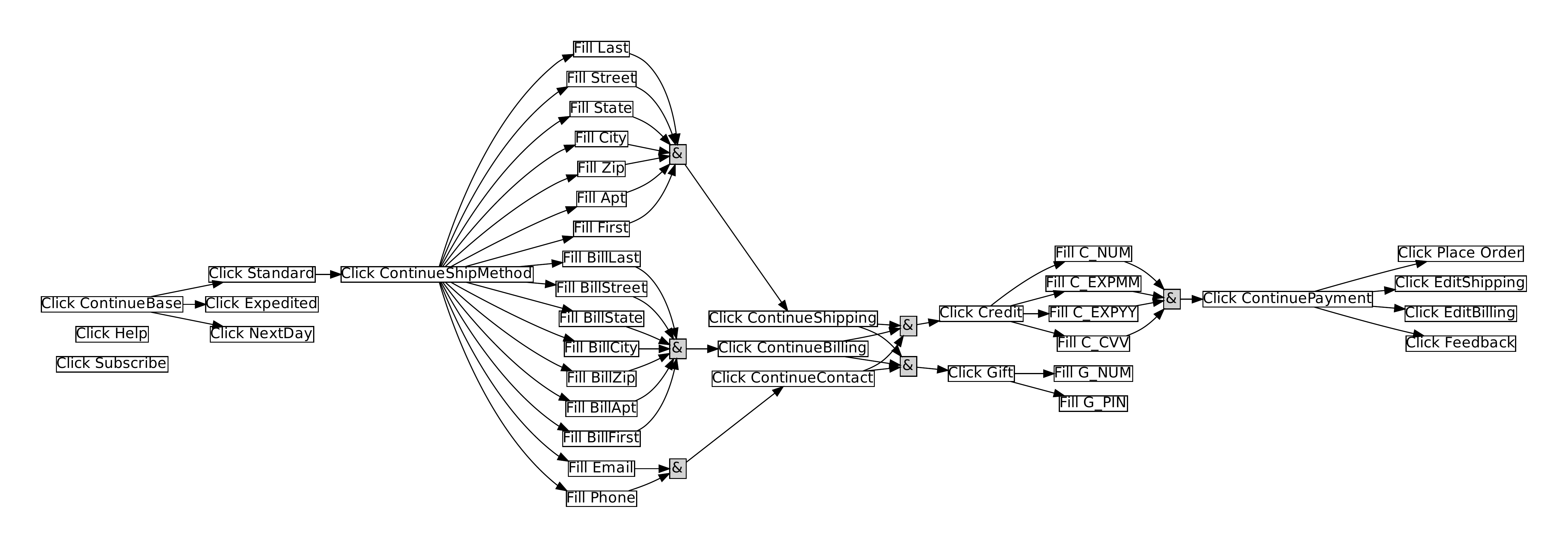}\\
    \includegraphics[draft=false,width=1.0\linewidth, valign=b]{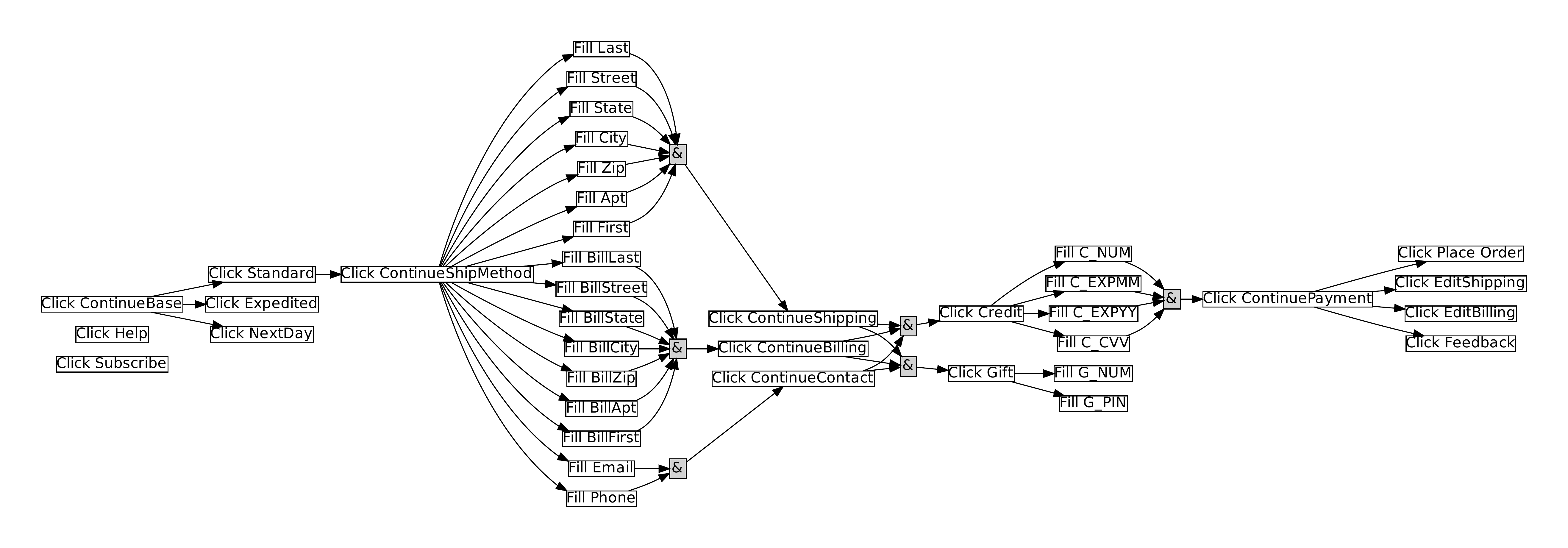}%
    \vspace{-8pt}
    \caption{%
        \figtop The ground-truth and \figbottom the inferred subtask graphs of \ikea{} website.
    }
    \label{fig:wob_ikea}
    \vspace{-7pt}
\end{figure}

\begin{figure}[H]
    \centering
    \includegraphics[draft=false,width=0.65\linewidth, valign=b]{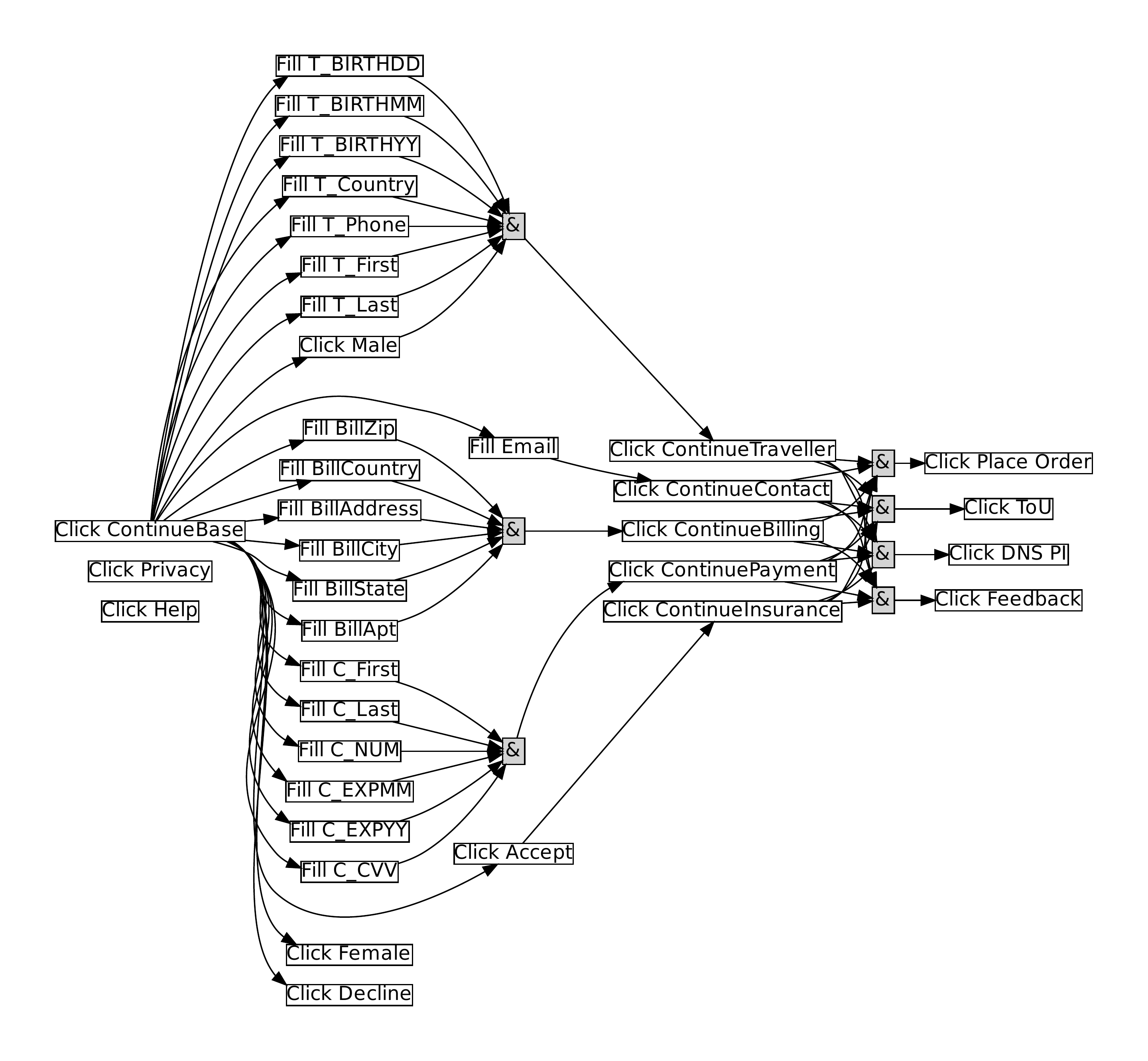}\\
    \includegraphics[draft=false,width=0.7\linewidth, valign=b]{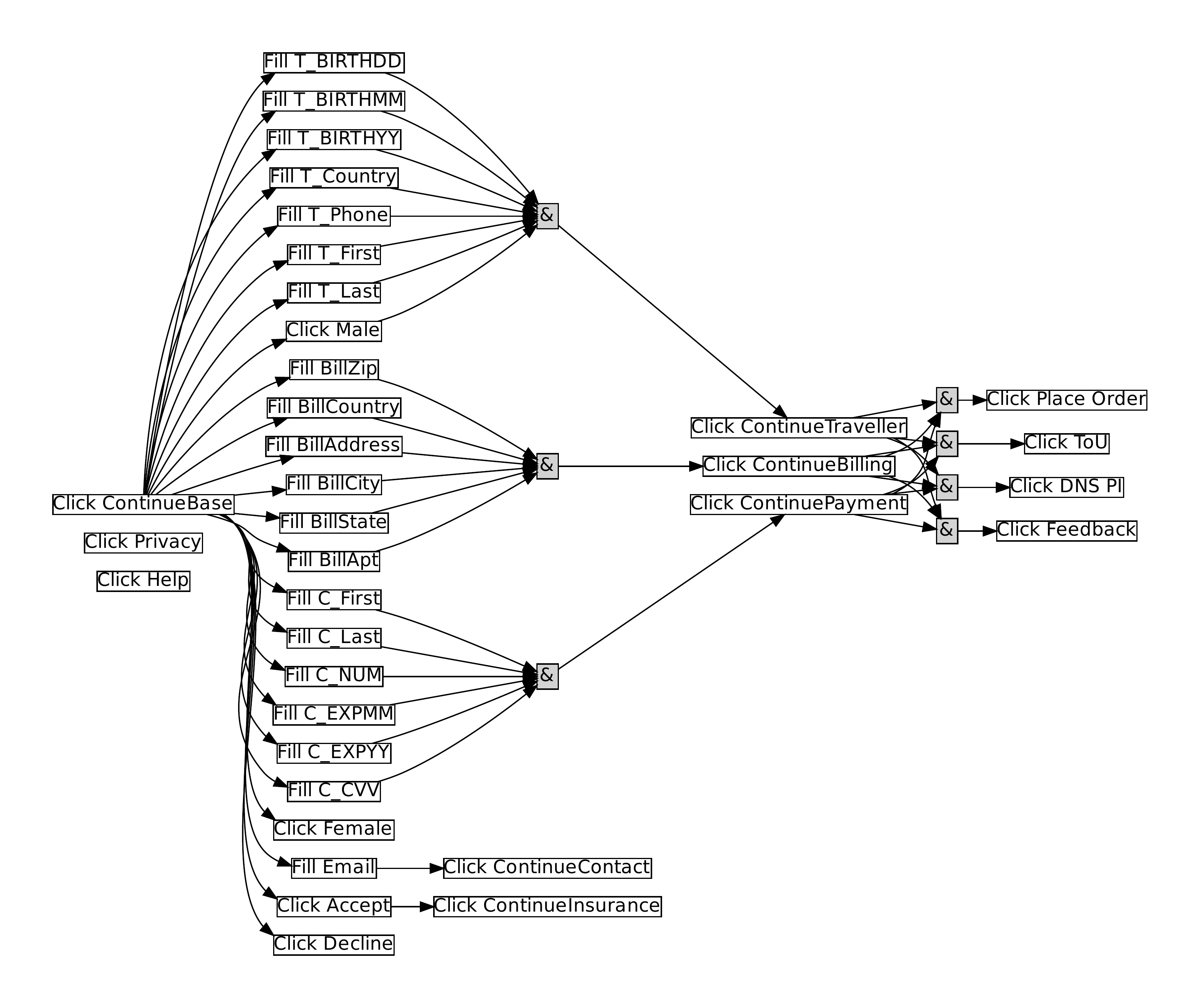}%
    \vspace{-8pt}
    \caption{%
        \figtop The ground-truth and \figbottom the inferred subtask graphs of \expedia{} domain.
    }
    \label{fig:wob_expedia}
    \vspace{-7pt}
\end{figure}

\begin{figure}[H]
    \centering
    \includegraphics[draft=false,width=1.0\linewidth, valign=b]{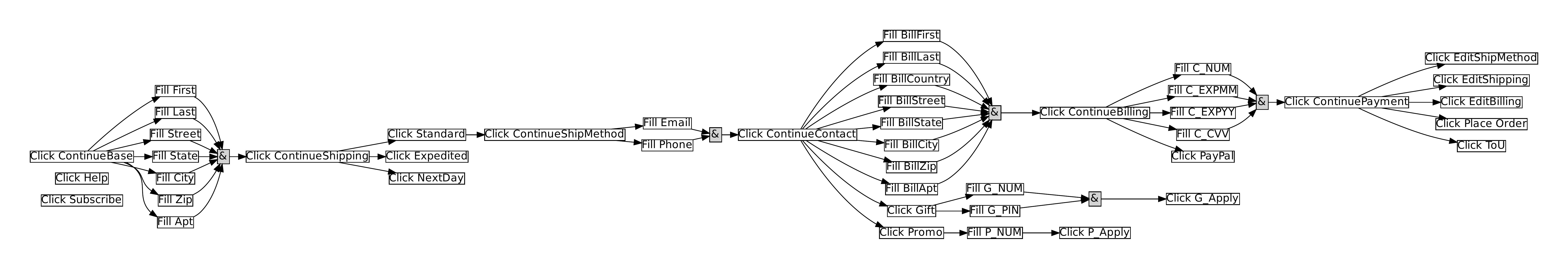}\\
    \includegraphics[draft=false,width=1.0\linewidth, valign=b]{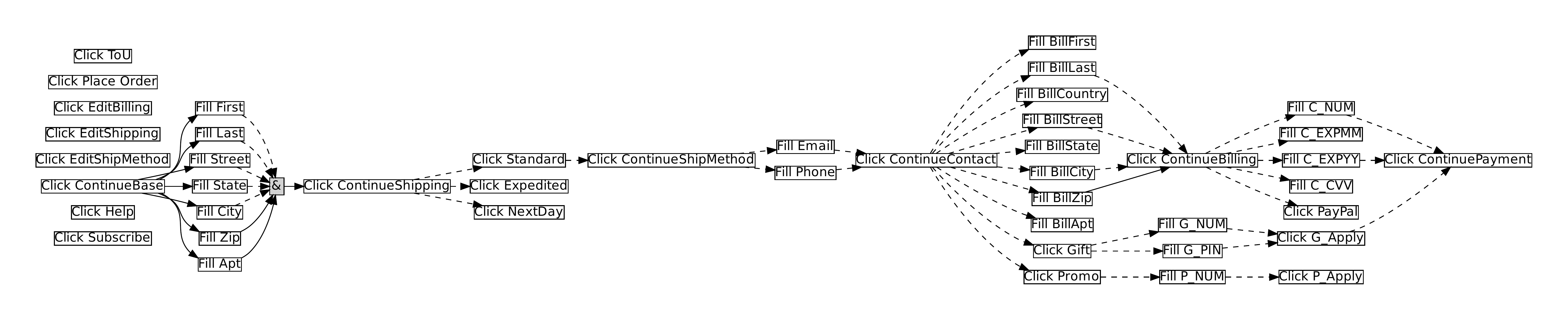}%
    \vspace{-8pt}
    \caption{%
        \figtop The ground-truth and \figbottom the inferred subtask graphs of \lego{} domain.
    }
    \label{fig:wob_lego}
    \vspace{-7pt}
\end{figure}

\begin{figure}[H]
    \centering
    \includegraphics[draft=false,width=1.0\linewidth, valign=b]{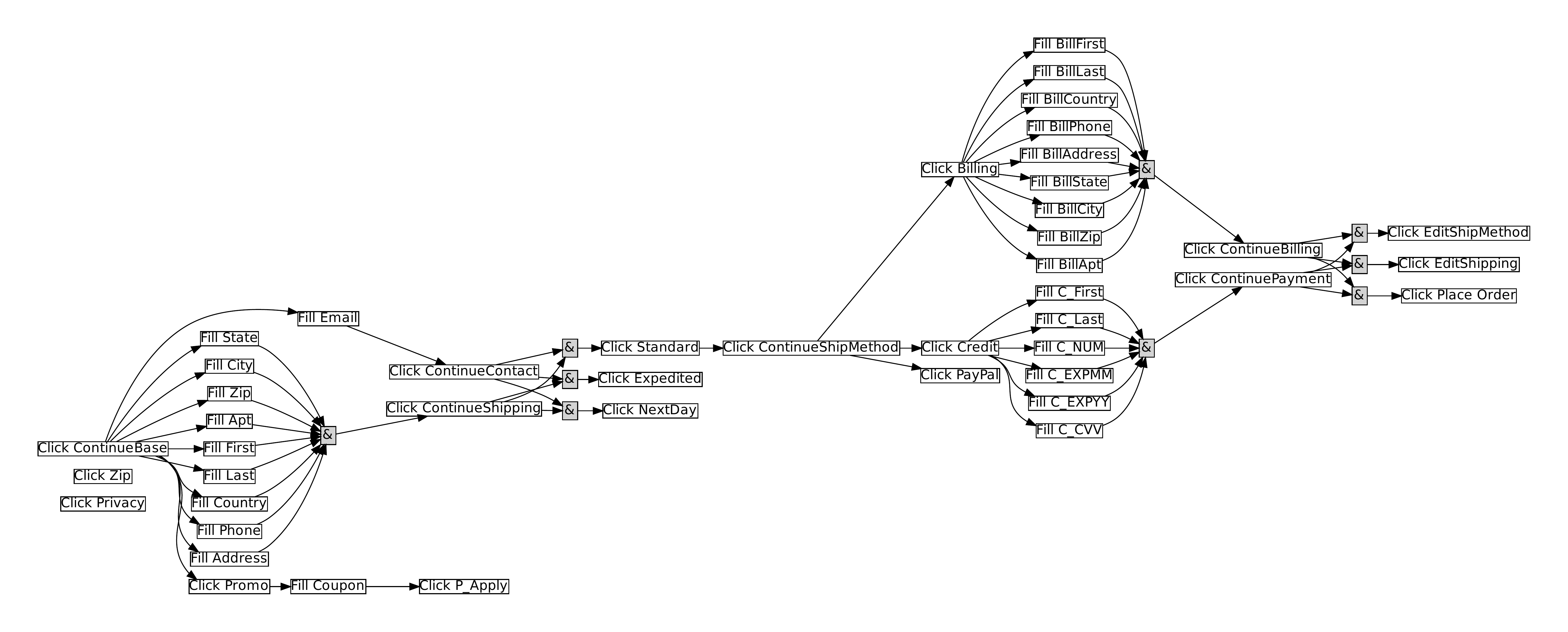}\\
    \includegraphics[draft=false,width=1.0\linewidth, valign=b]{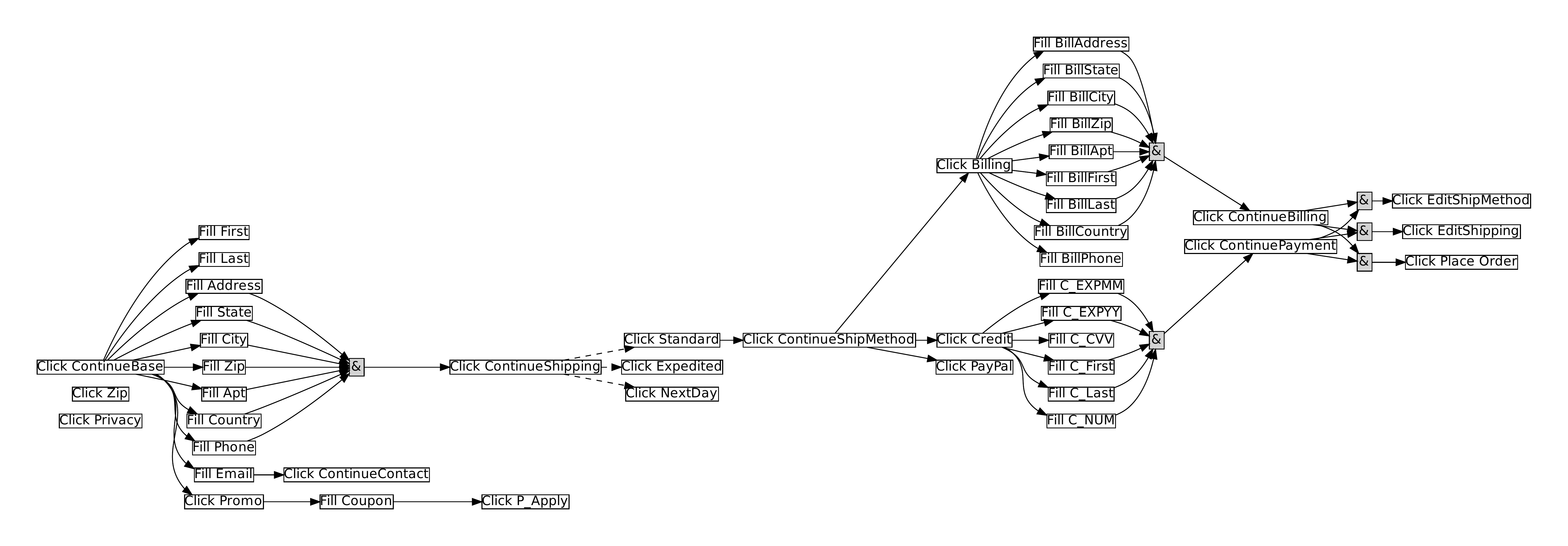}%
    \vspace{-8pt}
    \caption{%
        \figtop The ground-truth and \figbottom the inferred subtask graphs of \lenox{} domain.
    }
    \label{fig:wob_lenox}
    \vspace{-7pt}
\end{figure}

\begin{figure}[H]
    \centering
    \includegraphics[draft=false,width=1.0\linewidth, valign=b]{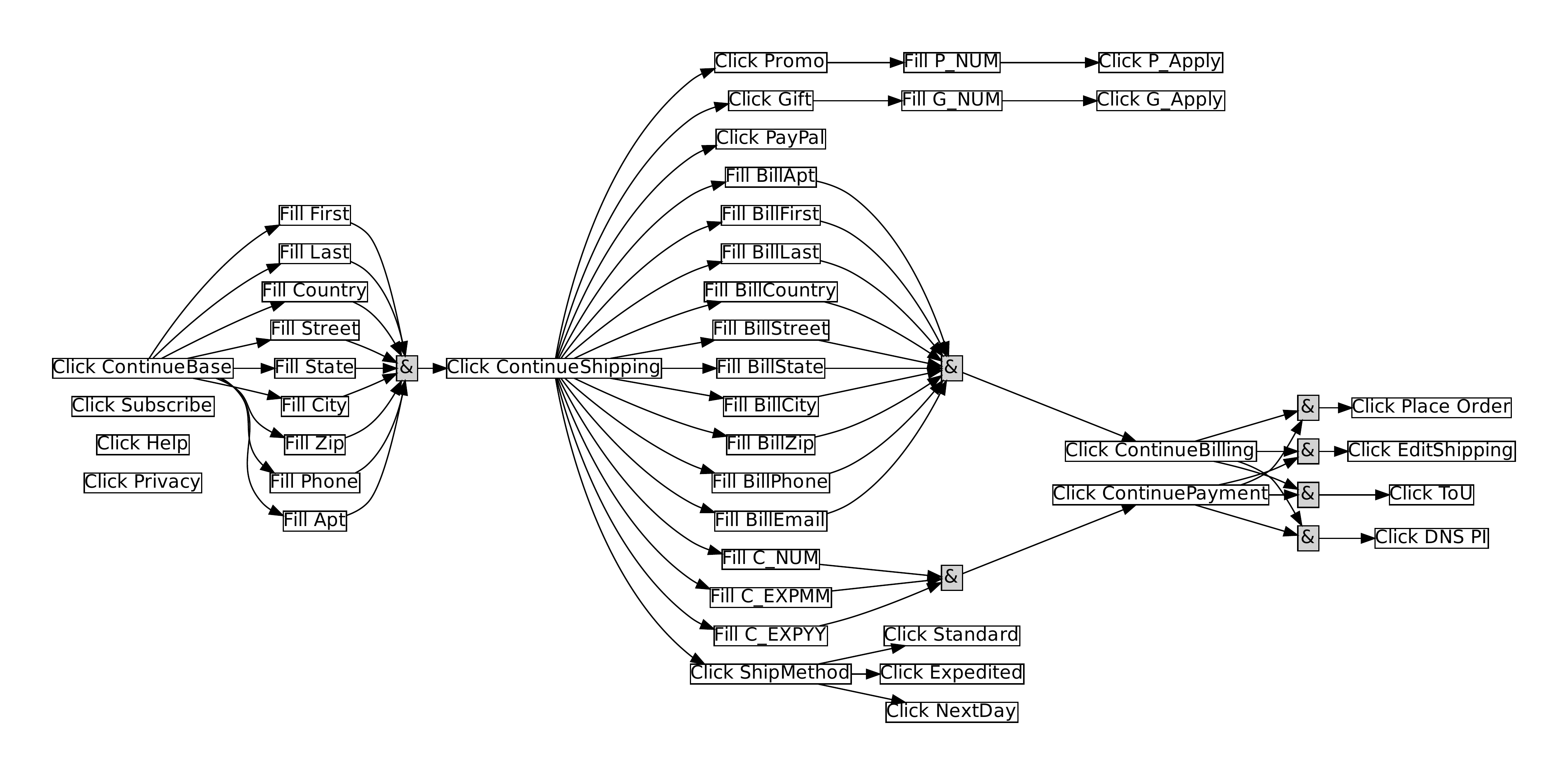}\\
    \includegraphics[draft=false,width=1.0\linewidth, valign=b]{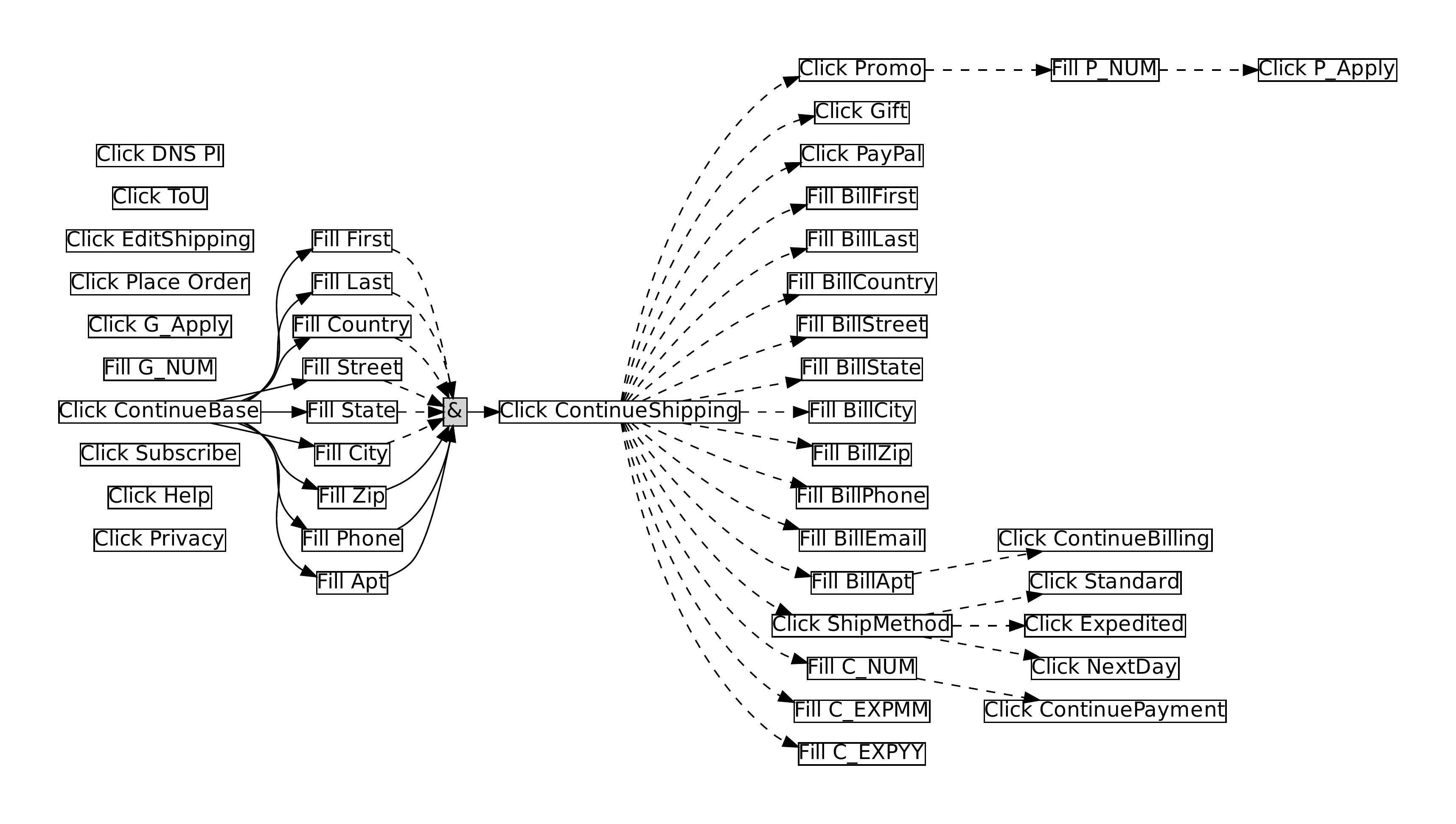}%
    \vspace{-8pt}
    \caption{%
        \figtop The ground-truth and \figbottom the inferred subtask graphs of \omahasteaks{} domain.
    }
    \label{fig:wob_omahasteaks}
    \vspace{-7pt}
\end{figure}

\begin{figure}[H]
    \centering
    \includegraphics[draft=false,width=1.0\linewidth, valign=b]{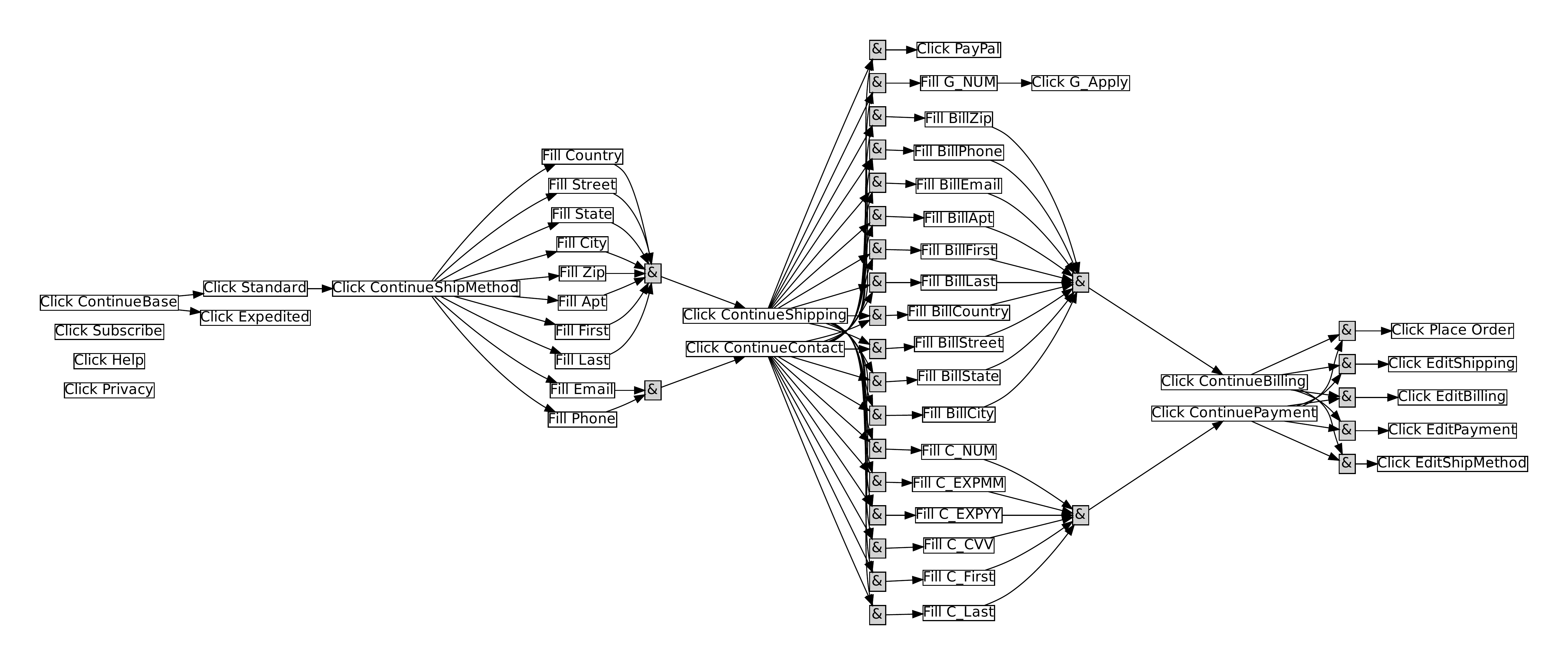}\\
    \includegraphics[draft=false,width=1.0\linewidth, valign=b]{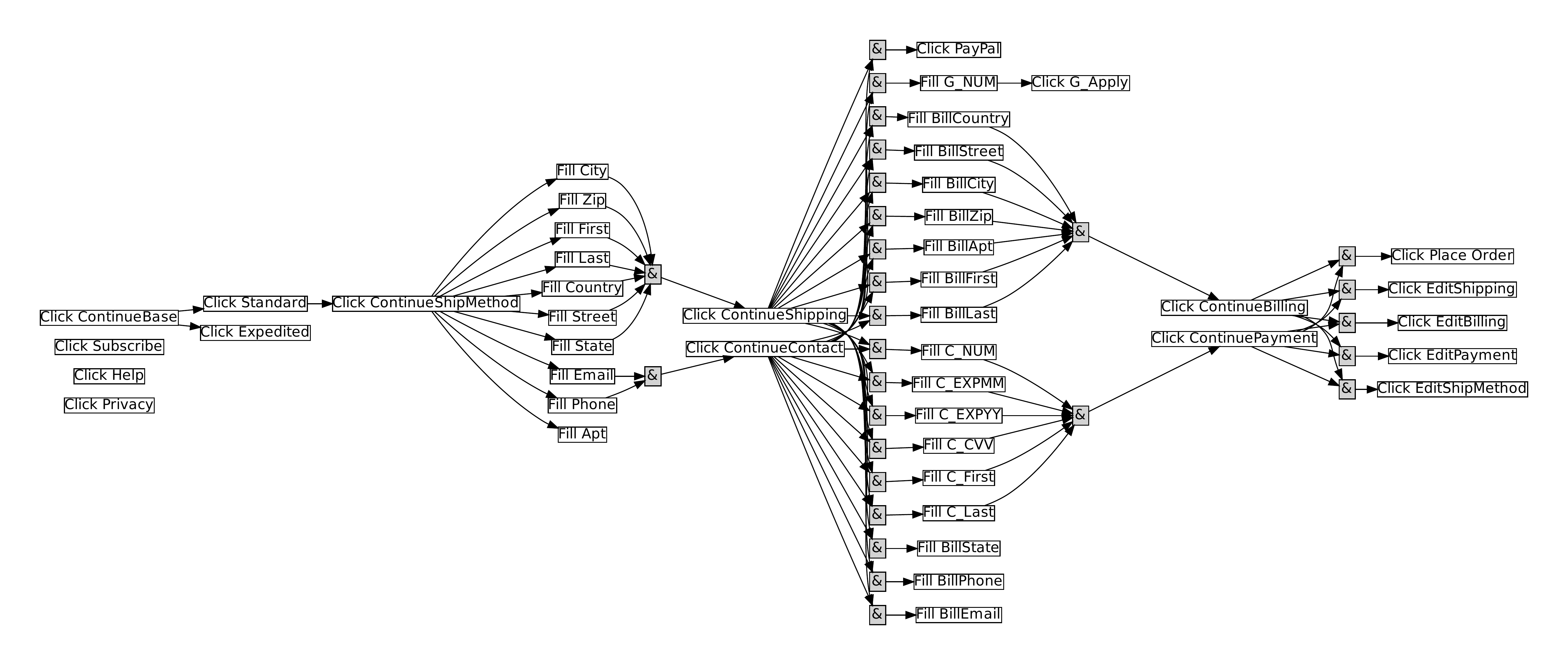}%
    \vspace{-8pt}
    \caption{%
        \figtop The ground-truth and \figbottom the inferred subtask graphs of \swarovski{} domain.
    }
    \label{fig:wob_swarovski}
    \vspace{-7pt}
\end{figure}

\begin{figure}[H]
    \centering
    \includegraphics[draft=false,width=1.0\linewidth, valign=b]{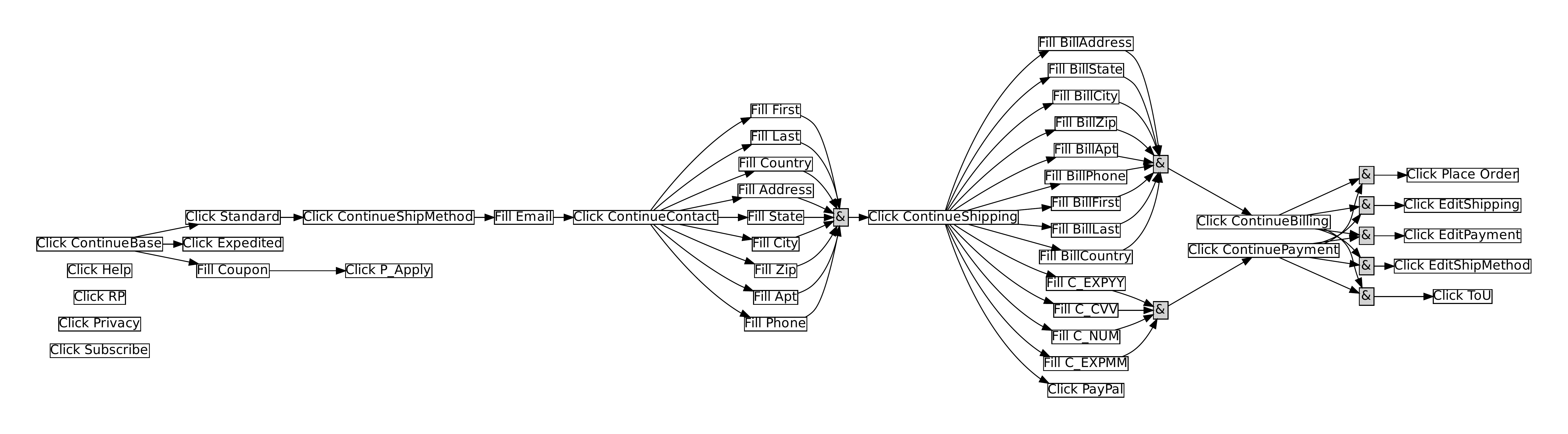}\\
    \includegraphics[draft=false,width=1.0\linewidth, valign=b]{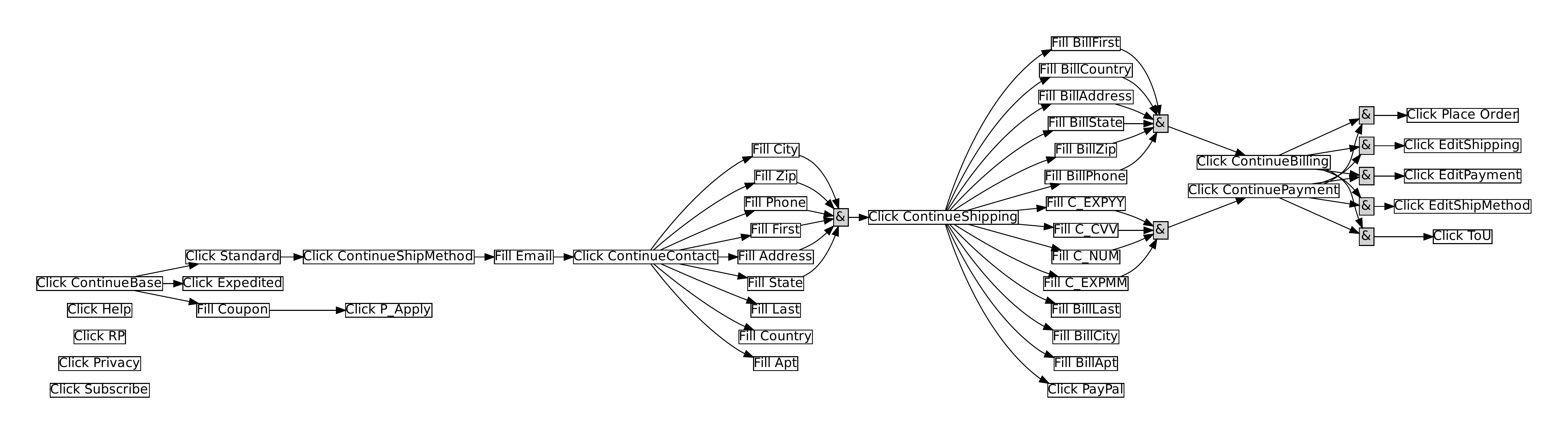}%
    \vspace{-8pt}
    \caption{%
        \figtop The ground-truth and \figbottom the inferred subtask graphs of \thriftbooks{} domain.
    }
    \label{fig:wob_thriftbooks}
    \vspace{-7pt}
\end{figure}

\begin{figure}[H]
    \centering
    \includegraphics[draft=false,width=1.0\linewidth, valign=b]{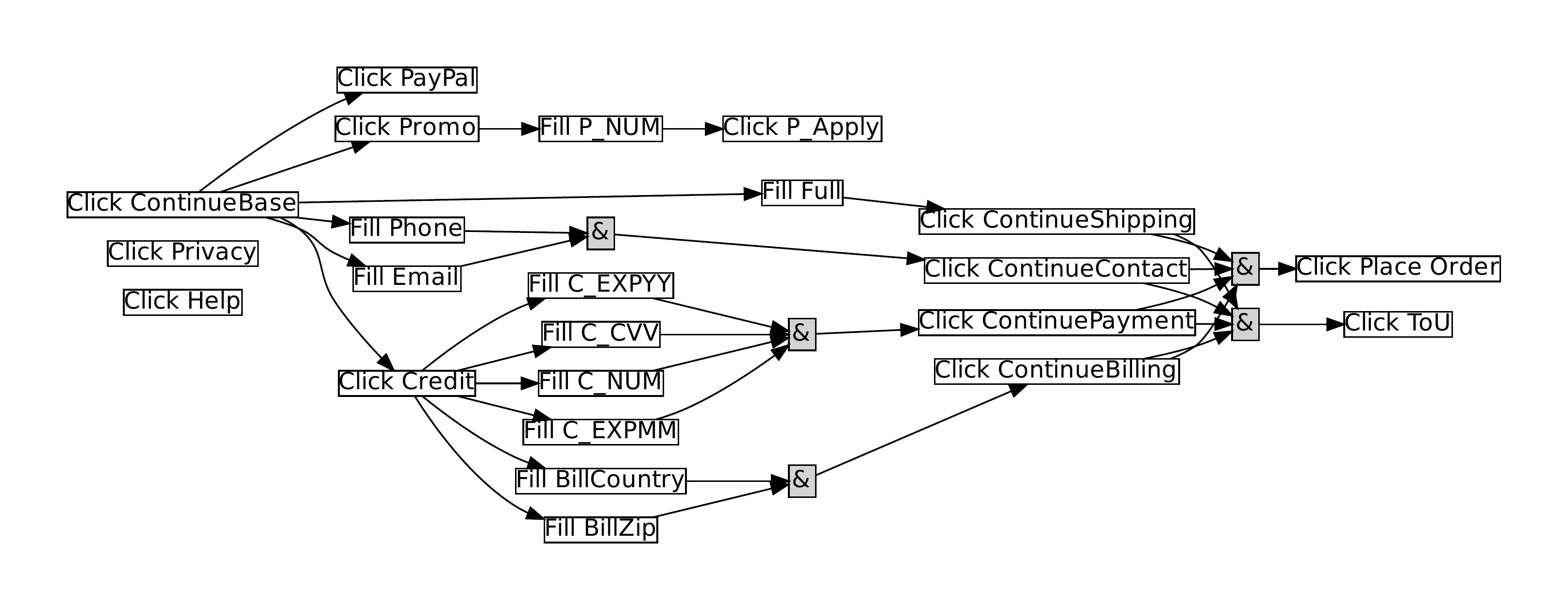}\\
    \includegraphics[draft=false,width=1.0\linewidth, valign=b]{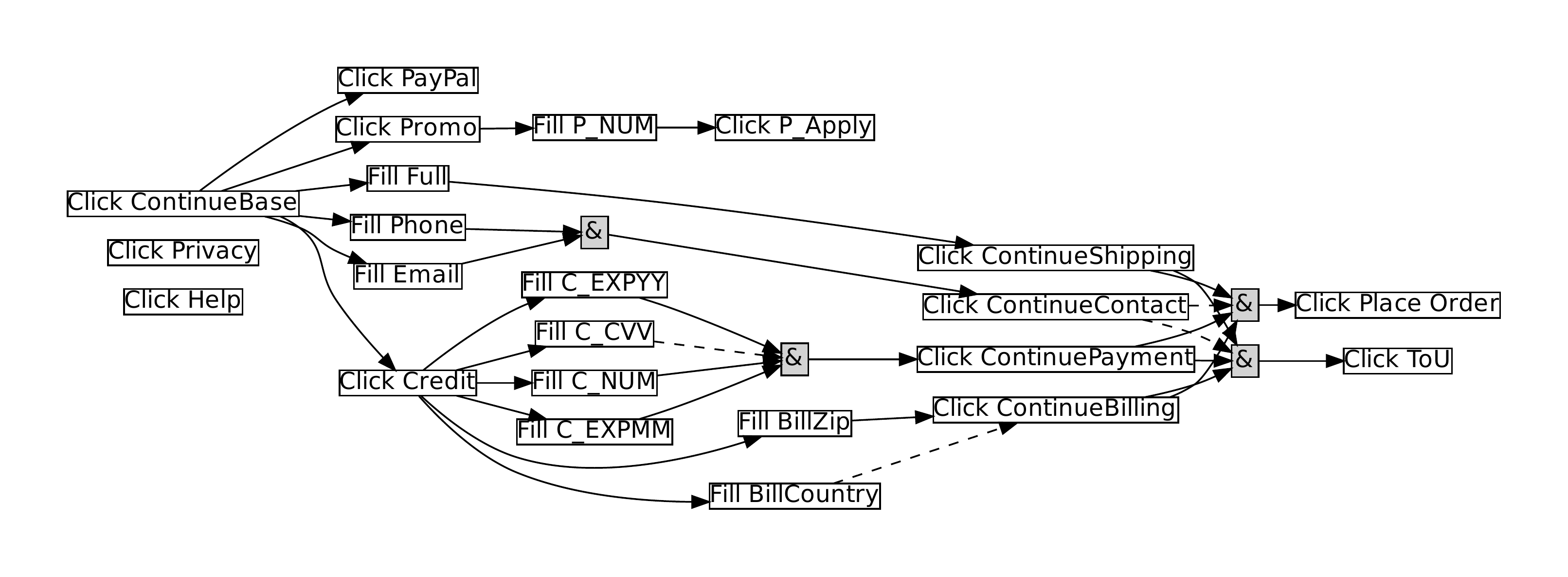}%
    \vspace{-8pt}
    \caption{%
        \figtop The ground-truth and \figbottom the inferred subtask graphs of \todaytix{} domain.
    }
    \label{fig:wob_todaytix}
    \vspace{-7pt}
\end{figure}

\begin{figure}[H]
    \centering
    \includegraphics[draft=false,width=1.0\linewidth, valign=b]{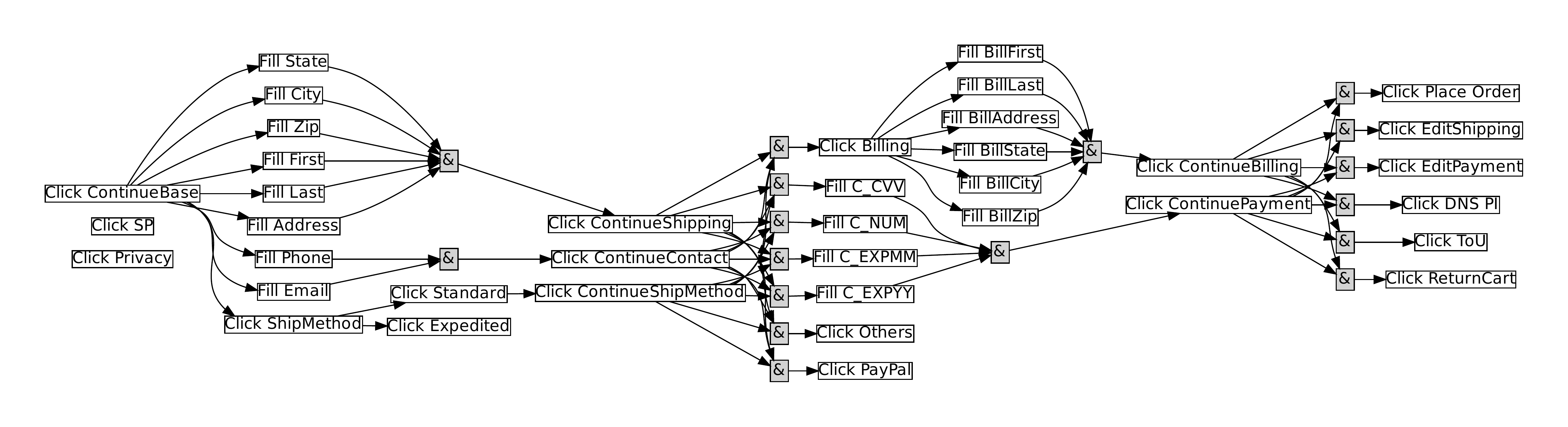}\\
    \includegraphics[draft=false,width=1.0\linewidth, valign=b]{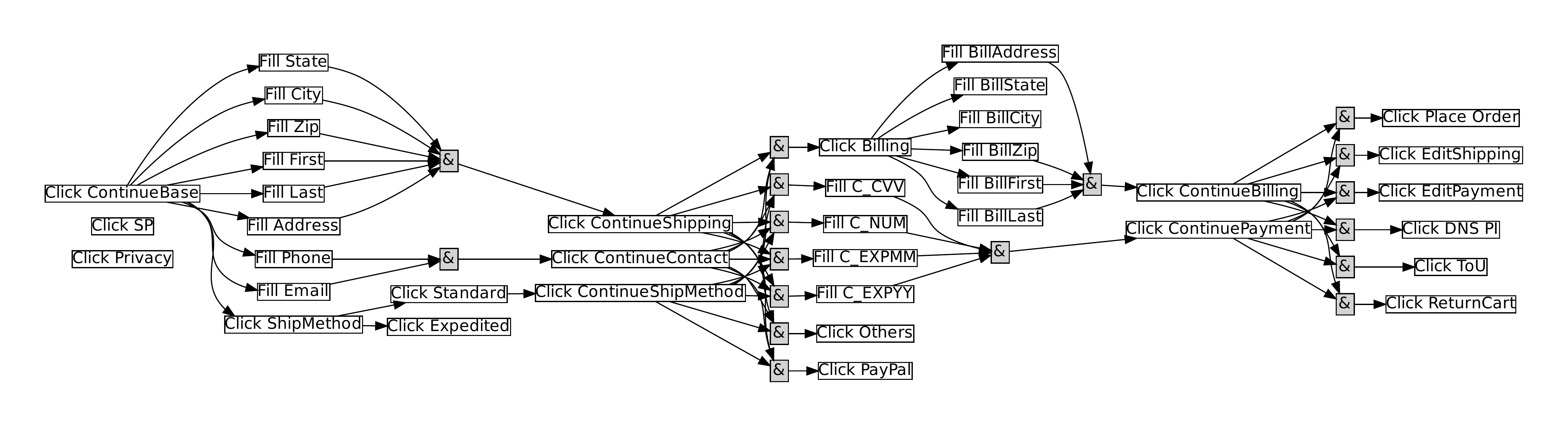}%
    \vspace{-8pt}
    \caption{%
        \figtop The ground-truth and \figbottom the inferred subtask graphs of \walgreens{} domain.
    }
    \label{fig:wob_walgreens}
    \vspace{-7pt}
\end{figure}

The agent's goal on every website is the same, that is placing the checkout order. However, the underlying subtask graphs, or task structure, of the websites are quite diverse, making the task much more challenging for the agent. Figures~\ref{fig:wob_walmart}-\ref{fig:wob_walgreens} visualize the ground truth subtask graph of all the websites. 
One of the major sources of diversity in subtask graphs is in the various ordering of the web pages. 
In a typical website's checkout process, some of the most common web pages include the shipping, billing, and payment web pages, each of which has a collection of corresponding subtasks. 
In~\Cref{fig:wob_walmart}, for example, the shipping web page is represented by the collection of the subtasks on the left side from \texttt{Fill\_Zip} to \texttt{Fill\_Last} and \texttt{Click\_ContinueShipping}, and these come \textit{before} the payment web page that is represented by the subtasks on the right side from \texttt{Click\_Credit} to \texttt{Click\_ContinuePayment}.
On the other hand, in \Cref{fig:wob_amazon}, the similar web pages are either connected in a different ordering, from payment to shipping web page, or placed on the same line side by side. Since the web pages can vary on how they are ordered, it allows the subtask graphs to have a variety of shapes such as deep and narrow as in \Cref{fig:wob_ikea} or wide and shallow as in \Cref{fig:wob_amazon}. Different shape of the subtask graphs means different precondition between the tasks, making it non-trivial for the agent to transfer its knowledge about one to the other. 

Another major source of diversity is the number of web elements in each web page. Let's compare the web elements of the shipping web page in \Cref{fig:wob_dicks} and \Cref{fig:wob_bestbuy}. These are the subtasks that are connected to \texttt{Click\_ContinueShipping} and as well as itself. We can see that the two websites do not have the same number of the web elements for the shipping web pages: the \bestbuy website requires more shipping information to be filled out than the \dicks website. Such variety in the number of web elements, or subtasks, allows the subtask graphs of the websites to have diverse preconditions as well.


\subsection{Distractor subtasks}\label{app:distractor}
In addition to the different task structures among the websites, there are also \textit{distractor} subtasks in the websites that introduces challenging components of navigating the real-world websites. 
There are two different types of distractor subtasks: the one that terminates the current episode with a negative reward and the another one that has no effect. 
The former, which we also call it the \textit{failure} distractor subtask, represents the web elements that lead the agent to some external web pages like \texttt{Help} or \texttt{Terms\_of\_Use} button. 
The latter is just called the distractor subtask, where executing the subtask does not contribute to progressing toward the goal (\eg, \texttt{Fill Coupon} subtask in \bestbuy). 
Each website has varying number of distractor subtasks and along with the shallowness of the task structure, the number of distractor subtasks significantly affects the difficulty of the task.

\section{Details of the agent implementation}\label{app:baseline}
We implement all of our algorithms, including both MTSGI and the baselines, on top of the recently introduced RL framework called Acme \citep{hoffman2020acme}. 

\subsection{MSGI}
Similar to the MTSGI agent, the MSGI agent uses the soft-version of UCB exploration policy as the adaptation policy, instead of its original hard-version due to a better performance. Furthermore, MSGI also uses inductive logic programming (ILP) in order to infer the subtask graph of the current task. But unlike the MTSGI agent that learns prior across the tasks through multi-task learning, the MSGI agent does not exploit the subtask graphs inferred from the previous tasks.
 
\subsection{HRL}
The hierarchical RL (HRL) agent is an option-based agent that can execute temporally extended actions. For the \mining{} domain, where there is a spatial component in the observation input, we use convolutional neural network (CNN) in order to encode the spatial information and get concatenated along with the other additional inputs, which are encoded using fully-connected (FC) networks. The concatenated embedding then gets passed on to GRU, which is followed by two separate heads for value and policy function outputs.

The details of the HRL architecture for \mining{} domain are: Conv1(16x1x1-1)-Conv2-(32x3x3-1)-Conv3(64x3x3-1)-Conv4(32x3x3-1)-Flatten-FC(256)-GRU(512). The HRL architecture for \wob{} domain is almost identical except it replaces the CNN module with fully-connected layers for processing non-spatial information: FC(64)-FC(64)-FC(50)-FC(50)-GRU(512). We use ReLU activation function in all the layers.

For training, we use multi-step actor-critic (A2C) in order to train HRL. We use multi-step learning of $n = 10$, learning rate 0.002, entropy loss weight of 0.01 and critic loss weight of 0.5. We use RMSProp optimizer to train the networks, where its decay rate is set to 0.99 and epsilon to 0.00001. We also clip the gradient by setting the norm equal to 1.0. Most importantly, the network parameters of the HRL agent gets reset for every new task since HRL is not a meta-RL agent.


\end{document}